**Title**: The grip of grammar on meaning uncertainty: cross-linguistic evidence, neural correlates, and clinical relevance

**Authors and affiliations**: Rui He[1], Claudio Palominos[1], Samuele Vallisa[1], Ni Yang[1], Han Zhang[2], Miguel Ángel Santos Santos[3], Neguine Rezaii[4], Sergi Valero[5,6], Yonghua Huang[7], Huan Li[8], Hong Jiang[9,10], Yongjun Peng[11], Maria Francisca Alonso-Sánchez[12], Frederike Stein[13,14], Tilo Kircher[13,14], Philipp Homan[15,16], Iris Sommer[17], Lena Palaniyappan[18,19,20], Wolfram Hinzen[1,21]

[1] Grammar and Cognition Lab, Department of Translation & Language Sciences, Universitat Pompeu Fabra, Barcelona, Spain.
[2] School of Foreign Studies, Guangzhou University, Guangzhou, China.
[3] Department of Neurology Sant Pau Memory Unit, Hospital de la Santa Creu i Sant Pau Biomedical Research Institute Sant Pau, Barcelona, Spain.
[4] Frontotemporal Disorders Unit, Department of Neurology, Massachusetts General Hospital, Harvard Medical School, Boston, USA.
[5] Ace Alzheimer Center Barcelona, Universitat Internacional de Catalunya, C/Gran Via de Carles III, 85 bis, 08028 Barcelona, Spain.
[6] Networking Research Center on Neurodegenerative Diseases (CIBERNED), Instituto de Salud Carlos III, Madrid, Spain.
[7] Department of Language Science and Technology, Saarland University, Saarbrücken, Germany.
[8] Department of Operation and Management, Zhuhai People's Hospital (The Affiliated Hospital of Beijing Institute of Technology, Zhuhai Clinical Medical College of Jinan University), Zhuhai, China.
[9] Department of Statistics, Zhuhai People's Hospital (The Affiliated Hospital of Beijing Institute of Technology, Zhuhai Clinical Medical College of Jinan University), Zhuhai, China.
[10] Faculty of Medicine, Macau University of Science and Technology, Macau, China.
[11] Guangdong Provincial Key Laboratory of Tumor Interventional Diagnosis and Treatment, Department of Radiology, Zhuhai People's Hospital (The Affiliated Hospital of Beijing Institute of Technology, Zhuhai Clinical Medical College of Jinan University), Zhuhai, China.
[12] CIDCL, Escuela de Fonoaudiología, Universidad de Valparaíso, Valparaíso, Chile.
[13] Department of Psychiatry and Psychotherapy, University of Marburg, Marburg, Germany
[14] Marburg University, School of Medicine, Department of Psychiatry and Psychotherapy, Marburg, Germany
[15] Department of Adult Psychiatry and Psychotherapy, University of Zurich, Zurich, Switzerland.
[16] Neuroscience Center Zurich, University of Zurich and ETH Zurich, Zurich, Switzerland.
[17] Center for Clinical Neuroscience and Cognition and Department of Psychiatry, University of Groningen, University Medical Center Groningen, Groningen, the Netherlands.
[18] Douglas Mental Health University Institute, Department of Psychiatry, McGill University, Montreal, Quebec, Canada.
[19] Department of Medical Biophysics, Schulich School of Medicine and Dentistry, Western University, London, Ontario, Canada.
[20] Robarts Research Institute, Schulich School of Medicine and Dentistry, Western University, London, Ontario, Canada.
[21] Institut Català de Recerca i Estudis Avançats (ICREA), Barcelona, Spain.

*Corresponding author(s). E-mail(s): rui.he@upf.edu;

**Competing interests:**

LP reports personal fees for serving as chief editor from the Canadian Medical Association Journals, speaker/consultant fee from Janssen Canada and Otsuka Canada, SPMM Course Limited, UK, Canadian Psychiatric Association; book royalties from Oxford University Press; investigator-initiated educational grants from Janssen Canada, Sunovion and Otsuka Canada outside the submitted work. PH has received grants and honoraria from Novartis, Lundbeck, Mepha, Janssen, Boehringer Ingelheim, OM Pharma and Neurolite outside of this work. TK received unrestricted educational grants from Servier, Janssen, Recordati, Aristo, Otsuka, neuraxpharm. All other authors report no relevant conflicts.

**Acknowledgements:**
We thank all data collectors, providers, and organizations; and participants for their generosity of sharing the data and willingness to take part in the studies. This research was supported by European Research Council (ERC-2023-SyG, 101118756). Views and opinions expressed are however those of the authors only and do not necessarily reflect those of the European Union or the Agency. Neither the European Union nor the granting authority can be held responsible for them. RH was financed by the China Scholarship Council (Grant No. 202108390062) during part of this work. The neuroimaging data was available on OpenNeuro, and most of the publicly available clinical data were obtained from the clinical banks of the TalkBank system. All data use strictly complied with the requirements of the corresponding system. The Zhuhai Alzheimer data acquisition was funded by the Department of Science and Technology of Guangdong Province (grant 2023A0505050118). The TOPSY psychotic data acquisition for this study was funded by Canadian Institutes of Health Research Foundation Grant (FDN 154296) to LP and was supported by the Canada First Excellence Research Fund to BrainSCAN, Western University (Imaging Core); Innovation fund for Academic Medical Organization of Southwest Ontario; Bucke Family Fund, The Chrysalis Foundation, The Children Hospital Foundation, and The Arcangelo Rea Family Foundation (London, Ontario). Compute Canada Resources (Application No. 1530) were used in the storage and analysis of imaging data. LP acknowledges research support from the Canada First Research Excellence Fund, awarded to the Healthy Brains, Healthy Lives initiative at McGill University (New Investigator Supplement); Monique H. Bourgeois Chair in Developmental Disorders and Graham Boeckh Foundation (Douglas Research Centre, McGill University) and a salary award from the Fonds de recherche du Quebec-Sante´ (FRQS). MFAS was funded by the National Agency for Research and Development (ANID), Scholarship Program, Becas Chile 2019, Postdoctoral Fellow 74200048 (MA), during the preprocessing of the TOPSY data. The Marburg psychosis data acquisition has been funded by the DFG-funded research unit FOR 2107 (project number 240413749), with funding allocated to TK (grants KI588/14-1, KI588/14-2, KI588/20-1, KI588/22-1). In addition, it was supported by the German Research Foundation (DFG) through grant STE3301/1-1 (project number 527712970) and by the Von Behring-Röntgen Society (project number 72_0013) both allocated to FS. FS and TK received funds from the DFG Collaborative Research Centre/Transregio 393 (CRC/TRR 393, project number 521379614). The ACE data acquisition was funded by Grifols SA, Life Molecular Imaging, Laboratorios Echevarne, Araclon Biotech, and Ace Alzheimer Center Barcelona, and was supported by the Spanish Ministry of Health from Instituto de Salud Carlos III (Madrid) (FISS PI10/00945) and by the Agència d'Avaluació de Tecnologia i Recerca Mèdiques. It was also funded by Departament de Salut de la Generalitat de Catalunya (Health Department of the Catalan Government) (390) Ace Alzheimer Center Barcelona is part of the Centro de Investigación Biomédica en Red sobre Enfermedades Neurodegenerativas (CIBERNED, Spain) and is one of the participating centers of the Dementia Genetics Spanish Consortium (DEGESCO). This study has also been funded by Instituto de Salud Carlos III (ISCIII) Acción Estratégica en Salud, integrated in the Spanish National RCDCI Plan and fnanced by ISCIII-Subdirección General de Evaluación and the Fondo Europeo de Desarrollo Regional (FEDER-Una manera de hacer Europa) grant PI19/00335.

**Author contributions:**

Rui He: Conceptualization, Methodology, Software, Formal analysis, Investigation, Data Curation, Resources, Writing - Original Draft, Writing - Review & Editing, Project administration, Visualization. Claudio Palominos: Conceptualization, Writing - Original Draft, Writing - Review & Editing, Project administration. Samuele Vallisa: Validation, Writing - Original Draft, Writing - Review & Editing, Project administration. Ni Yang: Data Curation. Han Zhang: Data Curation, Resources. Miguel Ángel Santos Santos: Data Curation, Resources. Neguine Rezaii: Data Curation, Resources. Sergi Valero: Data Curation, Resources, Writing - Review & Editing. Yonghua Huang: Data Curation, Resources. Huan Li: Data Curation, Resources. Hong Jiang: Data Curation, Resources. Yongjun Peng: Data Curation, Resources. Maria Francisca Alonso-Sánchez: Data Curation, Resources. Frederike Stein: Data Curation, Resources, Writing - Review & Editing. Tilo Kircher: Data Curation, Resources, Writing - Review & Editing. Philipp Homan: Writing - Original Draft, Writing - Review & Editing, Funding acquisition. Iris Sommer: Writing - Original Draft, Writing - Review & Editing, Funding acquisition. Lena Palaniyappan: Conceptualization, Data Curation, Resources, Writing - Original Draft, Writing - Review & Editing. Wolfram Hinzen: Conceptualization, Supervision, Project administration, Writing - Original Draft, Writing - Review & Editing, Funding acquisition.

**Abstract**

Isolated word meanings are inherently uncertain. This uncertainty reduces when they are combined and anchored in context. We propose that grammar compresses meaning uncertainty cross-linguistically, which is reflected in brain and selectively disrupted in disorders. Compression was operationalized as the relative difference between non-contextual surprisal estimated from lexical frequency, and contextual surprisal from grammar-sensitive models. In narratives from 20 languages, contextual surprisal reduced frequency-based surprisal. This reduction closely tracked the surprisal cost of reversing word order, and scaled with richer, non-redundant lexis as organized by more complex but optimal dependency structure. During fMRI, surprisal and its reduction explained BOLD activity for comprehension and production in overlapping but distinct regions. Uncertainty reduction was significantly attenuated in aphasia, dementia, and schizophrenia, but remained intact where primary deficit is not language. These findings position uncertainty reduction via grammar as a foundational concept that illuminates principles, brain basis, and disruptions of language.

## 1 Introduction

Words are symbols representing concepts but carry limited message content by themselves. Communicable information arises from their arrangement within sentences and broader discourse. Individual words, in particular, can carry divergent conceptual senses. For instance, *bank* may denote a financial institution or the side of a river, while *steep* can describe a sharp physical incline or, metaphorically, abrupt increases of almost anything. When combined into a noun phrase such as *steep bank*, these options are immediately constrained: *bank* no longer refers to financial institutions, and *steep* loses its metaphorical sense. Adding a determiner, as in *the steep bank*, further anchors the phrase to a particular bank typically recoverable from the prior context. Extending it to a full sentence uttered, for example *John fell on the steep bank yesterday*, a concrete event involving a particular person at a particular time and place will be specified. At this level, with grammar completely in place and the context of an utterance, meaning becomes fully referential: ambiguity is resolved, uncertainty is reduced, and thought is appropriately encapsulated for communication.

This progression from abstract lexical-conceptual meaning in isolated words to referential meaning in coherent sentences thus corresponds to a gradual reduction of uncertainty, which sits at the crossroads of randomness and determinism. Sentence production cannot be wholly random, or it would collapse into noise; nor can it be fully deterministic, because perfectly predictable messages convey no new information. Instead, language appears to exploit a controlled degree of randomness, strategically tuning uncertainty to support meaningful communication.[1,2] Uncertainty in this stochastic process can be quantified by the average negative log probability of words, or surprisal.[3] Starting from Shannon's seminal work on entropy estimated from letter frequencies,[3] linguistic uncertainty has been computed for decades using the frequency of individual words,[4,5] or else the combinations of words (n-grams) that better capture local context.[6,7] As expected, n-gram models typically yield lower uncertainty than unigram models, indicating that contextual information reduces uncertainty.[5,8] However, the context of a word often extends far beyond one or two neighbors, encompassing the entire sentence and discourse, thereby constraining not only which word is likely next but also which concrete situation is being described.[9] Large language models (LLMs), trained to predict a word from rich surrounding context, offer a context-sensitive way to approximate word probabilities. We therefore employ LLM-based surprisal to approximate uncertainty at the referential level, and hypothesize that it should be lower than surprisal from lexical frequency alone.

Constructing meaningful sentences to reduce uncertainty requires the coordination of at least two distinct systems: formal syntax, which imposes generative, rule-governed structure, and lexical semantics, which supplies content-specific input. From a relational perspective, phrases and sentences can be described as networks of asymmetric dependency relations between words. Within the Universal Dependencies framework, structures are generated by a small, finite set of relations, such as *nsubj* (nominal subject), *obl* (oblique nominal) and *amod* (adjectival modifier), which recursively expand heads into larger, infinite configurations (see Methods).[10,11] Each added dependency both enriches the message and prunes the remaining hypothesis space. Once *fell* is established as the predicate, it introduces open slots for an agent, an affected location or object, and a time, thereby constraining what kinds of lexical items can legitimately appear in those positions. Attaching *John* to *fell* fills the agent role, attaching *bank* constrains the event to involving a physical surface, and modifying *bank* by *steep* further suppresses the financial reading of *bank* and the metaphorical reading of *steep*. Such a formal syntactic system can be situated along the Chomsky hierarchy,[12] where context-free grammar marks a qualitative change in the probabilistic structure of linguistic sequences: only beyond this level can a system recursively generate unbounded, hierarchically organized dependencies[13,14], which, as argued above, are needed to progressively narrow down referential possibilities. Recent statistical physics studies have shown that language systems with at least context-free complexity can settle into a critical regime with long-range correlations,[15] intermediate between a random phase and an over-regular, frozen phase. In such a regime, each added dependency both opens and constrains options, so that deep syntactic "structure" emerges as the mechanism by which uncertainty over lexical inputs is reduced without collapsing them into a fixed code.[16] On this view, human grammar could be broadly conceived as an integrated system jointly encompassing formal structural rules and lexical-semantic contents as their input, thereby providing rich contextual constraints that help regulate lexical uncertainty and transforming it into referentially stable, information-rich messages.

This shift towards an optimized state through particular levels of syntactic organization shows reflexes in neural correlates and psycholinguistic metrics. Linguistic uncertainty has been linked to a wide range of behavioral and neural measures, including reading time,[17–19] eye movement,[20] electrophysiological signals (most notably, N400 and P600),[21] and activity across distributed cortical networks during naturalistic comprehension.[22–25] Increased uncertainty in spontaneous speech has been reported in clinical conditions that affect language, such as post-stroke aphasia,[26] probable Alzheimer's disease,[27] and psychosis.[28–30] A link between grammar and uncertainty has also been demonstrated, showing that syntactic structure building is tightly coupled to lexical surprisal during naturalistic comprehension and is modulated by sentence structure relative to word lists.[31,32] In patients with non-fluent primary progressive aphasia, Rezaii et al.[33] reported a trade-off between lexical richness and syntactic complexity, which helped maintain a relatively stable level of overall uncertainty. Together, these converging findings suggest a close relationship between grammatical organization and uncertainty regulation, and are broadly compatible with predictive-processing accounts in which linguistic expectations are shaped across multiple representational levels.

Here we contrast LLM-derived surprisal with the baselines estimated from lexical frequency, targeting uncertainty compression under constraints from grammatical context, to then relate this to independent measures of grammatical organization that capture the complexity of syntactic structure and lexical meaning. Subsequently we test whether surprisal and its reduction exhibit identifiable neural signatures during naturalistic comprehension and speech production. We hypothesize that these effects should particularly explain an information gain obtained through grammar in comprehension, and that in production, this gain may be attenuated, on the assumption that speakers coordinate grammar with a representation partially pre-planned before linear articulation begins. Finally, we examine whether these computations are selectively altered in clinical populations where language is affected. By linking linguistic, neural, and clinical levels of description, we argue that reducing uncertainty via grammar constitutes a core organizing principle of human language.

## 2 Results

### 2.1 Uncertainty reduced in 20 languages: from lexical to contextual surprisal

We first tested the hypothesis that grammatical context systematically reduces lexical uncertainty across typologically diverse languages. Using StoryDB Wikipedia movie plots,[34] we analyzed 20 languages selected to balance typological coverage and data availability (10 Indo-European and 10 non-Indo-European languages, with the latter drawn from distinct language families). Following the pipeline in **Figure 1**A, across all 20 languages, contextual surprisal was consistently lower than frequency-based surprisal, as seen in **Figure 1**B. Despite substantial diversity in linguistic typology, there is a robust pattern: surprisal is reduced when predictions are conditioned on grammatical context rather than lexical frequencies alone. By grammar, as stated above, we do not mean syntax in isolation, nor do we assume that BERT provides a pure syntactic parser. Rather, the measure captures the extent to which the ordered linguistic context constrains word choice through grammatical organization. We then reversed word order within sentences to test whether this advantage depends on ordered structure. Reversal increased contextual surprisal (**Figure 1**C), consistent with a loss of grammar-mediated predictability. This manipulation is used as a coarse perturbation of structured grammatical organization, not as a selective lesion of syntax. Surprisal reduction, defined as the proportional decrease from frequency-based to contextual surprisal, strongly correlated with the surprisal increase from original to reversed sentences (**Figure 1**C; meta-analytic Spearman's $\rho$=0.939, 95% CI: 0.929 to 0.950). Sentences benefiting most from contextual prediction also incurred the largest cost when ordered structure was disrupted. Word-order reversal was used here as a coarse perturbation of structured grammatical organization. Because the two measures rely on distinct contrasts, the association is unlikely to be purely tautological, although shared sentence-level variance may contribute.

## 2.2 Formal syntax and lexical semantic space jointly relate to uncertainty reduction

We next asked how much variability in uncertainty reduction can be accounted for by formal syntax and the lexical–semantic input it organizes, respectively. For each language, we summarized within-text normalized surprisal reduction and related these scores to a set of syntactic and lexical semantic metrics (**Figure 2**A).

### 2.2.1 Syntactic topology: balanced hierarchical chunking predicts stronger uncertainty reduction

We first tested whether uncertainty reduction scales with the *topological organization* of dependency structure. Sentences were parsed into dependency trees (head-dependent relations), and we quantified: (i) global tree balance using the B2 balance index,[35] and (ii) hierarchical heterogeneity using subtree-size unevenness. Higher B2 values indicate more evenly distributed branching, whereas subtree-size unevenness indicates broader variation in subtree sizes, consistent with more graded multi-level chunking. Across all 20 languages, surprisal reduction was positively correlated with both B2 index (**Figure 2**C; meta-analytic Spearman's $\rho$=0.213, 95% CI: 0.176 to 0.251) and subtree-size unevenness (**Figure 2**C; meta-analytic Spearman's $\rho$=0.264, 95% CI: 0.219 to 0.310). Sentences with more balanced and hierarchically heterogeneous dependency structure showed greater compression of lexical-conceptual uncertainty.

### 2.2.2 Packing constraints: hierarchical complexity and efficient linearization relate to uncertainty reduction

Beyond tree topology, we tested whether uncertainty reduction relates to how dependencies are organized in hierarchical and linear structure.[36] As languages must sustain enough hierarchical depth for predictions to propagate across clauses and phrases,[22,25] we quantified hierarchical complexity using the mean hierarchical distance, defined as the average vertical separation between heads and dependents in a dependency tree (**Figure 2**A).[37] Greater hierarchical depth allows uncertainty about upcoming materials to be resolved over longer spans of structure, rather than locally at each word. At the same time, distances between directly dependent words should reach an optimal value, favoring relatively short dependencies over much longer ones, without strictly attaining the formal minimum predicted by uniformly random linear arrangement.[38] Across all 20 languages, uncertainty reduction was positively correlated with both dependency optimality (**Figure 2**C; meta-analytic Spearman's $\rho$=0.231, 95% CI: 0.188 to 0.275), and mean hierarchical distance (**Figure 2**C; meta-analytic Spearman's $\rho$=0.229, 95% CI: 0.184 to 0.274). Sentences with stronger uncertainty reduction therefore tended to combine sufficient hierarchical depth with efficient linear dependency organization.

### 2.2.3 Lexical-semantic space: surprisal reduction benefits from non-redundant, compressible lexical input.

Finally, we tested whether the organization of lexical meaning before syntactic composition contributes to uncertainty reduction. We embedded all non-stopword tokens using distributional static fastText vectors,[39] and estimated the intrinsic dimensionality (ID) of the resulting semantic spaces using GRIDE algorithm[40] (**Figure 2**B, a-c). ID was estimated across tokens, indexing lexical diversity versus redundancy, and across embedding dimensions, indexing compressibility into latent semantic directions, as shown in **Figure 2**B, d. Across languages, surprisal reduction was positively correlated with ID in the token manifold (**Figure 2**C; meta-analytic Spearman's $\rho$=0.252, 95% CI: 0.184 to 0.319), and negatively correlated with ID in the feature manifold (**Figure 2**C; meta-analytic Spearman's $\rho$=−0.236, 95% CI: −0.293 to −0.180). Thus, stronger uncertainty reduction was associated with richer lexical content and a more compressible lexical–semantic space, suggesting that grammar-driven compression is facilitated when the lexical input is rich yet geometrically low-dimensional.

#### 2.2.4 Validating the results with generative language model

To test robustness, we replicated analyses in section 2.1 and 2.2 using a multilingual generative language model, which estimates token probabilities from preceding context only rather than bidirectional context. As generative models compute probabilities left-to-right, they offer a complementary test of uncertainty reduction. The results were largely replicated, as reported in the supplementary materials (section 5).

### 2.3 Overlapping but distinct neural correlates in language production vs. comprehension

As shown in **Figure 3**A, we compared word-by-word neural correlates of uncertainty metrics in spontaneous speech using two fMRI datasets: spontaneous speech production (recall) and naturalistic narrative comprehension. The first dataset recruited 16 participants to recall the events of the *Sherlock* episodes, immediately after they watched them, in an fMRI scanner. [41] The second dataset recruited 18 subjects listening to the recording of the recall by one of the participants from the first dataset. The fMRI images were preprocessed by fMRIprep[42], followed by regression of nuisance signals. Cortical and subcortical time series were extracted, respectively, using the Schaefer-400 parcellation[43] and the Melbourne S1 atlas[44] (**Figure 3**B). Linguistic measures were convolved the BOLD signals using a canonical hemodynamic response function and standardized within participants (**Figure 3**A). For each parcel and dataset, we fitted linear mixed-effects models predicting BOLD responses from each uncertainty metric, controlling for syllable count and including participant random intercepts. Across both datasets, lower surprisal and greater surprisal reduction were associated with stronger BOLD responses in distributed cortical and subcortical regions (**Figure 3**C,D), encompassing both perisylvian language cortex and, notably, regions outside it, including the subcortical structures.

During comprehension, the effects of surprisal and its reduction spanned a broad set of default-mode regions (including posterior temporal-parietal cortex, posterior cingulate cortex and parahippocampal gyrus), and extended to frontal control regions. Subcortically, associations were observed in the basal ganglia (pallidum, putamen, and caudate), amygdala and thalamus. Effects of frequency-based surprisal extended to early sensory cortices within the visual and somatomotor networks, as well as to the anterior temporal pole, a putative lexical-semantic hub,[45,46] with higher surprisal associated with reduced activation, consistent with an effect of decreased predictability at perceptual and conceptual levels. By contrast, BERT-based (contextual) surprisal and the surprisal reduction showed comparatively more focal cortical associations, largely overlapping with the frequency-based map but sparing early sensory cortices and the anterior temporal pole, while exhibiting additional effects in subcortical regions of basal ganglia, amygdala and thalamus.

During speech production (recall), overall associations with surprisal metrics were less extensive (**Figure 3**D).

Frequency-based surprisal was significant in bilateral precentral and postcentral regions and medial frontal areas, with additional effects extending to the precuneus but no reliable subcortical associations. In contrast, BERT-based surprisal and the normalized surprisal difference engaged postcentral gyrus, temporal pole, medial dorsal superior frontal gyrus, and subcortical regions including thalamus and caudate. This weaker production pattern may reflect the idea that little uncertainty is to be reduced during production, speakers typically operate with stronger internal representations of upcoming content, leaving less uncertainty to be resolved word by word.

### 2.4 Alterations in clinical populations

Finally, we asked whether uncertainty-related computations are selectively disrupted in clinical populations with clinically meaningful impairments in language function. We analyzed spontaneous speech from nine clinical cohorts, comparing each clinical group to healthy controls (**Figure 4**A), and relating uncertainty measures to independently collected clinical or neuropsychological indices. Unless stated otherwise, all reported $p$ values were corrected using two-step false-discovery rate procedure.

#### 2.4.1 Focal lesion and primary language disorders

In post-stroke aphasia, a canonical model of acquired language impairment following left-hemisphere lesions,[47] patients with mixed-type aphasia ($N$=49)[48] showed higher BERT-based surprisal ($z$=6.884, $p$<0.001) and attenuated surprisal reduction ($z$=-6.765, $p$<0.001), alongside significantly lower frequency-based surprisal ($z$=-2.528, $p$=0.011). In contrast, these effects were absent in a cohort with right-hemisphere damage (RHD; $N$=37),[49] who performed below the normal range in the attention and visuospatial domains, but within the normal range on the executive function, memory and language domains.[49] No significant effects were observed in this RHD cohort. Similar alterations were observed in primary progressive aphasia (PPA). PPA is a neurodegenerative syndrome with three variants, where language impairment is the leading initial symptom in disease progression.[50] In all three variants, the non-fluent ($N$=28), semantic ($N$=23), and logopenic ($N$=25) variants, we observed significantly lower frequency-based surprisal (non-fluent: $z$=-2.076, $p$=0.038; semantic: $z$=-6.135, $p$<0.001; lvPPA: $z$=-5.462, $p$<0.001), significantly higher surprisal from BERT (non-fluent: $z$=4.819, $p$<0.001; semantic: $z$=4.041, $p$<0.001; lvPPA: $z$=5.111, $p$<0.001), and significantly less surprisal reduction (non-fluent: $z$=-5.188, $p$<0.001; semantic: $z$=-5.139, $p$<0.001; lvPPA: $z$=-6.190, $p$<0.001) .

#### 2.4.2 Neurodegenerative cohorts across the Alzheimer spectrum

Across three Alzheimer-related cohorts, BERT-based surprisal was generally elevated and surprisal reduction attenuated. In the ADReSS cohort,[51] patients with probable Alzheimer's disease (pAD, $N$=122) showed higher surprisal from BERT ($z$=3.730, $p$<0.001), and less surprisal reduction ($z$=-4.690, $p$<0.001), with insignificantly lower frequency-based surprisal ($z$=-1.047, $p$=0.295). These results broadly replicated in a Chinese cohort where all patients were diagnosed as Alzheimer’s disease (AD, $N$=30) with cerebrospinal fluid (CSF) biomarkers. Surprisal from frequency ($z$=2.382, $p$=0.017) and BERT ($z$=3.7175, $p$=0.001) increased significantly, while surprisal reduction was attenuated ($z$=-2.393, $p$=0.017). The final cohort made finer-grained distinctions within the Alzheimer spectrum, including subjective cognitive decline (SCD, $N$=31),[52] where individuals report cognitive worsening despite largely normal test performance; patients with mild cognitive impairment (MCI, $N$=39),[53] marked by objective impairment without dementia-level functional decline; and patients with pAD ($N$=31). Patients with MCI ($z$=-2.729, $p$=0.019) and pAD ($z$=-2.688, $p$=0.022) both showed attenuated surprisal reduction. BERT-based surprisal was only significantly higher in MCI ($z$=2.219, $p$=0.040). No significant effects were observed for frequency-based surprisal. No significant effects were observed in adults with subjective cognitive decline SCD.

#### 2.4.3 Neurodevelopmental disorders

We next examined two neurodevelopmental conditions, autism spectrum disorder (ASD, $N$=46) and attention-deficit/hyperactivity disorder (ADHD, $N$=37), using the Dutch Asymmetries corpus.[54] ASD is defined by early-emerging social-communication differences and often involves pragmatic and structural language difficulties,[55] whereas ADHD is primarily characterized by attentional and executive-control difficulties, with language problems reported but not defining the disorder.[56] Relative to typically developing children, neither ASD nor ADHD showed reliable changes in frequency-based surprisal. Children with ASD exhibited higher BERT-based surprisal ($z$=2.456, $p$=0.021) and attenuated surprisal reduction ($z$=−2.590, $p$=0.021). Effects in ADHD were weaker and did not reach significance (BERT-based surprisal: $z$=2.062, $p$=0.059; surprisal reduction: $z$=−2.070, $p$=0.059).

#### 2.4.4 Psychosis cohorts

We further investigated psychosis, where disturbances of language and thought have been argued to be central in the pathology.[57] In the Canadian TOPSY cohort,[58] patients with first-episode, antipsychotic-naïve schizophrenia ($N$=72) showed elevated BERT-based surprisal ($z$=2.435, $p$=0.022) and attenuated surprisal reduction ($z$=−2.740, $p$=0.018), with no reliable frequency effect ($z$=−0.783, $p$=0.434). In less acute and less severe patients with chronic schizophrenia ($N$=20), both frequency-based and BERT-based surprisal were higher (frequency: $z$=2.603, $p$=0.014; BERT: z=2.733, $p$=0.014) and surprisal reduction was attenuated ($z$=−2.445, $p$=0.014). These patterns replicated in an independent Marburg schizophrenia cohort,[59] with higher BERT-based surprisal ($z$=3.224, $p$<0.001) and lower surprisal reduction ($z$=-3.177, $p$<0.001) in the schizophrenia group ($N$=20), alongside no reliable frequency effect ($z$=0.431, $p$=0.222). Notably, comparable effects were not observed in the schizoaffective ($N$=22) or major depression ($N$=42) groups from the same cohort.

#### 2.4.5 Correlations with clinical and neuropsychological scales

Together, these cohorts suggest reduced uncertainty reduction in disorders with prominent language or higher-order cognitive impairment, with weaker or absent effects when language was relatively spared or not the primary deficit, as in RHD, ADHD, schizoaffective disorder, and major depression. To better understand the cognitive and clinical relevance, we correlated uncertainty measures with behavioral and clinical indices within patient groups. Greater surprisal reduction was associated with better aphasia battery performance and memory/language scores in aphasia (**Figure 4**B,D). In Alzheimer-related cohorts, greater surprisal reduction correlated with better global cognitive performance (**Figure 4**C). In the Marburg cohort, surprisal reduction showed strong associations with verbal episodic memory and executive function (**Figure 4**E). In the TOPSY cohort, surprisal reduction was related to clinical indices of thought disorganization and impoverishment (**Figure 4**F).

## 3 Discussion

In this study, we examined how grammatical organization constrains uncertainty in lexical meaning space, using surprisal and its contextual reduction as computational approximates. Although context as integrated by LLMs exceeds grammar narrowly construed, convergent evidence linked surprisal reduction to syntactic structure, lexical-semantic organization, distributed neural responses, and selective clinical alterations.

Together, these findings motivate a mechanistic account in which grammar constrains meaning uncertainty. Linguistic analyses identify structural and lexical-semantic conditions under which this reduction is strongest; MRI analyses reveal distributed neural correlates; the clinical analyses show selective attenuation in disorders affecting language and higher-order cognition, which may reflect the language-specific vulnerability in these conditions.

From a linguistic perspective, referential meaning is central to human language and differs categorically from lexical-conceptual meaning. Lexical-conceptual meaning concerns general concepts associated with word forms, whereas referential meaning identifies entities, events, and relations in context.[60,61] Referential meaning depends on grammatical organization, through which content words are contextualized by function words and morphemes (*the*, *I*, *-ed,* etc.), which do not add lexical content but locate concepts and events in space, time, and discourse.[62] While lexical-conceptual meaning can be approximated by the distributional information of words, such as n-gram frequency and static distributional embeddings,[63] referential meaning remains harder to operationalize. LLMs provide a possible approximation via computing word probabilities and representations incorporating the linguistic context and thereby grammatical organization. In other words, while frequency-based surprisal only tracks the general availability of words in the language, BERT-based surprisal reflects how these probabilities are redistributed as context is integrated. Using this simple contrast, we confirmed across 20 typologically diverse languages that there was a consistent reduction of surprisal from lexical frequency to contextual surprisal, indicating that contextual integration reduces uncertainty relative to lexical baselines. This result is expected in a broad language-modelling sense, but gains interpretive force firstly from the word-order reversal analysis. Reversing sentences inflated contextual surprisal, and the magnitude of the contextual gain from frequency to BERT was tightly coupled to the cost of disrupting word order. Thus, what BERT adds beyond lexical availability is closely aligned with what reversal removes. Together, we therefore interpret the BERT-frequency contrast as a proxy for grammar-mediated constraint on meaning uncertainty, while recognizing that BERT is not a purely syntactic parser and that contextual information also includes semantic, discourse, and world-knowledge regularities. On this view, grammar goes beyond a purely formal artifact, but is a functional system defined by what it does: coordinating the composition of lexical meanings through dependency relations so as to constrain, sharpen, and stabilize the lexical sematic space, and allowing the emergence of a referential interpretation.

The correlational analyses further support the interpretation and clarify which aspects of linguistic structure contribute to surprisal reduction. Dependency topology was systematically related to uncertainty reduction: sentences with more balanced branching and greater hierarchical heterogeneity showed stronger compression of lexical uncertainty. These results suggest that grammar-mediated constraints are strongest when head–dependent relations are neither concentrated in a single hub nor restricted to uniformly shallow attachments. Instead, a richer distribution of dependency configurations may provide multiple, non-redundant sources of contextual constraint. Consistent with this interpretation, uncertainty reduction also increased in sentences combining sufficient hierarchical depth with efficient linear dependency organization. This profile aligns with the idea that effective syntactic constraints should be neither purely local as when being restricted to adjacent words, nor overly diffuse, with relations spread so widely that they become harder to exploit. In parallel, surprisal reduction was also related to the organization of lexical-semantic space. Stronger reductions occurred when lexical representations were both expressive and compressible. Token-manifold intrinsic dimensionality was positively associated with surprisal reduction, consistent with the need for sufficient lexical diversity rather than trivial repetition. By contrast, feature-manifold intrinsic dimensionality was negatively associated with surprisal reduction, suggesting that contextual constraints are easier to apply when lexical meanings occupy a more organized, lower-dimensional semantic manifold. Thus, the grammar-sensitive component of surprisal reduction has two faces: a formal condition, indexed by dependency structure, in which syntactic organization is topologically balanced, hierarchically articulated, and efficiently linearized; and a meaning condition, indexed by semantic geometry, in which lexical input is diverse enough to express meaningful distinctions but also structured enough to be compressed by context.

To connect these computational signatures to neurobiology, we examined neural correlates of surprisal and surprisal reduction during speech comprehension and production with similar stimuli. In both conditions, significant effects extended beyond perisylvian cortex and the canonical language-selective network, for example as defined by localizer tasks contrasting sentences with its scrambled version or jabberwocky sentences from pseudo-words.[64] This distribution suggests that uncertainty processing is not purely "linguistic" in one traditional, narrow sense, but recruits systems supporting situation-model construction and top-down control. Specifically, during narrative comprehension, associations involved default-mode regions, frontal control regions, and subcortical structures including basal ganglia, thalamus, and amygdala. Interpreting surprisal reduction as an index of grammar-mediated constraints on lexical predictability, these associations suggest a role for cortico–subcortical loops in implementing uncertainty compression via operations such as gating, updating, and precision modulation, which are often emphasized in predictive coding, rather than reflecting only representational encoding of the auditory input and its contextualization within the cortical hierarchy.[65] During speech production, associations were overall weaker and more spatially restricted. This asymmetry may reflect the difference between comprehension and production: listeners must incrementally resolve externally imposed uncertainty, whereas speakers may typically operate with stronger internal constraints and utterance plans. Nevertheless, surprisal reduction in production still involved cortico-subcortical systems, including motor-articulatory regions, medial frontal cortex, temporal pole, thalamus, and caudate. These regions overlap with systems affected in clinical populations with language deficits, such as aphasia[66,67], dementia[68–71], and psychosis.[72,73] If surprisal reduction reflects grammar-mediated uncertainty compression supported by these circuits, then disorders disrupting these systems should show measurable alterations in natural speech. This expectation is consistent with prior evidence that the probabilistic structure of spontaneous connected speech is altered across clinical populations.[27–29,74]

Across nine clinical cohorts, we observed a convergent attenuation of surprisal reduction across focal language disorders (post-stroke aphasia and three PPA variants), Alzheimer-spectrum conditions (MCI and AD), autism, and schizophrenia. This suggests that a shared vulnerability may lie in mechanisms transforming lexical–conceptual content into contextually constrained, referential meaning. Importantly, the pattern was not generic to all clinical diagnoses. Comparable effects were weaker or absent in conditions where language was relatively spared or not the primary deficit, including right-hemisphere damage, subjective cognitive decline, ADHD, schizoaffective disorder, and major depression. The contrast between schizophrenia and affective diagnoses within the same dataset was particularly informative, arguing against a purely general effect of clinical status or sampling.

In nine clinical cohorts, we observed a convergent attenuation of surprisal reduction across focal language disorders (post-stroke aphasia and three PPA variants), Alzheimer-spectrum conditions (MCI and AD), autism, and schizophrenia. Such attenuation suggests that a shared vulnerability may lie in the mechanisms transforming lexical conceptual content to referential meaning in these clinical populations. Importantly, the pattern is not generic to any clinical diagnosis. Conditions where language is not the primary deficit or is relatively spared (e.g., right-hemisphere damage with largely preserved language domains, subjective cognitive decline with no clinical meaningful marker, attention deficits, schizoaffective disorders, and major depression disorders). This selectivity argues against a trivial explanation in terms of sampling or general clinical status, and points to a more specific link between reduced uncertainty compression and disorders comprising deficits in core language computations. The contrast between schizophrenia and affective diagnoses within the same dataset is particularly informative. Correlations with clinical and neuropsychological measures further support the functional relevance of surprisal reduction. Greater surprisal reduction was associated with better communicative performance in aphasia and better global cognition in Alzheimer-related cohorts. It was also related to memory, especially episodic memory, and executive function, consistent with a high-level coupling between language and other cognitive systems. In schizophrenia, thought disorganization and impoverishment were associated with altered contextual surprisal and surprisal reduction, suggesting that clinically defined thought disorder has a measurable computational readout in spontaneous speech. This further supports a constrained language-thought linkage, where disturbances of thought become detectable when language fails to impose an appropriate probabilistic regime.

Several limitations should be noted. First, surprisal reduction remains an indirect measure of the conceptual-referential meaning distinction. Contextualization does not equate with referential meaning,[60,75] while relative lexical frequency captures only selected aspects of lexical-conceptual meaning. Further efforts should seek for more direct measures of both conceptual and referential meaning. Relatedly, our analyses were conducted at the sentence level to focus on grammatical context, but discourse-level constraints such as topic continuity and narrative structure are also central to meaning construction. Future extensions should quantify how uncertainty reduction accumulates across larger discourse units. Second, the neural analyses were correlational and rely on temporal alignment assumptions (hemodynamic response convolution and word-by-word regressors). Future studies could integrate measures with higher temporal resolution, and manipulate predictability experimentally within naturalistic paradigms. Third, the clinical analyses aggregated heterogeneous datasets differing in language, recording conditions, clinical instruments, and speech elicitation tasks. Although cross-cohort convergence is a strength, the present design cannot fully separate diagnosis-related effects from task-specific elicitation effects. The GEE models estimated overall clinical effects while accounting for correlated observations, but did not explicitly model task effects. Prospective studies with harmonized protocols and longitudinal follow-up are needed to establish diagnostic specificity and clinical utility. Finally, although we frame our results in terms of “reduction”, the theoretical target is better understood as an optimal *balance* between purely deterministic systems, where surprisal approaches zero, and fully chaotic systems, where surprisal remains very high. Future work should therefore complement surprisal-based contrasts with methods that directly quantify this balance.

# 4 Methods

## 4.1 Datasets

We analyzed three classes of datasets: cross-linguistic narratives from StoryDB, two fMRI datasets probing uncertainty-related neural correlates, and nine clinical cohorts assessing clinical selectivity. All analyses used orthographic transcripts and a shared surprisal-based uncertainty framework. Dataset-specific details are summarized below.

### 4.1.1 Crosslingual narratives from the StoryDB datasets

The StoryDB dataset contains movie plots from Wikipedia in 42 different languages, and is publicly available https://drive.google.com/drive/folders/1RCWk7pyvIpubtsf-f2pIsfqTkvtV80Yv.[34] We analyzed 20 languages, which were selected to balance typological coverage and within-language data availability, including 10 Indo-European and 10 non-Indo-European languages. Each of the 10 non-Indo-European languages was drawn from a different language family. With the exception of Tamil, each selected language contributed approximately 1,000 or more texts. Languages from the Kartvelian and Kra–Dai families, as well as constructed languages, were excluded owing to insufficient data. To improve comparability across languages, we restricted the corpus to passages with an average sentence length between 10 and 50 words and with no sentence exceeding 100 words. From the filtered English set, we randomly sampled 500 passages to define a reference distribution. For each remaining language, we then selected a target set of passages to match the English sample using propensity score matching (PSM) implemented in PsmPy (version 0.3.13). Specifically, for each language separately, we concatenated the 500 English passages (coded as treatment=1) with all eligible passages from the target language (coded as treatment=0), and estimated propensity scores via logistic regression on the number of sentences and averaged sentence length. We performed 1:1 nearest-neighbor matching on the logit propensity score (matcher=propensity_logit), without replacement, using a caliper of 0.2, and dropped unmatched observations. The matched non-English passages (*n*=500 when available) were retained for downstream analyses. For languages with insufficient eligible passages to support matching to the target sample size, we included all passages meeting the length criteria. **Table 1** summarizes the resulting dataset.

#### 4.1.2 FMRI datasets for language comprehension and production

We analyzed two fMRI datasets sampling naturalistic language production and comprehension using similar contents. The production data were collected by Chen et al.[77] and publicly available on OpenNeuro (https://openneuro.org/datasets/ds001132/versions/1.0.0). Transcripts of the recordings were also publicly available from the main authors.[78] We reused the cleaned and annotated transcripts from Giglio et al (https://osf.io/qjmky/overview).[79] A total of 16 participants were included in this dataset. Participants produced an unguided spoken recall of the events during fMRI scanning, which was recorded simultaneously, immediately after watching the first part of BBC's *Sherlock*. The comprehension dataset comprised an independent group of 18 participants who listened during fMRI to the recording of one participant's spoken recall from the production dataset (sub-19 to sub-36). These data were collected by Zadbood et al.[80] and publicly available on OpenNeuro (https://openneuro.org/datasets/ds001110/versions/00003). These two datasets were selected due to their comparability. First, the scanning parameters were identical, as reported in the corresponding papers. Second, the tasks were similar in content, with the linguistic input of the compression as one of the linguistic outputs of the production on the same contents.

#### 4.1.3 Clinical spontaneous speech cohorts

We analyzed nine clinical cohorts to characterize the clinical selectivity and relevance of surprisal metrics in spontaneous speech. Cohort-specific details are summarized below, and additional metadata (sample sizes, language, elicitation task, and clinical scales) are provided in the supplementary materials. There were participants who produced only very short sentences that were all excluded and hence not entered our analyses. Here we reported the number and statistics based on participants who entered our analyses.

1. Post-stroke aphasia samples were obtained from the Olness corpus and available on AphasiaBank (https://talkbank.org/aphasia/access/English/NonProtocol/Olness.html). This dataset provided spontaneous speech samples in English from 49 individuals with mixed-type aphasia and 32 controls with no brain injury as described in prior work by Olness and Ulatowska.[81] The speech was elicited by a variety of discourse tasks including picture description, story recall, narrative completion, and free speech on given topics.
2. The right-hemisphere damage (RHD) cohort was recruited by Minga et al.[49] and available on RHDBank (control: https://talkbank.org/rhd/access/English/Control.html; RHD:

https://talkbank.org/rhd/access/English/RHD.html). Participants in this cohort were English speakers from the United States, including 35 controls and 37 patients. The patients had single right hemisphere stroke at least six months before the recruitment. The speech was elicited by a variety of discourse tasks from the RHDBank Discourse Protocol, including free speech, picture descriptions, story retelling, procedural speech, and question answering. All participants were evaluated using the Cognitive Linguistic Quick Test. In this cohort, these patients performed below the normal range in the attention and visuospatial domains, but within the normal range (on the low end) on the executive function, memory and language domains.

3. The primary progressive aphasia (PPA) cohort was shared by Rezaii et al.[82] and publicly available on OSF (https://osf.io/sr3ag/). This cohort includes picnic picture descriptions from 53 controls and 76 patients with PPA. Patients were categorized into three variants of PPA, including 28 non-fluent variants (nfvPPA), 23 semantic variants (svPPA), and 25 logopenic variants (lvPPA). All data were in English.
4. The ADReSS challenge provided speech transcripts from 122 patients with probable Alzheimer's disease (pAD) and 115 matched controls.[51] Speech was in English and elicited by picture description using the Cookie Theft picture.
5. The ACE dataset collected speech data with a fine-grained categorization of the Alzheimer's disease (AD) continuum, including 17 healthy older adults, 31 healthy older adults with subjective cognition decline, 39 patients with mild cognitive impairments, and 31 patients with pAD. This dataset was collected by the first lab, with recruitment details in the supplementary materials. All participants completed two speech tasks, past-directed narratives and picture descriptions. They could choose to speak in Spanish or Catalan.
6. The Zhuhai AD cohort recruited patients with AD whose diagnoses were confirmed using cerebrospinal fluid (CSF) biomarkers, alongside cognitively healthy controls. CSF-confirmed AD speech datasets are uncommon, as most speech studies rely on clinical diagnosis without biomarker verification. Because contemporary diagnostic and staging frameworks increasingly define Alzheimer's disease biologically, biomarker confirmation from CSF is clinically important for improving diagnostic specificity, particularly in specialist memory-clinic pathways.[83] This datasets included spontaneous speech from 34 controls and 30 patients with AD elicited by a wide range of different tasks. All participants were native Chinese speakers, and could choose to speak in either Mandarin or Cantonese.
7. The Asymmetries Project collected children speech during a structured storytelling task in Dutch.[84] This dataset included 69 typically developing children, 46 children with Autistic Spectrum Disorder (ASD), and 37 children with Attention Deficit Hyperactivity Disorder (ADHD). The data are available on ASDBank (https://talkbank.org/asd/access/Dutch/Asymmetries.html) and CHILDES (https://talkbank.org/childes/access/DutchAfrikaans/Asymmetries.html).
8. The first psychosis cohort included spontaneous speech in English from 39 healthy controls (HC), 72 patients with first-episode psychosis (FEP), and 20 patients with chronic schizophrenia (CSZ), as a part of the Tracking Outcomes in Psychosis (TOPSY) study (https://clinicaltrials.gov/study/NCT02882204).[28,85] In brief, FEP subjects had <2 weeks of lifetime antipsychotic exposure and in most cases were assessed in the first week of referral to the first-episode psychosis team. Only the data from FEP whose diagnosis remained stable (as schizophrenia, excluding those who had bipolar disorder or depressive psychosis) after 6 months of follow-up are included in this study. CSZ consisted of 20 participants that were clinically stable on long-acting injectable medications with >3 years since illness onset and no recorded hospitalization in the past year and receiving community-based care from physicians affiliated to a first-episode clinic (PEPP, London Ontario). Importantly, all participants were recruited regardless of the status of disorganization/thought disorder in their prior history, which was in order not to bias our sample towards language-related symptomatology. All participants were asked to describe three pictures from the Thematic Apperception Test and were given one minute for each image.
9. The second psychosis cohort recruited German speakers from Marburg, including 43 healthy controls (HC),

42 patients with major depressive disorder (MDD), 22 patients with schizoaffective disorders, and 20 patients with non-affective schizophrenia.[59] All participants were asked to describe four pictures from rom the Thematic Apperception Test and were given three minutes for each image.

### 4.2 Data preprocessing

#### 4.2.1 StoryDB text preprocessing

For the StoryDB narratives, raw texts were processed separately for each language. We loaded the plot files and excluded entries with missing text per-language. As the raw passages contain residual HTML markers, we removed the HTML tags and recovered plain texts using the python package lxml (version 5.3.0). The clean texts were then segmented into sentences using the Trankit language-specific pipelines (version 1.1.2).[86]

#### 4.2.2 fMRI preprocessing

Functional MRI data were preprocessed using fMRIPrep v21.0.1 executed in a Singularity container.[42] Additional information on the preprocessing is available in the supplement. Preprocessing was run per participant, with the output images registered in a standard MNI space (MNI152NLin2009cAsym). Repetition time (TR) was read from the meta files as 1.5 seconds. We performed parcel-wise time-series extraction using a combined cortical–subcortical atlas from Schaefer 400 parcels[43] for the cortical regions and Melbourn S1 atlas[44] for the subcortical regions. The python package nilearn (version 0.12.0) was used for parcel-wise time-series extraction (NiftiLabelsMasker, no z-standardization), and nuisance regression (nilearn.interfaces.fmriprep.load_confounds). Confounds were selected using Nilearn's fMRIPrep interface with the default strategy, which includes head-motion estimates, discrete cosines transformation basis regressors to handle low-frequency signal drifts, and white matter and cerebrospinal fluid signals. The two parcellations are visualized in **Figure 3**B.

#### 4.2.3 Spontaneous speech transcripts preprocessing

All transcripts were provided by the original corpus sources and were manually transcribed, with the exception of the Zhuhai CSF-confirmed AD dataset. For the Zhuhai recordings, manual transcription was performed by trained research assistants who were native or near-native speakers of the relevant variety (Mandarin and/or Cantonese), and transcripts were subsequently quality-checked by the first author, who is fluent in both Mandarin and Cantonese. All recordings were transcribed in simplified Chinese characters to ensure consistency across participants and tasks. For Cantonese speech, Cantonese-specific orthographic characters were used where appropriate to preserve lexical distinctions in the original variety. All recordings were double-checked against the audio to resolve ambiguities in segmentation and lexical choice, and discrepancies were resolved by consensus. The transcribed texts, as similar to StoryDB narratives, were segmented into sentences using Trankit pipelines for subsequent analysis.

### 4.3 Surprisal estimation

#### 4.3.1 Frequency-based lexical surprisal

**Figure 1**A illustrates the pipeline. We estimated a frequency-based lexical surprisal using unigram relative frequencies with the python package wordfreq (version 3.1.1).[87] For each sentence $x = (t_1, \dots, t_{N(x)})$, we first obtained a word-level tokenization using the wordfreq tokenizer with all punctuations excluded, and the relative frequency of each token was derived from the package as the estimated context-free lexical probability $p_{\mathrm{freq}}(t)$.

As Basque is not supported by wordfreq, we estimated lexical frequency of Basque using a large external Basque Wikipedia corpus available on Hugging Face (orai-nlp/ZelaiaHandi). Specifically, we loaded the train split of this dataset (there is no other split), tokenized each text with the same tokenizer, aggregated corpus-wide token counts, and computed the relative frequencies of every unique token. Sentence-level frequency surprisal was computed as the sum of word-level surprisal:

$$S_{\text{freq}}(x) = -\sum_{i=1}^{N(x)} \log_2 p_{\text{freq}}\,(t_i) \tag{1}$$

To avoid infinite surprisal for out-of-vocabulary or zero-frequency items, we applied a small probability floor:

$$p_{\text{freq}}(t) \leftarrow max\big(p_{\text{freq}}(t), 10^{-9}\big) \tag{2}$$

4.3.2 Contextual surprisal derived from language models

To estimate contextual surprisal conditioned on rich sentence context, we used BERT family models to derive the probability of each token given the whole sentence as the context. For each of the 20 languages, we used a language-specific BERT-family model (for names and commit hash see Supplement). Given a sentence, we encoded it into subword tokens (including [CLS] and [SEP]) and excluded sentences whose encoded length exceeded 512 subword tokens. The probability of each subword token, $p_{\text{BERT}}\big(\,t_i \mid x_{\setminus i}\,\big)$, was computed by iteratively masking each subword token in the sequence (excluding special tokens), running the model's forward pass, and extracting the probability assigned to the true subword at the masked position. These probabilities were converted to subword surprisal $-\log_2 p_{\text{BERT}}\big(\,t_i \mid x_{\setminus i}\,\big)$. Zero probabilities were floored at $10^{-9}$. As the models operates over subwords while downstream analyses were conducted at the word level and excluded punctuation, we aligned the subword sequence (excluding [CLS]/[SEP]) to the punctuation-aware word tokenization from wordfreq using the python package spacy_alignments (version 0.9.1). Word-level contextual surprisal was then computed as the sum of subword surprisals aligned to each word, while skipping punctuation tokens (i.e., tokens present in the punctuation-aware list but absent from the punctuation-removed list). Sentence-level contextual surprisal was computed as the sum of word-level surprisal across the sentence.

$$S_{\text{BERT}}(x) = -\sum_{i=1}^{N(x)} \log_2 p_{\text{BERT}}\,\big(\,t_i \mid x_{\setminus i}\,\big) \tag{3}$$

where $\mathrm{x}_{\setminus \mathrm{i}}$ denotes the sentence with the token (or subword span) corresponding to $\mathrm{t}_\mathrm{i}$ masked. We additionally included results from generative language models, who model the probabilities of tokens only based on preceeding tokens, in the supplementary materials (Section 5) as a validation.

4.3.3 Word-order reversal control

To probe sensitivity to grammatical structure, we contrasted intact sentences with a word-order–distorted variant $x^{rev}$, obtained by reversing token order. This manipulation keeps the multiset of lexical items fixed but strongly perturbs grammatical sequencing, yielding a within-sentence comparison between intact and grammar-distorted inputs. We then generated a reversed-order variant computed $S_{\text{BERT}}(x^{rev})$ under the same model and preprocessing pipeline.

### 4.4 Syntactic analysis: dependency parsing and feature extraction

4.4.1 Dependency parsing

Sentences were parsed into dependency relations using the Trankit pipeline, as shown in **Figure 2**A. Token-level universal dependency annotations were obtained from the pipeline, including token text, token index, head index, and dependency relation labels. Each sentence was thus represented as a list of tuples $(t_i, i, h_i, rel_i)$, where $t_i$ is the token string, $i$ is the token index, $h_i$ is the head index ($h_i = 0$ indicating the root), and $rel_i$ is the dependency relation. As an example, we take the sentence *The mother forgot to turn off the water*. The sentence is parsed as: *[('The', 1, 2, 'det'), ('mother', 2, 3, 'nsubj'), ('forgot', 3, 0, 'root'), ('to', 4, 5, 'mark'), ('turn', 5, 3, 'xcomp'), ('off', 6, 5, 'compound:prt'), ('the', 7, 8, 'det'), ('water', 8, 5, 'obj')]*. Here, *forgot* (id=3) is the root; it takes *mother* (id=2) as its subject and *turn* (id=5) as an open clausal complement. The verb *turn* in turn governs *to* (marker), *off* (particle) and *water* (object), with determiners *the* attached to *mother* and *water*.

Before computing tree metrics, we removed punctuation attachments by pruning tokens with dependency relation (identified by the relation label *punct*), reattaching their children to the punct token's head to preserve connectivity. We then reindexed the remaining tokens consecutively (starting at 1) and ensured that exactly one node was treated as root (assigning a root if none was marked). From each pruned, reindexed dependency structure, we computed four syntactic descriptors capturing complementary aspects of grammatical organization (**Figure 2**A).

#### 4.4.2 Mean hierarchical distance: hierarchical complexity of the syntactic structure

To capture hierarchical complexity, we computed node depths from the root in the directed dependency graph (edges from heads to dependents). Let $\mathrm{depth}(\mathrm{v})$ denote the distance (in edges) from the root to node $\mathrm{v}$. We summarized hierarchical depth as the mean depth across non-root nodes:

$$\text{Mean hierarchical distance} = \frac{1}{|\mathrm{V}| - 1} \sum_{\mathrm{v} \in \mathrm{V} \setminus \{\mathrm{root}\}} \mathrm{depth}(\mathrm{v}) \tag{4}$$

In the example above, the root *forgot* has depth 0; its dependents *mother* and *turn* have depth 1; the dependents of those nodes (e.g., *The* under *mother*, and *to*, *off*, *water* under *turn*) have depth 2; and the determiner *the* attached to *water* has depth 3. Mean hierarchical distance increases when dependencies are distributed across more hierarchical levels rather than concentrated locally, indicating a more complex hierarchical syntactic structure underlying the superficial linear sequence.

#### 4.4.3 Dependency optimality: efficiency of linear packing

We quantified how efficiently a sentence packs dependencies into a linear order using a normalized optimality score $\Omega$ on dependency distance. Dependency distance quantifies the linear distance between a syntactic head and its dependent, and the total dependency length $D$ is the sum of these distances across all dependencies in a sentence:

$$D = \sum | h_i - i | \tag{5}$$

$D$ is commonly treated as a proxy for cognitive cost.[37] While languages show robust dependency length minimization relative to random baselines, dependency distances are typically not globally minimum.[88,89] Instead, natural language appears to reflect an efficiency trade-off: dependencies are kept sufficiently local to mitigate activation decay and interference, while remaining compatible with expectation-based and prediction-based constraints.[88,90] Ferrer-i-Cancho et al. therefore motivate an optimality measure to indicate such a balance, which rescales the observed $D$ relative to (a) the minimum possible total length $D_{\mathrm{min}}$ under an optimal linear arrangement of the same rooted tree and (b) a random-order baseline $D_{\mathrm{rand}}$ given by the expected sum of edge lengths under a uniformly random linearization. The optimality score Ω is thus defined as:

$$\Omega = \frac{D_{\text{rand}} - D}{D_{\text{rand}} - D_{min}} \tag{6}$$

Larger $\Omega$ indicates a more efficient linearization relative to the random baseline. In the example, $D_{\text{rand}}$ is computed as $(8^2 - 1)/3 = 21$, while the minimum baseline, as obtainable from the right tree of **Figure 2**A, is 10. Thus the $\Omega$ equals to 0.917. In this study, $\Omega$ was computed using the python tool Linear Arrangement Library (LAL, version 24.10).

#### 4.4.4 Subtree size unevenness

We quantified the heterogeneity of hierarchical organization via subtree size unevenness. In a dependency tree, for each node $v$, let $s(v)$ denote its subtree size, defined as the number of nodes dominated by $v$ including itself. We then formed a probability mass function over nodes,

$$p(v) = \frac{s(v)}{\sum_{u \in V} s(u)} \tag{7}$$

and computed unevenness as:

$$\text{Subtree size unevenness} = -\sum_{v \in V} p(v) \times \log_2 p\,(v) \tag{8}$$

In this example, the root *forgot* dominates the entire sentence (largest subtree), *turn* dominates a large verbal phrase (to *turn off the water*), *water* dominates a smaller noun phrase (including its determiner), and function words such as *to* and *off* are leaves (subtree size 1). Subtree size unevenness is high when the tree contains a rich mixture of small and large subtrees (as in this example, where both large verbal substructures and smaller nominal substructures coexist), and lower when most nodes have similar subtree sizes (as in near-linear chains or highly degenerate star-like trees).

#### 4.4.5 B2 index

Finally, we measured global topological balance using the rooted-tree balance index B2. After converting the dependency structure to a rooted tree representation, we computed B2 as implemented in the python package tskit (version 0.6.4). Higher B2 values indicate more balanced branching and thus less topological imbalance, whereas lower values indicate increasingly unbalanced, star- or chain-like configurations.

### 4.5 Intrinsic dimensionality estimation

To quantify the geometric organization of lexical meaning spaces, we estimated intrinsic dimensionality (ID) from context-free word embeddings using fastText models.[91] For each language, we used the corresponding pretrained fastText model trained on Common Crawl (https://fasttext.cc/docs/en/crawl-vectors.html), and constructed a text-level lexical semantic space. Within each text, we tokenized all sentences and, by default, removed the stopwords. Stopwords are defined in Stopwords ISO (version 0.6.1) for all languages except Tamil. Tamil stopwords were retrieved from a curated stopword list (https://gist.github.com/arulrajnet/e82a5a331f78a5cc9b6d372df13a919c). Tokens were lowercased and stripped and empty tokens were discarded. To focus on lexical diversity rather than raw repetition, we retained the set of unique non-stopword tokens within each passage. Each retained token was then encoded to a 300-dimensional fastText vector, yielding an embedding matrix

$$X \in R^{N(x_{\text{unique}}) \times D}, D = 300 \tag{9}$$

where $N(x_{\text{unique}})$ is the number of unique tokens in the passage and each row corresponds to one token embedding, after removing tokens whose fastText representations were all zeros. The embedding matrix was then z-scored.

ID quantifies the effective number of degrees of freedom underlying a high-dimensional representation, as needed to locally describe the data geometry. In panel 2B(a), the points largely fall along a single trend: although the data live in 2D, they behave as if they vary along one single underlying direction, so $\text{ID} \approx 1$. In panel 2B(b), the points spread out over an area rather than a line, so describing their variability requires two underlying directions, and thus $\text{ID} \approx 2$. Panel (c) provides an intuitive way to see this. If we zoom out (increase the radius $r$) and count how many neighbors fall within that radius, the count grows more slowly for a line-like structure than for an area-like structure. Concretely, for a roughly d-dimensional structure, the number of points within radius $r$ increases approximately like $r^d$. Therefore, on the log-log plot, the slope reflects the effective dimensionality. In other words, a slope near 1 indicates line-like (one-dimensional) variation, whereas a slope near 2 indicates area-like (two-dimensional) variation.

We estimated ID on two complementary manifolds, as seen in **Figure 2**B-d. In the token manifold, ID was estimated across token embeddings, capturing how many effective degrees of freedom were needed to describe the geometric spread of lexical items within a text. In the feature manifold, ID was estimated across embedding dimensions, capturing how many latent directions were needed to account for variation in the semantic representation. As token-manifold estimates can depend on the number of sampled tokens, token-manifold ID was then divided by $N$ for normalization. In our analyses, ID therefore provides an intuitive measure of the effectiveness and compressibility of the lexical semantic space: higher token-manifold ID indicates less redundancy and greater diversity across lexical items within a text, while lower feature-manifold ID indicates more compressible semantic space.

Generalized ratios intrinsic dimension estimator (GRIDE) was employed as the main estimator for ID, coupled with the kstar estimation of the neighborhood scale.[92] GRIDE algorithm was implemented in the python package DADApy (version 0.3.3). This algorithm employs minimal assumptions and has been previously applied on language embeddings, assuming that the data lie on a single manifold.[93] As a robustness check, we also computed the ID using maximum-likelihood (MLE) estimator. MLE a more canonical approach that infers ID from the distribution of local neighbor distances under a Poisson-process approximation.[94] In MLE, the number of nearest neighbors is set as 15 following suggestions from sensitivity analysis in the previous work[95]. MLE was implemented in the python package scikit-dimension (skdim, version 0.3.4).

### 4.6 Text-level feature extraction and summary

All analyses were conducted at the text level to enable cross-lingual comparisons under a common unit of observation. For each text $T = \left(x_1, \dots, x_{n(T)}\right)$, where each sentence represented as a token sequence $x = \left(t_1, \dots, t_{N(x)}\right)$, we first computed surprisal values and syntactic measures at the sentence level. ID values, as noted before, were computed at the text level. Syntactic measures were averaged across sentences. Number of tokens, as well as the three surprisal values, $S_{\text{freq}}(x)$, $S_{\text{BERT}}(x)$, and $S_{\text{BERT}}(x^{rev})$, were summed across sentences within the text. Then, we normalized the text-level surprisal (i.e. the sum of the sentence-level surprisal) by total number of words of the text $N(T)$, thus we have:

$$\overline{T_{\text{freq}}}(x) = \frac{1}{N(T)} \sum_{i=1}^{n(T)} S_{\text{freq}}(x) \tag{10}$$

$$\overline{T_{\text{BERT}}}(x) = \frac{1}{N(T)} \sum_{i=1}^{n(T)} S_{\text{BERT}}(x) \tag{11}$$

$$\overline{T_{\text{BERT}}}(x^{\text{rev}}) = \frac{1}{N(T)} \sum_{i=1}^{n(T)} S_{\text{BERT}}(x^{\text{rev}}) \tag{12}$$

To quantify the reduction of uncertainty from grammar-insensitive to grammar-sensitive conditions, we derived a normalized surprisal-reduction index contrasting the lexical with contextual surprisal ($\Delta Diff_{\text{FB}}(x)$), and an analogous index contrasting the reversed-order (grammar-distorted) with the intact sentences ($\Delta Diff_{\text{REV}}(x)$).

$$\Delta Diff_{\text{FB}}(x) = \frac{\overline{T_{\text{freq}}}(x) - \overline{T_{\text{BERT}}}(x)}{\overline{T_{\text{freq}}}(x)} \tag{13}$$

$$\Delta Diff_{\text{REV}}(x) = \frac{\overline{T_{\text{BERT}}}(x^{\text{rev}}) - \overline{T_{\text{BERT}}}(x)}{\overline{T_{\text{BERT}}}(x^{\text{rev}})} \tag{14}$$

### 4.7 Statistical analyses

#### 4.7.1 Cross-lingual comparisons of surprisal values under three conditions

To evaluate whether surprisal estimates differed systematically across modeling conditions and whether such patterns were universal across different languages, we compared the differences among $\overline{T_{\text{freq}}}(x)$, $\overline{T_{\text{BERT}}}(x)$, and $\overline{T_{\text{BERT}}}(x^{\text{rev}})$, using Friedman test (Pingouin, version 0.5.5) followed by post hoc pairwise comparisons using the Siegel-Friedman procedure (scikit_posthocs, version 0.11.4). *P*-values from the pairwise comparisons were corrected using the false discovery rate (FDR)).

#### 4.7.2 Associations between uncertainty measures and linguistic features

To quantify how surprisal reduction relates to structural and semantic properties of the texts, we estimated the Spearman rank correlations between $\Delta Diff_{\text{FB}}(x)$ and the six linguistic features described above. Correlations were computed using a pairwise approach within each language, with *p* values corrected using FDR separately for each feature across languages (i.e., for a given linguistic feature, *p* values from all languages were jointly adjusted). In the supplementary materials, we reported the Spearman rank correlations between the surprisal ($\overline{T_{\text{freq}}}(x)$ and $\overline{T_{\text{BERT}}}(x)$) and the six linguistic features, and also the Spearman rank correlations between the surprisal reduction and ID values estimated using the MLE algorithm. Additionally, we correlated the two surprisal difference values, $\Delta Diff_{\text{FB}}(x)$ and $\Delta Diff_{\text{REV}}(x)$, similarly using Spearman rank correlations and meta-analysis.

For each feature, we summarized the correlation strength (i.e. Spearman's $\rho$) across 20 languages using meta-analytic tools. Specifically, we fitted a random-effects meta-analysis using restricted maximum likelihood (REML) estimation for each feature, treating each language as an independent study contributing one effect size. We then derived the pooled random-effects estimate (total effect) with its 95% confidence interval (CI) from the model. REML was implemented in R 4.5.1 using the meta package (version 8.2-1). To evaluate cross-language heterogeneity, we computed Cochran's Q statistic and its p-value for each feature. Total effect with its 95% CI was reported in the results section, and also in the supplementary together with the Cochran heterogeneity test results and forest plots.

#### 4.7.3 Regression on fMRI BOLD signals

To regress linguistic surprisal and its reduction on BOLD signals, we first computed frequency-based surprisal, BERT-based surprisal, and their proportional differences per word. We also counted the number of syllables per word to account for low-level speech effect. Syllable counts were derived from the Python package syllapy (version 0.7.2). As uncertainty estimates were defined at the word level but fMRI signals were sampled at discrete scan times, we carried out temporal alignment between BOLD signals and linguistic variables (linguistic surprisal, surprisal reduction, and syllable counts). Specifically, we used the word offsets as event times and convolved the resulting event-wise feature series with a canonical hemodynamic response function (HRF) at TR=1.5 s, which yielded scan-aligned regressors for every word-derived feature. The scan-aligned regressors were all standardized.

After temporal alignment, we fitted linear mixed-effects models to regress the linguistic features on the time series separately for each of the 416 regions of interest (ROIs), including 400 cortical ROIs and 16 subcortical ROIs. The dependent variable was the ROI-specific fMRI time series (scan-level), and predictors were the HRF-convolved scan-level regressors. For each ROI, we estimated four models using the Python package statsmodels (version 0.14.0), each focusing on a single uncertainty-related regressor while controlling for a low-level speech covariate:

$$\mathrm{ROI}_{\mathrm{r,t}} = \beta_0 + \beta_1 X_t + \beta_2\, \mathrm{SyllabelNum_t} + b_{0,\mathrm{subj}(t)} + \varepsilon_{r,t} \tag{15}$$

where $\mathrm{ROI}_{\mathrm{r,t}}$ denotes the fMRI signal in ROI $r$ at scan $t$, $X_t$ is one of the HRF-convolved surprisal regressors (frequency-based surprisal, BERT-based surprisal, or their proportional differences), and $\mathrm{SyllabelNum_t}$ is the syllable count regressor to control for low-level speech effect. Models included a random intercept $b_{0,\mathrm{subj}(t)}$ for subjects. Models were fitted using restricted maximum likelihood (REML) optimization with the L-BFGS algorithm. To control for multiple comparisons across ROIs, we applied FDR correction across the 416 ROIs for each of the three surprisal regressors.

#### 4.7.4 Clinical cohort analyses of surprisal metrics: group effects and symptom correlations

Despite cohort differences, statistical comparisons for the nine cohort followed a common pipeline. We started from sentence-level outputs containing word-level uncertainty measures. Sentences shorter than three tokens were excluded. We computed the word count and surprisal metrics per participant. If participants completed more than one task, we averaged the measures across tasks to derive subject-level scores. Firstly, we compared the three surprisal metrics using Friedman test followed by Siegel-Friedman pairwise comparisons, similarly to the comparisons using StoryDB data. Results were reported in the supplementary materials. Secondly, to quantify diagnostic group effects on uncertainty measures while accounting for repeated task observations per participant, we fit generalized estimating equations (GEE) separately for each outcome at the task level, if participants completed more than one task; or else, generalized linear models (GLM) for each outcome at the participant level. To be consistent across all cohorts, the models included only basic demographics (age and sex) and word count as the confounds, and diagnostic group as the predictors with healthy or typical populations as the reference category. GEE/GLM models used a Gaussian family to fit the data, and the goodness of fit was checked via a deviance-based goodness-of-fit test. If the fitness was not good, the models would change to Gamma distribution instead. Finally, all models fit the data well. *P* values were adjusted for multiple comparisons using two-step FDR correction. Finally, we correlated surprisal metrics to clinical symptom and cognitive scale scores within the patient groups, using Spearman rank correlations. GEE/GLM models were carried out with statsmodels and the correlation analyses were implemented in pingouin.

### Data availability

The StoryDB data is publicly available at https://drive.google.com/drive/folders/1RCWk7pyvIpubtsf-f2pIsfqTkvtV80Yv. The fMRI data are available at https://openneuro.org/datasets/ds001132/versions/1.0.0 ad https://openneuro.org/datasets/ds001110/versions/00003. A cleaned and annotated version of transcripts can be found at https://osf.io/qjmky/overview. The Olness aphasia corpus, Minga RHD corpus, Dutch Asymmetries corpus, and ADReSS challenge data are available in the clinical banks of TalkBank (https://talkbank.org/). The English PPA dataset is available at https://osf.io/sr3ag/. The TOPSY dataset are made available to qualified researchers through https://talkbank.org/psychosis/, a collaboration between DISCOURSE in Psychosis consortium (https://discourseinpsychosis.org/) and TalkBank. The ACE dataset, Zhuhai dataset, and Marburg dataset cannot be made publicly available due to ethical restrictions, but are available upon reasonable request.

### Code availability

Scripts and results are available at https://github.com/RuiHe1999/grammar_uncertainty and https://doi.org/10.17605/OSF.IO/HFRMD.

**Tables**

**Table 1. Descriptions of the StoryDB dataset.** [a]

| Languages | | | Text No. | | Averaged sentence count | Averaged word count | Averaged sentence length |
|---|---|---|---|---|---|---|---|
| Names | Codes [b] | Families and branches [c] | Before filtering | After filtering | | | |
| Arabic | ar | Afro-Asiatic | 3358 | 16.92 | 16.92 | 330.05 | 20.27 |
| Catalan | ca | Indo-European, Italic-Romance | 738 | 13.83 | 13.83 | 313.01 | 22.61 |
| Czech | cs | Indo-European, Balto-Slavic (Slavic) | 996 | 23.25 | 23.25 | 426.96 | 20.22 |
| German | de | Indo-European, Germanic | 23741 | 22.57 | 22.57 | 450.72 | 21.81 |
| Greek | el | Indo-European, Greek | 2362 | 21.08 | 21.08 | 439.29 | 21.64 |
| English | en | Indo-European, Germanic | 59172 | 21.44 | 21.44 | 437.45 | 21.75 |
| Spanish | es | Indo-European, Italic-Romance | 10167 | 22.59 | 22.59 | 496.90 | 22.04 |
| Basque | eu | Language isolate | 4863 | 13.40 | 13.40 | 228.65 | 21.63 |
| Persian | fa | Indo-European, Indo-Iranian | 4761 | 20.34 | 20.34 | 439.92 | 21.92 |
| French | fr | Indo-European, Italic-Romance | 30655 | 21.63 | 21.63 | 460.91 | 22.13 |
| Hungarian | hu | Uralic | 2325 | 24.78 | 24.78 | 452.91 | 21.22 |
| Indonesian | id | Austronesian | 3183 | 20.62 | 20.62 | 435.41 | 22.19 |
| Japanese | ja | Japonic | 7532 | 21.76 | 21.76 | 513.66 | 23.02 |
| Korean | ko | Koreanic | 2866 | 20.03 | 20.03 | 469.88 | 22.25 |
| Dutch | nl | Indo-European, Germanic | 14719 | 22.86 | 22.86 | 422.47 | 21.41 |
| Polish | pl | Indo-European, Balto-Slavic (Slavic) | 11799 | 20.49 | 20.49 | 346.69 | 21.81 |
| Tamil | ta | Dravidian | 1125 | 19.18 | 19.18 | 280.28 | 17.53 |
| Turkish | tr | Turkic | 185 | 19.49 | 19.49 | 276.47 | 15.33 |
| Vietnamese | vi | Austro-Asiatic | 2856 | 21.44 | 21.44 | 460.90 | 21.80 |
| Chinese | zh | Sino-Tibetan | 2261 | 20.21 | 20.21 | 451.95 | 22.41 |

[a] Available at https://drive.google.com/drive/folders/1RCWk7pyvIpubtsf-f2pIsfqTkvtV80Yv.

[b] Language code as in the ISO 639-1 standard.

[c] Language family information as in the 28 edition of *Ethnologue: Languages of the World*.[96]

## Figures

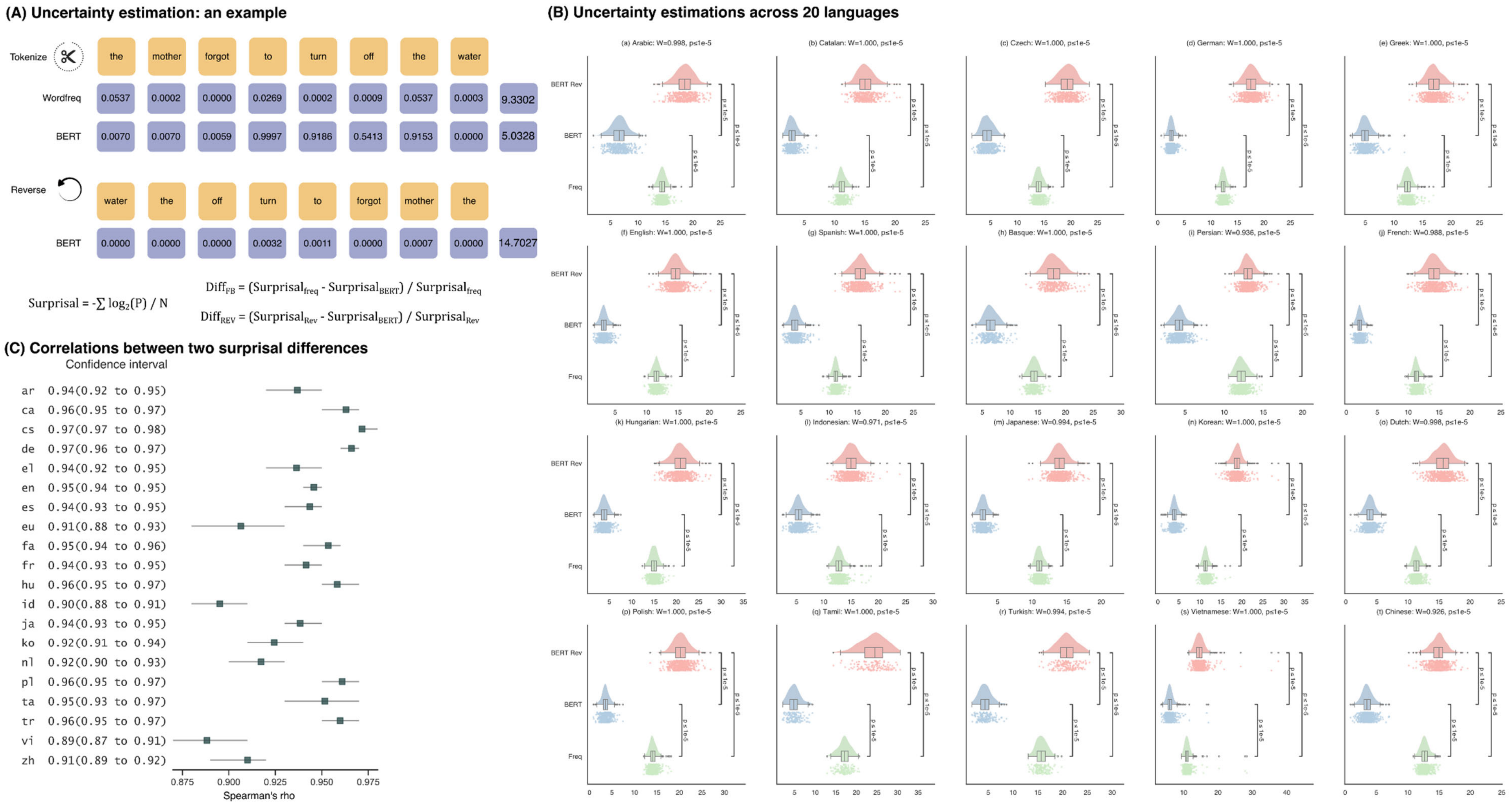

**Figure 1. Cross-linguistic uncertainty reduction from lexical to contextual surprisal.** (A) Example of uncertainty estimation. Each word is assigned a non-contextual surprisal based on its lexical frequency (WordFreq) and a contextual surprisal from a grammar-sensitive language model (BERT). A third row shows surprisal for the same words presented in reversed order, which preserves the multiset of words but disrupts grammatical structure. Sentence-level surprisal corresponds to the average across words. (B) Uncertainty estimation across 20 languages (a–t). For each language, distributions of sentence-level surprisal are shown for frequency-based estimates (Freq, green), contextual surprisal in the original order (BERT, blue), and contextual surprisal for reversed sentences (BERT Rev, red). Violin plots with overlaid points represent individual texts. Differences across conditions were tested using Friedman's test (reported in each panel title) with Siegel's post-hoc pairwise comparisons (reported in the brackets). Across all languages, contextual surprisal is lower than frequency-based surprisal, and reversing word order reliably increases contextual surprisal. (C) Forest plot of the correlations between uncertainty reduction and the surprisal cost of reversal. For each language, normalized uncertainty reduction (proportional reduction from frequency-based to contextual surprisal) is correlated with the increase in contextual surprisal induced by reversal (Spearman's $\rho$ with 95% confidence intervals). Squares indicate the meta-analytic Spearman's $\rho$ across languages, showing a strong positive relationship between the benefit of grammatical context and its disruption by word-order scrambling.

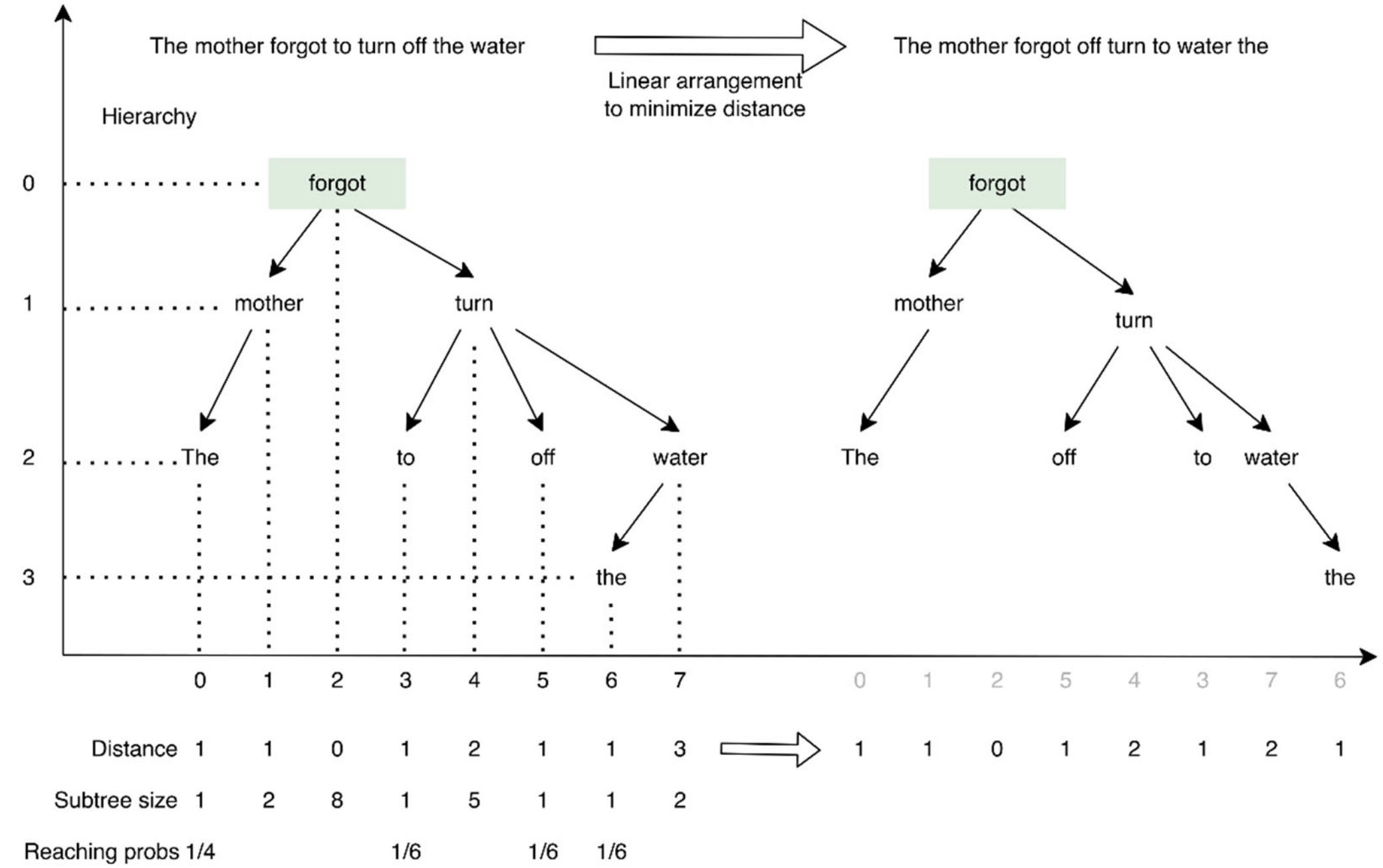
(A) Syntax parsing: dependency distance, and topological imbalance
The mother forgot to turn off the water
Linear arrangement
to minimize distance
The mother forgot off turn to water the
Hierarchy
forgot
mother
turn
The
to
off
water
the
Distance 1 1 0 1 2 1 1 3
1 1 0 1 2 1 2 1
Subtree size 1 2 8 1 5 1 1 2
Reaching probs 1/4 1/6 1/6 1/6

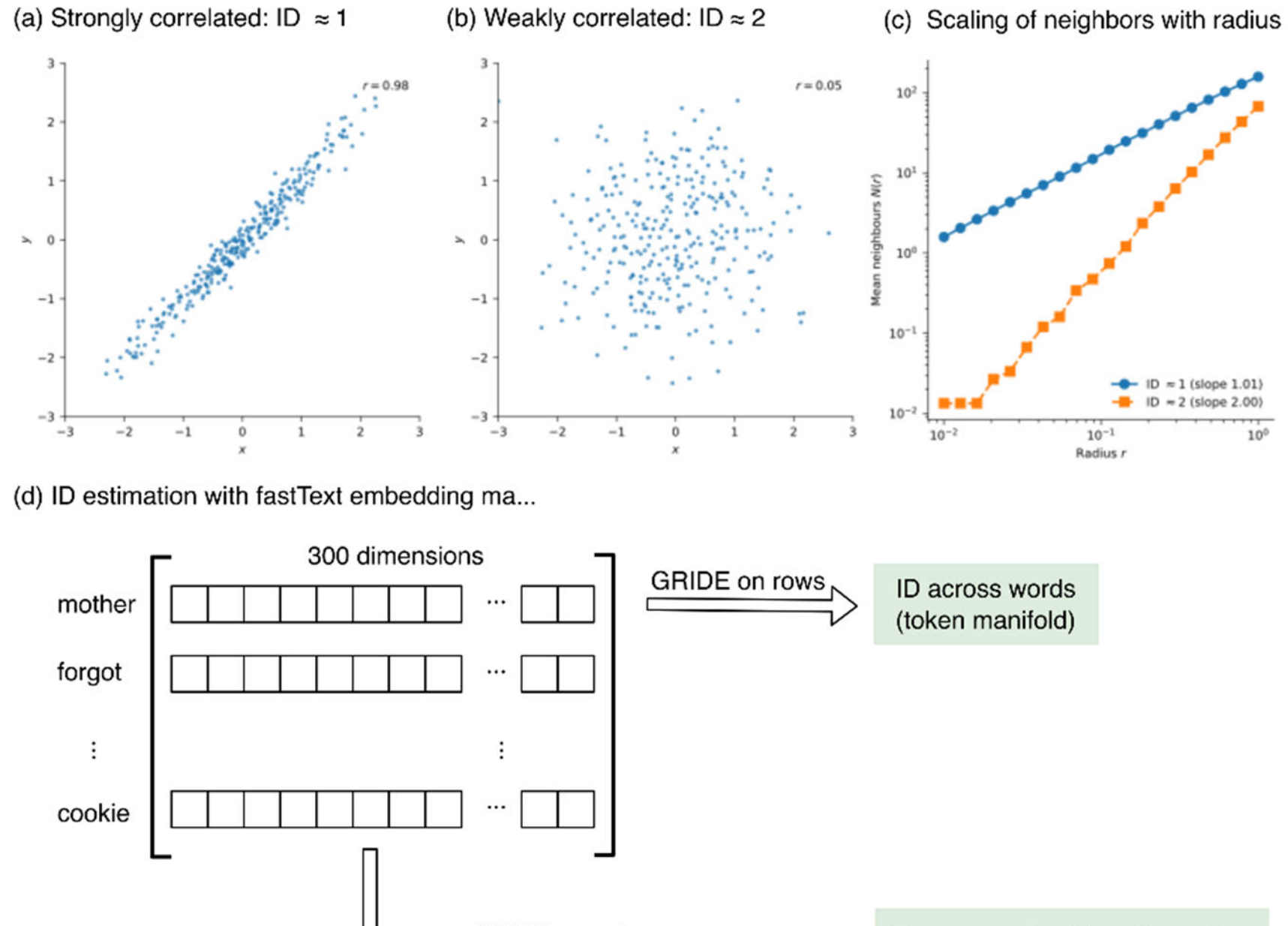
(B) Lexical semantic space: intrinsic dimensionality (ID)
(a) Strongly correlated: ID ≈ 1
(b) Weakly correlated: ID ≈ 2
(c) Scaling of neighbors with radius
ID = 1 (slope 1.01)
ID = 2 (slope 2.00)
Radius r
Mean neighbours N(r)
(d) ID estimation with fastText embedding ma...
300 dimensions
mother
forgot
cookie
GRIDE on rows
ID across words
(token manifold)
GRIDE on columns
ID across embedding dimensions
(feature manifold)

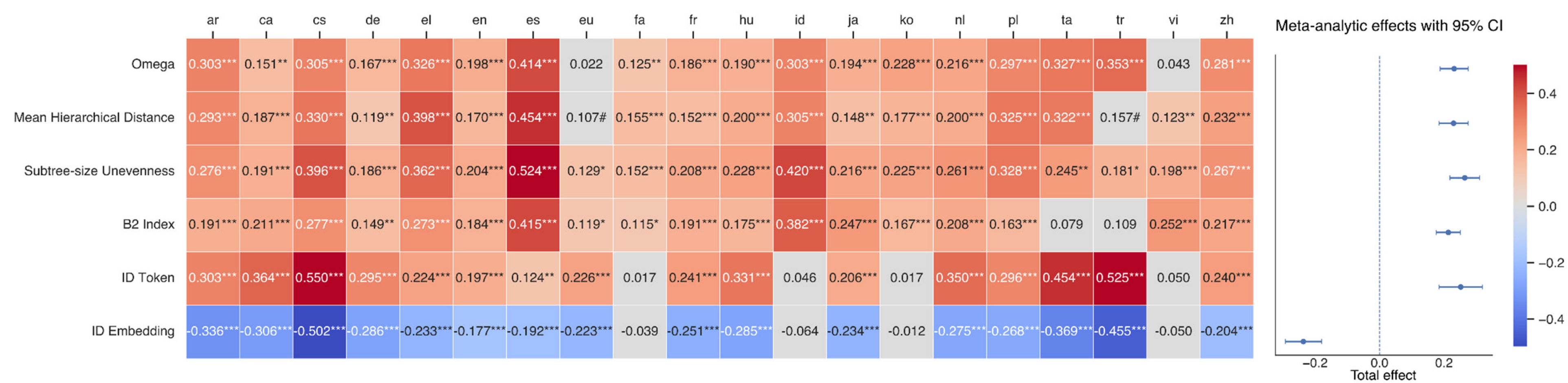
(C) Correlations between uncertainty reduction, syntactic structure, and intrinsic dimensions of meaning
ar ca cs de el en es eu fa fr hu id ja ko nl pl ta tr vi zh
Omega 0.303*** 0.151** 0.305*** 0.167*** 0.326*** 0.198*** 0.414*** 0.022 0.125** 0.186*** 0.190*** 0.303*** 0.194*** 0.228*** 0.216*** 0.297*** 0.327*** 0.353*** 0.043 0.281***
Mean Hierarchical Distance 0.293*** 0.187*** 0.330*** 0.119** 0.398*** 0.170*** 0.454*** 0.107# 0.155*** 0.152*** 0.200*** 0.305*** 0.148** 0.177*** 0.200*** 0.325*** 0.322*** 0.157# 0.123** 0.232***
Subtree-size Unevenness 0.276*** 0.191*** 0.396*** 0.186*** 0.362*** 0.204*** 0.524*** 0.129* 0.152*** 0.208*** 0.228*** 0.420*** 0.216*** 0.225*** 0.261*** 0.328*** 0.245** 0.181* 0.198*** 0.267***
B2 Index 0.191*** 0.211*** 0.277*** 0.149** 0.273*** 0.184*** 0.415*** 0.119* 0.115* 0.191*** 0.175*** 0.382*** 0.247*** 0.167*** 0.208*** 0.163*** 0.079 0.109 0.252*** 0.217***
ID Token 0.303*** 0.364*** 0.550*** 0.295*** 0.224*** 0.197*** 0.124** 0.226*** 0.017 0.241*** 0.331*** 0.046 0.206*** 0.017 0.350*** 0.296*** 0.454*** 0.525*** 0.050 0.240***
ID Embedding -0.336*** -0.306*** -0.502*** -0.286*** -0.233*** -0.177*** -0.192*** -0.223*** -0.039 -0.251*** -0.285*** -0.064 -0.234*** -0.012 -0.275*** -0.268*** -0.369*** -0.455*** -0.050 -0.204***
Meta-analytic effects with 95% CI
Total effect

**Figure 2. Formal syntactic organization and lexical semantic space jointly shape uncertainty reduction.** (A) Illustration of dependency-based metrics. Sentences are represented as trees of head–dependent relations. The numbers on the horizontal axis mark the linear positions of the words in the sentence. Dependency distance is defined as the head–dependent distances for each word (number of intervening words between a head and its dependent), whose sum is optimized to deviate from both the minimal value as found by linear arrangement (on the right), and the maximal value where dependent words are as far away from each other as possible. Subtree size gives the number of nodes in the subtree rooted at each word (for example, the subtree of *mother* contains two words). Reaching probabilities shows the probability of reaching each node in the tree under a simple traversal scheme, which is used to compute B2 index as a summary of topological imbalance. (B) Illustration of intrinsic dimensionality (ID) in lexical semantic space. Subpanels (a–b) illustrate that two variables can lie in the same 2D space but differ in their *effective* degrees of freedom: when they are nearly colinear, the data are confined to an approximately one-dimensional manifold (ID ≈ 1), whereas weakly correlated variables span a two-dimensional region (ID ≈ 2). Subpanel (c) shows the corresponding scaling of the mean number of neighbors within radius $r$, $N(r)$, on a log–log scale. The slope of this relationship provides an estimate of the ID. Subpanel (d) indicates how the IDs of the non-stopword token embedding matrices from fastText ID are estimated, either across tokens (token manifold) or across embedding dimensions (feature manifold). (C) Heatmap of Spearman correlations between normalized surprisal reduction and each structural or semantic metric across 20 languages (columns). Colored cells indicate significant correlations between uncertainty reduction and linguistic variables (rows) in a certain language data (columns). Warm colors indicate positive correlations, cool colors negative correlations, and asterisks mark FDR-corrected significance levels. * $p<0.05$, ** $p<0.01$, *** $p<0.001$. The forest plot on the right shows meta-analytic correlation coefficients for each metric with 95% confidence intervals.

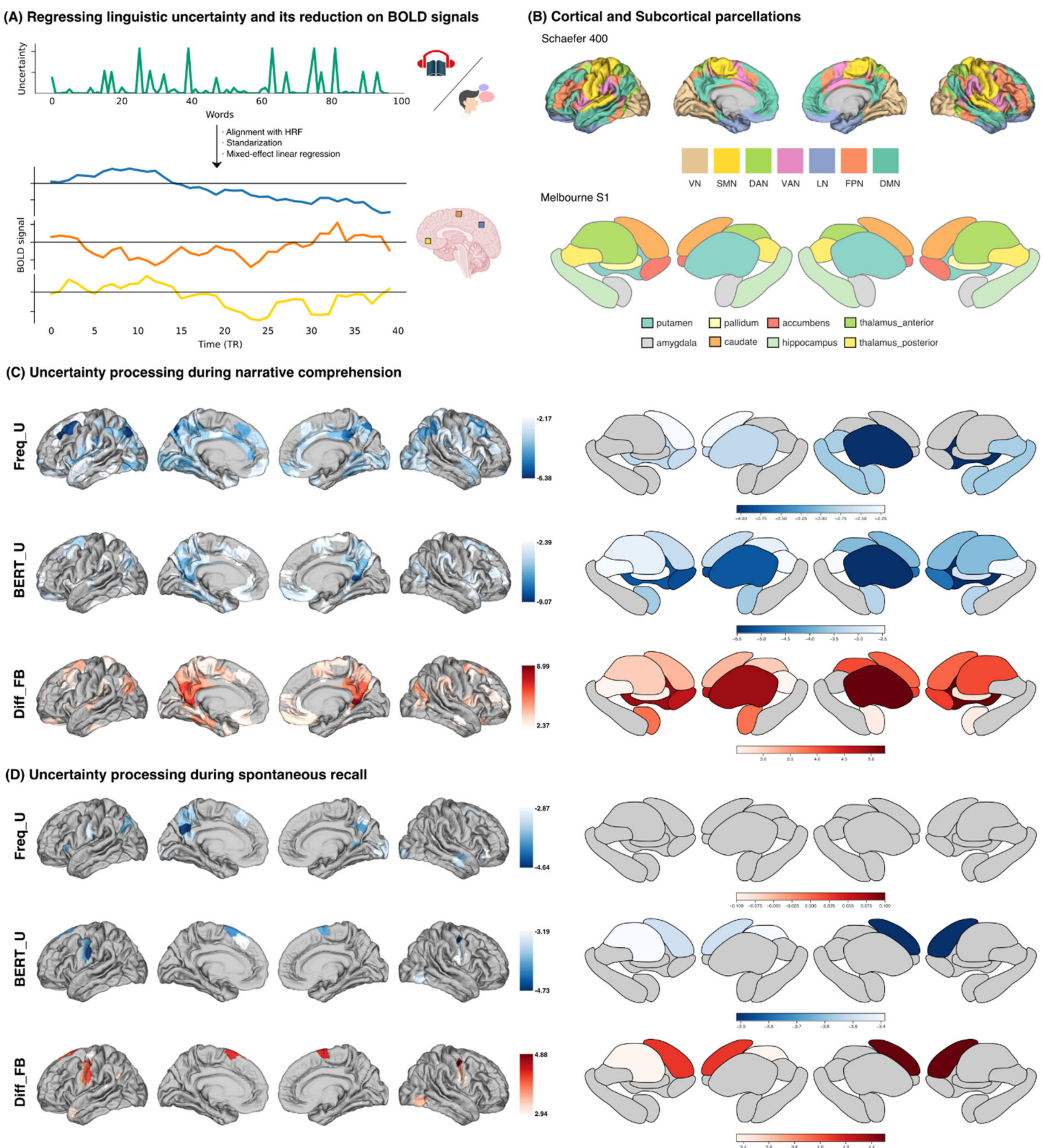

**Figure 3. Neural correlates of linguistic uncertainty and its reduction.** (A) Schematic of the modelling approach. Word-by-word estimates of linguistic uncertainty were aligned to fMRI time series using a canonical hemodynamic response function (HRF), standardised, and entered as predictors in mixed-effects linear regression of BOLD signals for every region of interest. (B) Cortical and subcortical parcellations. Cortical parcels follow the Schaefer-400 atlas, grouped into seven large-scale networks (visual, somatomotor, dorsal attention, ventral attention, limbic, frontoparietal and default-mode networks). Subcortical regions follow the Melbourne S1 parcellation, including striatal and thalamic subregions, hippocampus and amygdala. (C) Uncertainty processing during narrative comprehension (Sherlock audio). Cortical (left) and subcortical (right) plots show regression coefficients for frequency-based surprisal (Freq_U), contextual surprisal from BERT (BERT_U) and their normalized difference (Diff_FB, indexing uncertainty reduction). Brain regions with statistically significant effects were colored. Blue colors indicate negative associations (higher uncertainty with lower BOLD), red colors positive associations. (D) Uncertainty processing during spontaneous recall (Sherlock recall). Conventions as in (C).

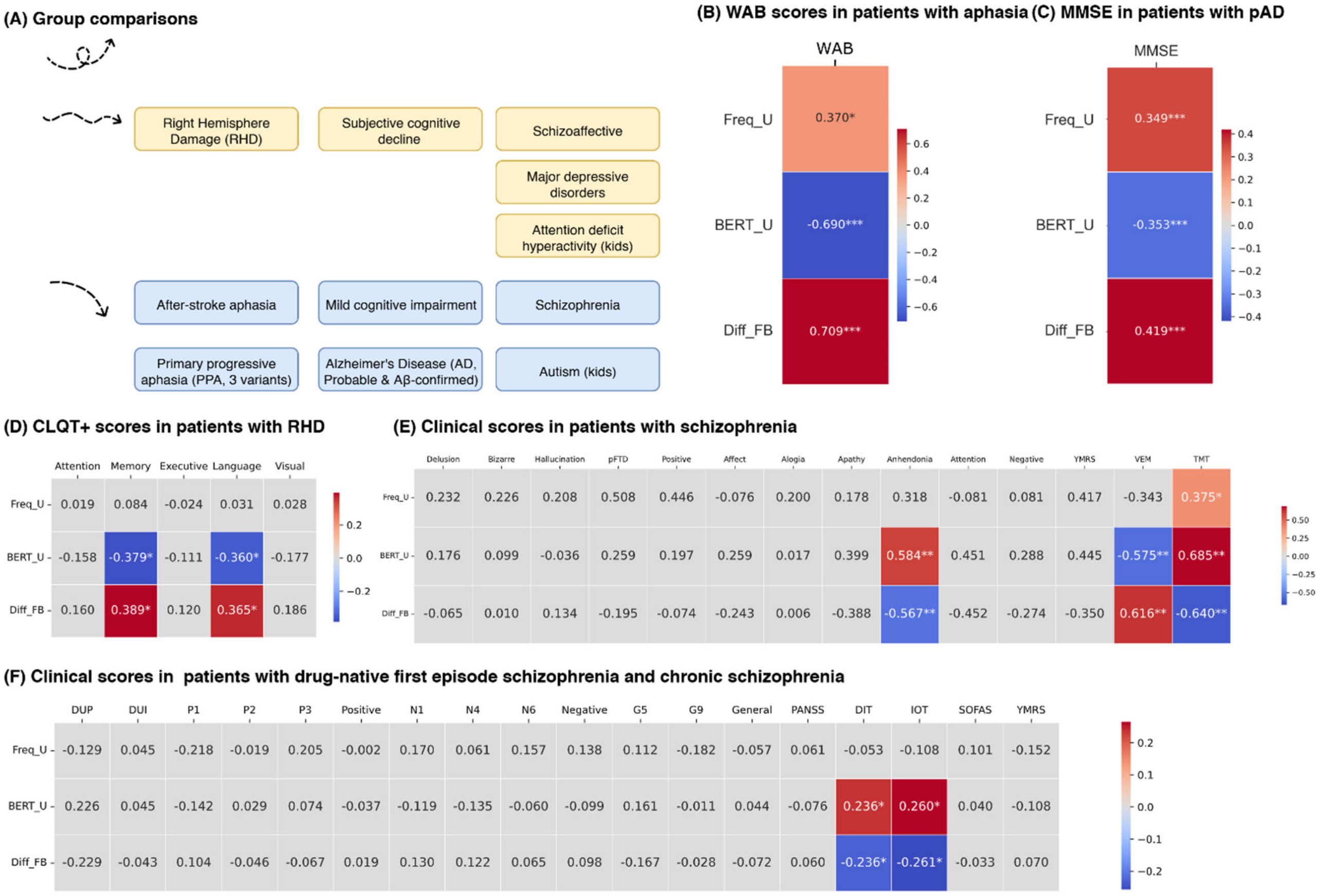

**Figure 4. Selective disruption of uncertainty optimization in clinical populations.** Panel A indicated the direction of changes in surprisal reduction in different disorders. In panels B–F, each cell shows a Spearman's correlation coefficient between an uncertainty measure and a behavioral or clinical variable; numbers indicate the correlation value, and color encodes its sign and magnitude. Colored cells indicate significant correlations of three surprisal values (Freq_U: frequency-based surprisal, BERT_U: BERT-based surprisal, and Diff_FB: their differences) to clinical and neuropsychological scores. Warm colors indicate positive correlations, cool colors negative correlations, and asterisks mark FDR-corrected significance levels. * $p$<0.05, ** $p$<0.01, *** $p$<0.001. (A) Overview of group comparisons of uncertainty reduction in the nine clinical cohorts. No clinical group exhibited significantly larger uncertainty reduction. Significantly lower uncertainty reduction was observed in groups marked by the down arrow in blue boxes. Insignificant changes were observed in groups marked by the horizontal wave arrows in yellow boxes. (B) Correlations between uncertainty measures and Western Aphasia Battery (WAB) scores in patients with post-stroke aphasia. (C) Correlations between uncertainty measures and Mini-Mental State Examination (MMSE) scores in patients with probable Alzheimer's disease (pAD). (D) Correlations between uncertainty measures and Cognitive Linguistic Quick Test Plus (CLQT+) scores in patients with right hemisphere damage (RHD). Significant correlations were observed with the memory (M) and language (L) domain scores, and not in Attention (Att), executive function (EF), and visual (V) domains. (E) Correlations between uncertainty measures and clinical scores in patients with schizophrenia in the Marburg cohort. Significant correlations were observed with anhedonia, verbal episodic memory (VEM), and executive function as examined with trail-making tests (TMT). Full terms of abbreviations available in Supplementary Table 9. (F) Correlations between uncertainty measures and clinical scores in patients with schizophrenia, in both very acute stage of the drug-naïve first episode and relatively stable stage. Significant correlations were observed with thought disorganization (DIT) and impoverishment (IOT). Full terms of abbreviations available in Supplementary Table 8.

## (A) Uncertainty estimation: an example

| | | | | | | | | | |
|---|---|---|---|---|---|---|---|---|---|
| Tokenize | the | mother | forgot | to | turn | off | the | water | |
| Wordfreq | 0.0537 | 0.0002 | 0.0000 | 0.0269 | 0.0002 | 0.0009 | 0.0537 | 0.0003 | 9.3302 |
| BERT | 0.0070 | 0.0070 | 0.0059 | 0.9997 | 0.9186 | 0.5413 | 0.9153 | 0.0000 | 5.0328 |
| Reverse | water | the | off | turn | to | forgot | mother | the | |
| BERT | 0.0000 | 0.0000 | 0.0000 | 0.0032 | 0.0011 | 0.0000 | 0.0007 | 0.0000 | 14.7027 |

$$\text{Surprisal} = -\sum \log_2(P) / N$$

$$\text{Diff}_{FB} = (\text{Surprisal}_{freq} - \text{Surprisal}_{BERT}) / \text{Surprisal}_{freq}$$

$$\text{Diff}_{REV} = (\text{Surprisal}_{Rev} - \text{Surprisal}_{BERT}) / \text{Surprisal}_{Rev}$$

## (B) Uncertainty estimations across 20 languages

(a) Arabic: W=0.998, p≤1e-5
(b) Catalan: W=1.000, p≤1e-5
(c) Czech: W=1.000, p≤1e-5
(d) German: W=1.000, p≤1e-5
(e) Greek: W=1.000, p≤1e-5
(f) English: W=1.000, p≤1e-5
(g) Spanish: W=1.000, p≤1e-5
(h) Basque: W=1.000, p≤1e-5
(i) Persian: W=0.936, p≤1e-5
(j) French: W=0.988, p≤1e-5
(k) Hungarian: W=1.000, p≤1e-5
(l) Indonesian: W=0.971, p≤1e-5
(m) Japanese: W=0.994, p≤1e-5
(n) Korean: W=1.000, p≤1e-5
(o) Dutch: W=0.998, p≤1e-5
(p) Polish: W=1.000, p≤1e-5
(q) Tamil: W=1.000, p≤1e-5
(r) Turkish: W=0.994, p≤1e-5
(s) Vietnamese: W=1.000, p≤1e-5
(t) Chinese: W=0.926, p≤1e-5

BERT Rev, BERT, Freq; p ≤ 1e-5

## (C) Correlations between two surprisal differences

| | Confidence interval |
|---|---|
| ar | 0.94(0.92 to 0.95) |
| ca | 0.96(0.95 to 0.97) |
| cs | 0.97(0.97 to 0.98) |
| de | 0.97(0.96 to 0.97) |
| el | 0.94(0.92 to 0.95) |
| en | 0.95(0.94 to 0.95) |
| es | 0.94(0.93 to 0.95) |
| eu | 0.91(0.88 to 0.93) |
| fa | 0.95(0.94 to 0.96) |
| fr | 0.94(0.93 to 0.95) |
| hu | 0.96(0.95 to 0.97) |
| id | 0.90(0.88 to 0.91) |
| ja | 0.94(0.93 to 0.95) |
| ko | 0.92(0.91 to 0.94) |
| nl | 0.92(0.90 to 0.93) |
| pl | 0.96(0.95 to 0.97) |
| ta | 0.95(0.93 to 0.97) |
| tr | 0.96(0.95 to 0.97) |
| vi | 0.89(0.87 to 0.91) |
| zh | 0.91(0.89 to 0.92) |

Spearman's rho

**(A) Syntax parsing: dependency distance, and topological imbalance**

The mother forgot to turn off the water

Linear arrangement to minimize distance

The mother forgot off turn to water the

Hierarchy

0 forgot
1 mother turn
2 The to off water
3 the

forgot
mother turn
The off to water
the

0 1 2 3 4 5 6 7

0 1 2 5 4 3 7 6

| | | | | | | | | | | | | | | | | |
|---|---|---|---|---|---|---|---|---|---|---|---|---|---|---|---|---|
| Distance | 1 | 1 | 0 | 1 | 2 | 1 | 1 | 3 | ⇒ | 1 | 1 | 0 | 1 | 2 | 1 | 2 | 1 |
| Subtree size | 1 | 2 | 8 | 1 | 5 | 1 | 1 | 2 | | | | | | | | | |
| Reaching probs | 1/4 | | | 1/6 | | 1/6 | 1/6 | | | | | | | | | | |

**(B) Lexical semantic space: intrinsic dimensionality (ID)**

(a) Strongly correlated: ID ≈ 1

$r = 0.98$

(b) Weakly correlated: ID ≈ 2

$r = 0.05$

(c) Scaling of neighbors with radius

Mean neighbours $N(r)$

Radius $r$

ID ≈ 1 (slope 1.01)

ID ≈ 2 (slope 2.00)

(d) ID estimation with fastText embedding ma...

300 dimensions

mother

forgot

⋮

cookie

GRIDE on rows → ID across words (token manifold)

GRIDE on columns → ID across embedding dimensions (feature manifold)

**(C) Correlations between uncertainty reduction, syntactic structure, and intrinsic dimensions of meaning**

| | ar | ca | cs | de | el | en | es | eu | fa | fr | hu | id | ja | ko | nl | pl | ta | tr | vi | zh |
|---|---|---|---|---|---|---|---|---|---|---|---|---|---|---|---|---|---|---|---|---|
| Omega | 0.303*** | 0.151** | 0.305*** | 0.167*** | 0.326*** | 0.198*** | 0.414*** | 0.022 | 0.125** | 0.186*** | 0.190*** | 0.303*** | 0.194*** | 0.228*** | 0.216*** | 0.297*** | 0.327*** | 0.353*** | 0.043 | 0.281*** |
| Mean Hierarchical Distance | 0.293*** | 0.187*** | 0.330*** | 0.119** | 0.398*** | 0.170*** | 0.454*** | 0.107# | 0.155*** | 0.152*** | 0.200*** | 0.305*** | 0.148** | 0.177*** | 0.200*** | 0.325*** | 0.322*** | 0.157# | 0.123** | 0.232*** |
| Subtree-size Unevenness | 0.276*** | 0.191*** | 0.396*** | 0.186*** | 0.362*** | 0.204*** | 0.524*** | 0.129* | 0.152*** | 0.208*** | 0.228*** | 0.420*** | 0.216*** | 0.225*** | 0.261*** | 0.328*** | 0.245** | 0.181* | 0.198*** | 0.267*** |
| B2 Index | 0.191*** | 0.211*** | 0.277*** | 0.149** | 0.273*** | 0.184*** | 0.415*** | 0.119* | 0.115* | 0.191*** | 0.175*** | 0.382*** | 0.247*** | 0.167*** | 0.208*** | 0.163*** | 0.079 | 0.109 | 0.252*** | 0.217*** |
| ID Token | 0.303*** | 0.364*** | 0.550*** | 0.295*** | 0.224*** | 0.197*** | 0.124** | 0.226*** | 0.017 | 0.241*** | 0.331*** | 0.046 | 0.206*** | 0.017 | 0.350*** | 0.296*** | 0.454*** | 0.525*** | 0.050 | 0.240*** |
| ID Embedding | -0.336*** | -0.306*** | -0.502*** | -0.286*** | -0.233*** | -0.177*** | -0.192*** | -0.223*** | -0.039 | -0.251*** | -0.285*** | -0.064 | -0.234*** | -0.012 | -0.275*** | -0.268*** | -0.369*** | -0.455*** | -0.050 | -0.204*** |

Meta-analytic effects with 95% CI

Total effect

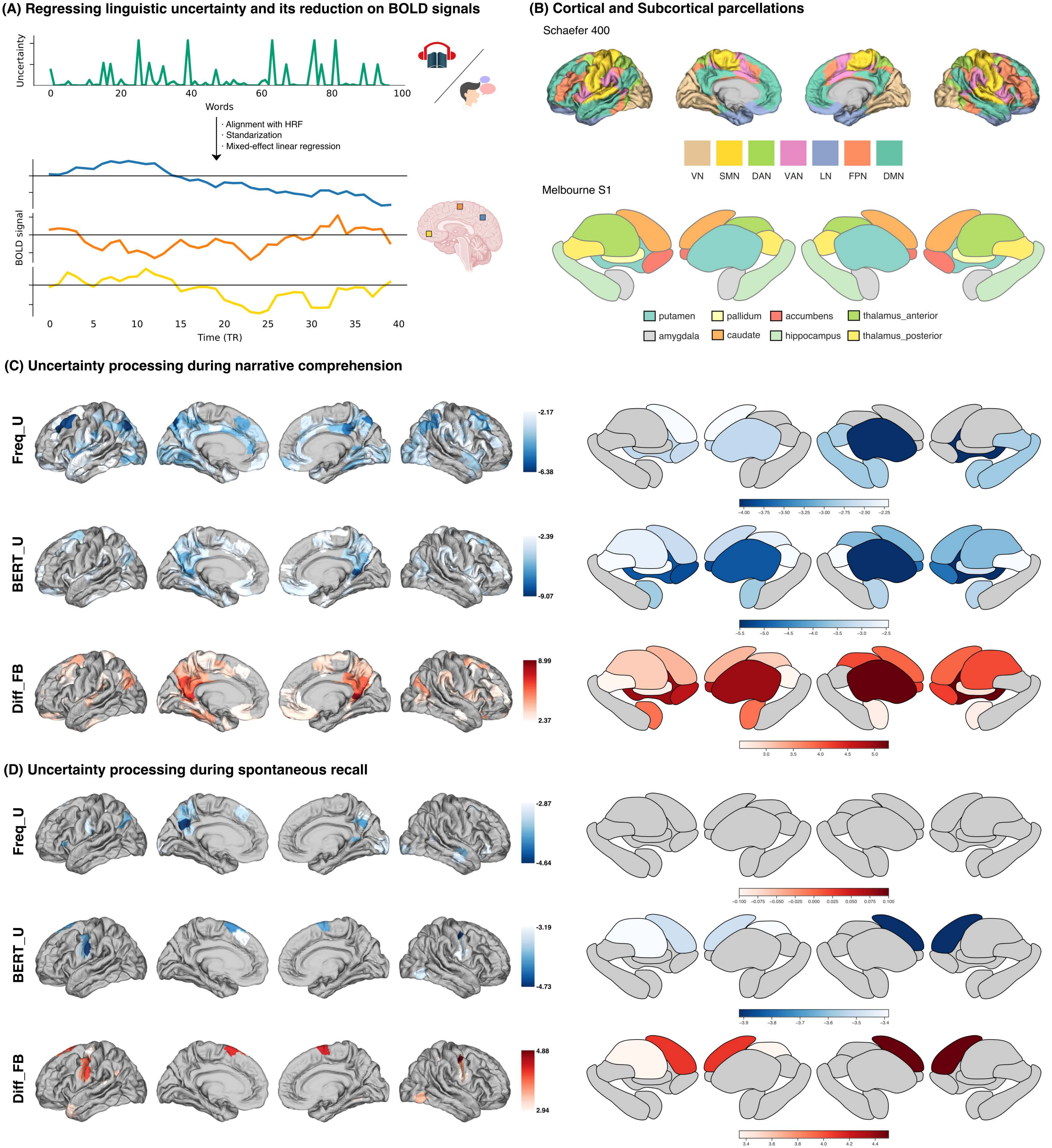

(A) Regressing linguistic uncertainty and its reduction on BOLD signals
Uncertainty
Words
· Alignment with HRF
· Standarization
· Mixed-effect linear regression
BOLD signal
Time (TR)
(B) Cortical and Subcortical parcellations
Schaefer 400
VN
SMN
DAN
VAN
LN
FPN
DMN
Melbourne S1
putamen
pallidum
accumbens
thalamus_anterior
amygdala
caudate
hippocampus
thalamus_posterior
(C) Uncertainty processing during narrative comprehension
Freq_U
BERT_U
Diff_FB
(D) Uncertainty processing during spontaneous recall
Freq_U
BERT_U
Diff_FB

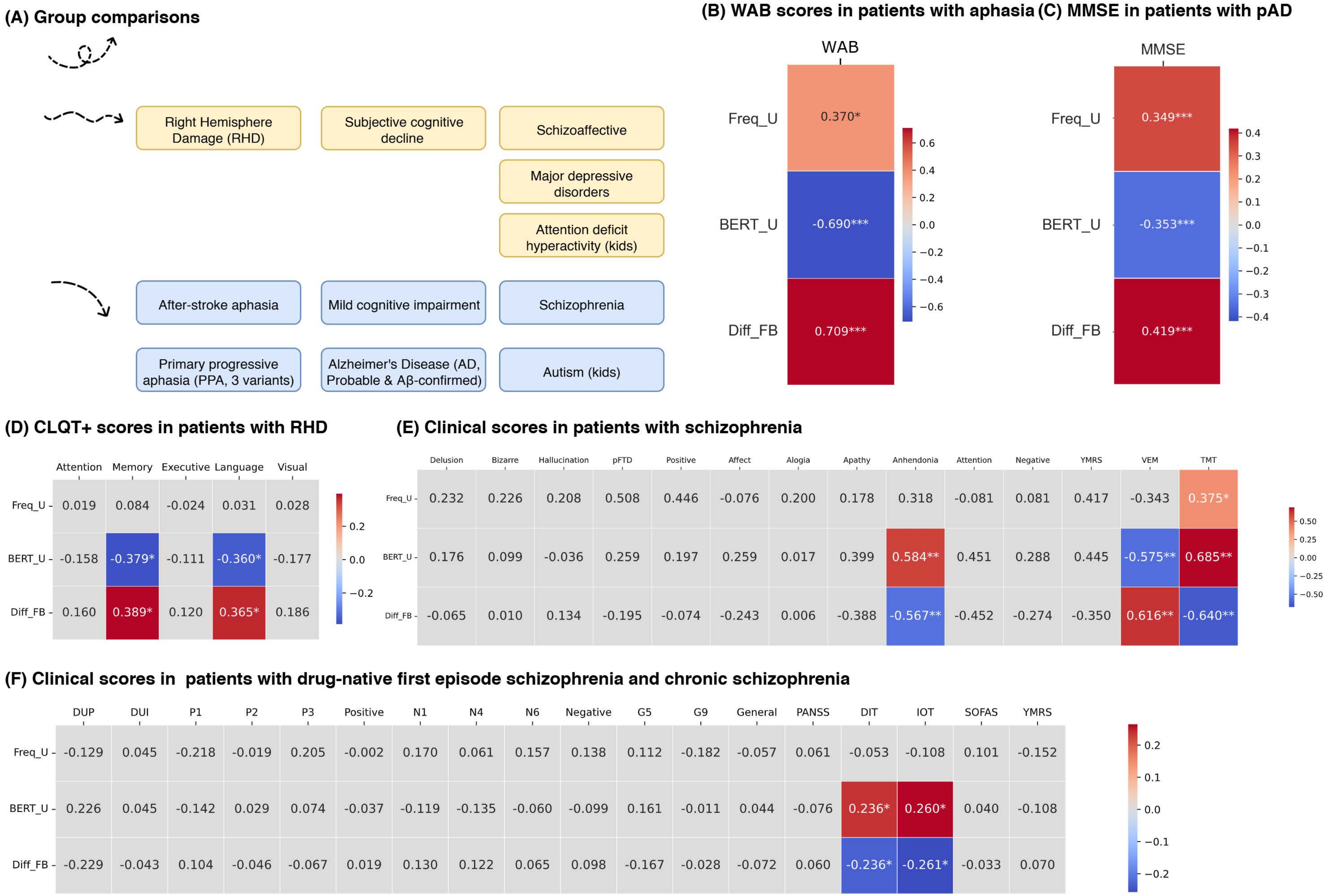

(A) Group comparisons

Right Hemisphere Damage (RHD)
Subjective cognitive decline
Schizoaffective
Major depressive disorders
Attention deficit hyperactivity (kids)
After-stroke aphasia
Mild cognitive impairment
Schizophrenia
Primary progressive aphasia (PPA, 3 variants)
Alzheimer's Disease (AD, Probable & Aβ-confirmed)
Autism (kids)

(B) WAB scores in patients with aphasia

WAB
Freq_U 0.370*
BERT_U -0.690***
Diff_FB 0.709***

(C) MMSE in patients with pAD

MMSE
Freq_U 0.349***
BERT_U -0.353***
Diff_FB 0.419***

(D) CLQT+ scores in patients with RHD

(E) Clinical scores in patients with schizophrenia

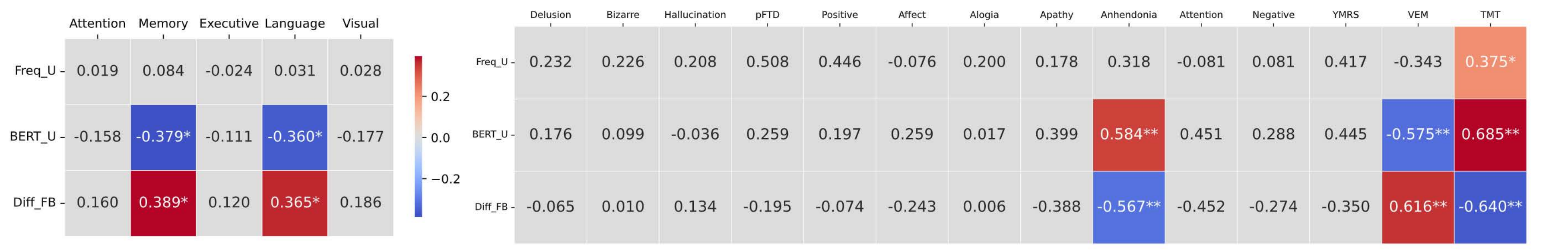

(F) Clinical scores in patients with drug-native first episode schizophrenia and chronic schizophrenia

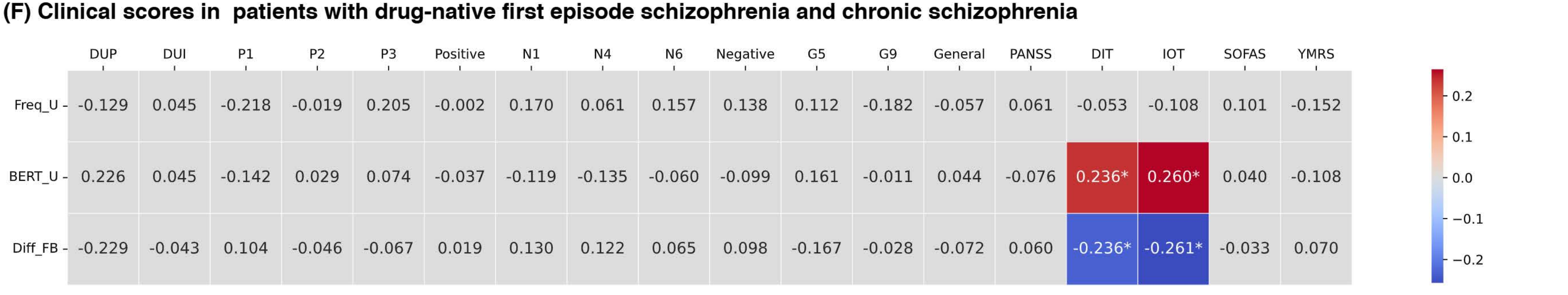

**Supplementary information**

# Contents

## 1 Detailed information about the clinical cohorts

The number of participants may differ from the number of participants in the original dataset due to exclusion based on sentence length. There were participants who produced only very short sentences that were all excluded and hence not entered our analyses. Here we reported the number and statistics based on participants who entered our analyses.

### 1.1 Olness corpus

This corpus included speech transcripts from 32 healthy controls without brain injuries and 49 patients with post-stroke aphasia and left hemisphere lesion.[1] Participants went through a wide variety of discourse tasks, with details available in the original paper. The dataset is available on the AphasiaBank: https://talkbank.org/aphasia/access/English/NonProtocol/Olness.html. Severity of the patients was assessed using the Western Aphasia Battery (WAB).

Supplementary Table 1. Demographics and clinical data of the Olness corpus.

| Variables | Non-brain-injuries | Aphasia | Test | p values |
|---|---|---|---|---|
| Count | 32 | 49 | / | / |
| Age | 54.50 (4.75) | 59.00 (13.00) | U test | 0.085 |
| Sex (Female%) | 65.62% | 48.98% | $X^2$ test | 0.213 |
| Ethics (African American%) | 56.25% | 42.86% | $X^2$ test | 0.341 |
| Education | 3.50 (2.00) | 3.00 (3.00) | U test | 0.395 |
| Socioeconomic status (SES) | 4.00 (3.00) | 5.00 (4.00) | U test | 0.674 |
| Western Aphasia Battery (WAB) scores | 99.05 (1.12) | 84.55 (16.17) | U test | 0.000 |
| Word Count | 115.84 (73.89) | 74.68 (68.39) | U test | 0.002 |

Note: Continuous variables were represented as median (interquartile range).

Missing values:

Age: NBI: 0 APH: 0

Sex: NBI: 0 APH: 0

Ethics: NBI: 0 APH: 0

Education: NBI: 0 APH: 0

SES: NBI: 0 APH: 0

WAB: NBI: 0 APH: 1

Word Count: NBI: 0 APH: 0

Overall, the aphasia and control groups were comparable in age, sex distribution, African American representation, years of education and socioeconomic status (SES), but the aphasia group produced significantly fewer words.

### 1.2 RHDBank English Minga Control Corpus

This corpus included speech transcripts from 35 controls and 37 patients with right hemisphere damage (RHD).

The dataset is available on the RHDBank (control: https://talkbank.org/rhd/access/English/Control.html; RHD: https://talkbank.org/rhd/access/English/RHD.html). Details were reported in the original paper.[2]

Supplementary Table 2. Demographics and clinical data of the RHDBank English Minga Control Corpus.

| Variables | Controls | RHD | Test | p values | Normal range |
|---|---|---|---|---|---|
| Count | 35 | 37 | / | / | |
| Age | 49.98 (20.82) | 56.22 (11.96) | U test | 0.004 | |
| Sex (Female%) | 65.71% | 51.35% | $X^2$ test | 0.319 | |
| Education | 17.00 (3.00) | 16.00 (5.00) | U test | 0.865 | |
| CLQT Attention | 204.50 (9.25) | 186.00 (56.00) | U test | 0.000 | 180–215 |
| CLQT Memory | 171.00 (12.25) | 166.00 (20.00) | U test | 0.177 | 155–185 |
| CLQT Executive Function | 33.00 (3.00) | 27.00 (7.00) | U test | 0.000 | 24–40 |
| CLQT Language | 33.00 (3.75) | 31.00 (3.00) | U test | 0.010 | 29–37 |
| CLQT Visuospatial | 98.00 (5.00) | 83.00 (25.00) | U test | 0.000 | 82–105 |
| Word Count | 195.22 (99.63) | 237.88 (134.93) | U test | 0.037 | |

Note: Continuous variables were represented as median (interquartile range).

Missing values:

Age: HC: 0 RHD: 0

Sex: HC: 0 RHD: 0

Education: HC: 0 RHD: 0

CLQT Attention: HC: 1 RHD: 0

CLQT Memory: HC: 1 RHD: 0

CLQT Executive Function: HC: 1 RHD: 0

CLQT Language: HC: 1 RHD: 0

CLQT Visuospatial: HC: 1 RHD: 0

Word Count: HC: 0 RHD: 0

### 1.3 Rezaii PPA cohort

This cohort includes picnic picture descriptions from 53 controls and 76 patients with PPA. Patients were categorized into three variants of PPA, including 28 non-fluent variants (nfvPPA), 23 semantic variants (svPPA), and 25 logopenic variants (lvPPA). Details were reported in the original paper.[3]

Supplementary Table 3. Demographics and clinical data of the Rezaii PPA dataset.

| Variables | HC | nfvPPA | svPPA | lvPPA | Test | p values |
|---|---|---|---|---|---|---|
| Count | 53 | 28 | 23 | 25 | / | / |
| Age | 63.93 (9.55) | 69.61 (12.84) | 64.83 (11.60) | 68.84 (5.94) | KW test | 0.024 |
| Sex (Female%) | 62.26% | 55.56% | 59.09% | 36.00% | $X^2$ test | 0.177 |
| Education | 16.00 (2.00) | 16.50 (3.75) | 16.00 (2.00) | 16.00 (2.25) | KW test | 0.020 |
| Word Count | 174.00 (163.00) | 59.50 (48.75) | 87.00 (53.00) | 125.00 (70.00) | KW test | 0.000 |

Note: Continuous variables were represented as median (interquartile range). KW test: Kruskal-Wallis test.

Missing values:

Age: HC: 0 nfvPPA: 0 svPPA: 0 lvPPA: 0

Sex: HC: 0 nfvPPA: 1 svPPA: 1 lvPPA: 0

Education: HC: 1 nfvPPA: 2 svPPA: 2 lvPPA: 1

Word Count: HC: 0 nfvPPA: 0 svPPA: 0 lvPPA: 0

### 1.4 The ADReSS challenge dataset

This cohort includes Cookie Theft picture description from 122 patients with probable Alzheimer's disease (pAD) and 115 matched controls. The dataset has been carefully selected so as to mitigate common biases often overlooked in evaluations of AD detection methods, including repeated occurrences of speech from the same participant (common in longitudinal datasets), variations in audio quality, and imbalances of gender and age distribution. Details were reported in the challenge description paper.[4]

Supplementary Table 4. Demographics and clinical data of the ADReSS challenge dataset.

| Variables | HC | pAD | Test | p values |
|---|---|---|---|---|
| Count | 115 | 122 | / | / |
| Age | 66.00 (10.00) | 70.00 (10.00) | U test | 0.000 |
| Sex (Female%) | 0.00% | 0.00% | $X^2$ test | 0.951 |
| Education | 13.00 (4.00) | 12.00 (4.00) | U test | 0.000 |
| MMSE | 29.00 (2.00) | 18.50 (7.75) | U test | 0.000 |
| Word Count | 101.00 (64.50) | 73.50 (52.25) | U test | 0.000 |

Note: Continuous variables were represented as median (interquartile range).

Missing values:

Age: HC: 0 pAD: 0

Sex: HC: 0 pAD: 0

Education: HC: 22 pAD: 0

MMSE: HC: 1 pAD: 0

Word Count: HC: 0 pAD: 0

### 1.5 The ACE dataset

Patients were recruited and assessed at the Memory Clinic from ACE Alzheimer Center Barcelona. All participants were recruited and diagnosed at Fundació ACE and were native Spanish and/or Catalan speakers or native bilingual speakers of both languages. All participants were required to be functionally literate and have at least basic primary school experience. Additionally, participants could not have diagnoses of any other neurological disorder or impairment, including aphasia, and needed to be without severe auditory or visual abnormalities including glaucoma, cataracts, or any major vision or hearing impairments that might create difficulty in completing an interview or a picture description task. Participation was voluntary, and participants were informed that they could stop the interview at any time, or withdraw their data from the project at any point during or after the study. All data were collected from February 2020 to September 2021. Informed consent was obtained from all participants. The referral center ethics committee (Hospital Clínic i Provincial of Barcelona) approved the patient recruitment and collection protocols were in accordance with ethical standards according to the World Medical Association Declaration of Helsinki - Ethical Principles for Medical Research Involving Human Subjects.

All diagnoses were assigned at a daily consensus conference among neurologists, neuropsychologists and social workers. Participants received standardized neurobehavioral exams, including neurological examination, neuropsychological testing, and social work evaluations. Information about vascular risk factors (including hypertension, hypercholesterolemia, diabetes mellitus, history of stroke, heart disease) and family history of dementia was provided by the patients or their caregivers. Healthy older adults (HOA, n = 17) were considered to be cognitively normal by a Neurologist and a Neuropsychologist, that is, there were no cognitive complaints by the subject or informant, no evidence by history of functional impairment due to declining cognition, a Mini-Mental State Examination (MMSE) score >24, and no cognitive impairment as measured by the neuropsychological battery from the diagnostic unit. Subjective cognitive decline (SCD, n = 31) was defined as the coexistence of cognitive complaints and a score of ⩾ 8 on the Spanish Modified Questionnaire of Memory Failures Every day (MFE-30). Inclusion criteria were (a) subjects older than 49 years, (b) Mini Mental State Exam (MMSE) ⩾ 27, (c) Clinical Dementia Rating (CDR) = 0, and (d) performance in the Fundació ACE Neuropsychological Battery (NBACE) within the normal range for age and education. Mild cognitive impairment (MCI, n = 39) patients fulfilled Petersen's MCI diagnostic criteria, including subjective memory complaints, normal general cognition, preserved performance of daily living activities, absence of dementia, and a measurable impairment in one or more cognitive functions. Patients had a Clinical Dementia Rating Scale (CDR) of 0.5, and were classified into aMCI and naMCI, single or multiple domains, including subjective memory complaints, normal general cognition, preserved performance in activities of daily living, absence of dementia, and measurable impairment in one or more cognitive functions. Moreover, taking into account Lopez's classification, probable or possible MCI subtypes were added in function of the absence or presence, respectively, of comorbidities that could otherwise explain their cognitive deficits. Probable Alzheimer's disease (pAD, n = 31) diagnoses were based on NINCDS/ADRDA criteria up to 2014 and on NIA-AA criteria thereafter. Patients with potential causes of dementia other than Alzheimer's were excluded (i.e. Lewy bodies, Parkinson's, pure vascular dementia, and frontotemporal dementia). Only patients with a Clinical Dementia Rating (CDR) score between 1 and 4 were included.

All patients with MCI completed the neuropsychological battery used in Ace Alzheimer Center Barcelona (NBACE). This diagnostic procedure assesses eight cognitive domains: 1) Orientation - temporal, spatial, and personal orientation; 2) Attention and working memory - digit spans (forwards and backwards) subtests from the Wechsler adult intelligence scale–third edition (WAIS-III); 3) Processing speed and Executive function - the automatic inhibition subtest from the Syndrom-Kurztest (SKT), phonetic verbal fluency (words beginning with 'P' in 1 min), semantic verbal fluency ('animals' in 1 min), and similarities subtest from WAIS-III (abbreviated to the first 10 items); 4) Language - repetition (two words and two sentences), verbal comprehension (correctly execute two simple, two semi-complex, and two complex commands extracted from the Alzheimer's disease assessment scale (ADAS) and the Barcelona test battery), and an abbreviated 15-item Boston naming test; 5) Verbal Learning and Memory - word list learning test from the Wechsler memory

scale–third edition (WMS-III) (without using the interference list); 6) Praxis - block design subtests from WAIS-III (abbreviated so that items 6–9 were scored only for accuracy (1 point) without a time bonus), imitation praxis (four items), and ideomotor praxis (four items), 7) Visual gnosis - two Poppelreuter-type overlap figures, Luria's clock test, and the 15-objects test; and 8) Global cognition - the Spanish version of the clock test.

It is important to note that data collection, which began in February 2020, was impacted by the onset of the COVID-19 pandemic. Fortunately, Fundació ACE was able to implement measures to continue in-person protocols. In-person interviews for the spontaneous speech protocol began again in October 2020 and continued until September 2021, always strictly following guidelines set in place by Fundació ACE and governing bodies in place.

All participants completed two speech tasks, past-directed narratives and picture descriptions. They could choose to speak either in Spanish or Catalan.

Supplementary Table 5. Demographics and clinical data of the ADReSS challenge dataset.

| Variables | HOA | SCD | MCI | pAD | Test | p values |
|---|---|---|---|---|---|---|
| Count | 17 | 31 | 39 | 31 | / | / |
| Age | 66.00 (5.25) | 69.00 (10.00) | 75.00 (7.50) | 81.00 (8.50) | KW test | 0.000 |
| Sex(female%) | 0.00% | 0.00% | 0.00% | 0.00% | $X^2$ test | 0.139 |
| Language | 35.29% | 22.58% | 7.69% | 12.90% | X2 test | 0.058 |
| Education | 12.00 (8.00) | 10.00 (8.00) | 8.00 (6.00) | 8.00 (4.00) | KW test | 0.000 |
| Word Count | 198.50 (148.00) | 226.00 (234.25) | 128.00 (106.50) | 79.00 (69.25) | KW test | 0.000 |
| MMSE | 29.50 (1.00) | 29.00 (2.00) | 27.00 (3.00) | 23.00 (4.00) | KW test | 0.000 |
| AWM | 0.38 (1.23) | 0.08 (1.57) | -1.17 (1.56) | -1.79 (0.95) | KW test | 0.000 |
| EF | 1.08 (2.45) | 1.31 (3.21) | -3.24 (4.87) | -7.96 (6.17) | KW test | 0.000 |
| Language | 0.33 (0.00) | 0.33 (0.00) | -0.32 (1.62) | -1.61 (5.09) | KW test | 0.000 |
| Memory | 0.37 (3.44) | -0.50 (3.62) | -5.35 (6.34) | -10.83 (3.48) | KW test | 0.000 |
| Orientation | 0.46 (0.00) | 0.46 (0.00) | 0.46 (2.22) | -1.77 (5.54) | KW test | 0.000 |
| Praxis | 0.55 (0.18) | 0.55 (0.00) | 0.55 (2.43) | -5.62 (7.35) | KW test | 0.000 |
| Visuospatial | 1.10 (0.91) | 1.10 (1.27) | -1.62 (4.50) | -6.76 (7.90) | KW test | 0.000 |

Note: Continuous variables were represented as median (interquartile range). Language: represented by the ratio of Catalan speech. MMSE: mini-mental state examination. AWM: Attention working memory. EF: Executive function. KW test: Kruskal-Wallis test.

Missing values:

Age: HC: 1 SCD: 0 MCI: 0 pAD: 0

Sex: HC: 0 SCD: 0 MCI: 0 pAD: 0

Language: HC: 0 SCD: 0 MCI: 0 pAD: 0

Education: HC: 0 SCD: 0 MCI: 0 pAD: 0

WordCount: HC: 0 SCD: 0 MCI: 0 pAD: 0

MMSE: HC: 1 SCD: 0 MCI: 2 pAD: 2

Attention working memory: HC: 0 SCD: 0 MCI: 0 pAD: 0

Executive funciton: HC: 1 SCD: 0 MCI: 0 pAD: 0

Language: HC: 0 SCD: 0 MCI: 0 pAD: 0
Memory: HC: 0 SCD: 0 MCI: 0 pAD: 0
Orientation: HC: 0 SCD: 0 MCI: 0 pAD: 0
Praxis: HC: 1 SCD: 0 MCI: 0 pAD: 0
Visuospatial: HC: 0 SCD: 0 MCI: 0 pAD: 0

### 1.6 The Zhuhai CSF-confirmed AD cohort

The Zhuhai Alzheimer's disease (AD) cohort is part of an ongoing study designed to support AI-assisted detection of Alzheimer's disease using episodic-memory-based assessments and multimodal biomarkers. Recruitment targets older adults with AD and cognitively normal controls, with concurrent collection of clinical, neuropsychological scales, spontaneous speech, and imaging measures. As recruitment and curation are ongoing, the present analysis uses an interim subset of the cohort (controls, n = 10; AD, n = 5), primarily as an independent validation sample for the findings observed in the previously mentioned larger datasets.

Participants were older adults (≥50 years) who were native Chinese speakers (Mandarin and/or other Chinese varieties, including Cantonese). Participation was voluntary and required written informed consent from the participant, consistent with local ethical requirements and research governance. Participants in the AD group met clinical criteria for mild-to-moderate Alzheimer's disease, characterized by insidious onset and slow progression with prominent memory impairment that could co-occur with deficits in other cognitive domains. In addition to clinical diagnosis, this cohort was designed to include biomarker-supported AD ascertainment, with cerebrospinal fluid (CSF) biomarkers used to confirm AD etiology where available, addressing a key limitation in many speech-based AD studies that rely on clinical diagnosis alone.

Participants were included in the AD group if they met all of the following:

1. Age ≥ 50 years.
2. Native Chinese speaker (including Mandarin and/or other Chinese varieties).
3. Diagnosed with mild or moderate Alzheimer's disease.
4. CDR score ≥ 0.5.
5. Prominent memory decline, potentially accompanied by impairments in other cognitive domains.
6. Insidious onset and slow progression.

Participants were excluded from the AD group if any of the following applied:

1. Severe AD / dementia stage beyond mild-to-moderate severity.
2. Clinical diagnosis of vascular dementia (VaD).
3. MRI contraindications or inability to meet imaging/speech acquisition requirements (e.g., pacemaker, cochlear implant or other implanted electronic devices; severe claustrophobia; conditions incompatible with MRI scanning).
4. Severe sensory or speech limitations affecting participation (e.g., deafness, mutism, loss of voice).
5. History of stroke with focal neurological signs, and imaging findings consistent with significant cerebral

small vessel disease (Fazekas score > 2).

6. Neurodevelopmental delay/intellectual disability or other known conditions that could independently cause cognitive impairment.
7. Refusal to sign informed consent at baseline by the participant or family/caregiver.
8. Insufficient clinical documentation (CRF) to characterize symptoms and features.
9. Acute/critical illness or any condition judged to make participation unsafe or infeasible.

Controls were included if they met all of the following:

1. Age ≥ 50 years.
2. Native Chinese speaker (including Mandarin and/or other Chinese varieties).
3. No diagnosis of cognitive impairment.

Controls were excluded if they met any AD-group exclusion criteria, or if they met diagnostic thresholds suggestive of mild/moderate AD (e.g., CDR ≥ 0.5 with clinically significant cognitive impairment).

Participants completed a standardized set of clinical and laboratory assessments, including blood tests (e.g., routine hematology, liver/kidney function, lipid and glucose profiles), cognitive assessments, language/speech testing, and neuroimaging as part of the multimodal screening protocol. The study also provided transportation compensation to support participation.

All participants completed a standardized clinical workup and multimodal assessment battery. Routine laboratory testing included a complete blood count and urinalysis. Global cognitive screening included the Mini-Mental State Examination (MMSE) and Montreal Cognitive Assessment (MoCA), alongside the Clinical Dementia Rating (CDR) for staging. Neuroimaging was acquired to support etiological characterization and to exclude prominent vascular pathology. Structural MRI included T1-weighted and T2-weighted sequences with FLAIR. Functional imaging included approximately 10 minutes of resting-state fMRI and arterial spin labeling (ASL) to characterize cerebral perfusion. Imaging findings were used in conjunction with clinical history to exclude vascular dementia and substantial small vessel disease.

Spontaneous speech was elicited using an adapted version of the DISCOURSE protocol, designed to elicit naturalistic connected speech suitable for quantitative analysis. Participants completed standardized elicitation tasks under this protocol, and responses were recorded and transcribed. Tasks included:

1. Self-related speech: Tell me about yourself.
2. Past-directed speech: Describe the most important event in your life
3. Picture descriptions of: (1) Cookie Theft; (2) Birthday party; and (3) the farmland picture from thematic appreciation test.
4. Caroon description: describe what happened in a cartoon of six images
5. Immediate recall: read the crown story aloud and recall the story immediately after the reading

Supplementary Table 6. Demographics and clinical data of the Zhuhai CSF-confirmed AD cohort.

| Variables | Control | AD | Test | p values |
|---|---|---|---|---|
| Count | 34 | 30 | / | / |
| Age | 66.50 (7.00) | 72.50 (13.50) | U test | 0.004 |
| Sex (Female%) | 58.82% | 66.67% | $X^2$ test | 0.698 |
| MMSE | 26.50 (4.75) | 17.00 (11.75) | U test | 0.000 |
| MoCA | 22.00 (6.00) | 12.00 (9.00) | U test | 0.000 |
| Word Count | 142.50 (78.32) | 97.00 (98.46) | U test | 0.020 |

Note: Continuous variables were represented as median (interquartile range).

Missing values:

Age: HC: 0 AD: 0

Sex: HC: 0 AD: 0

MMSE: HC: 0 AD: 0

MoCA: HC: 0 AD: 0

Word Count: HC: 0 AD: 0

### 1.7 The Dutch Asymmetries Corpus

The Asymmetries Project collection contains Dutch language productions gathered in Groningen and neighboring towns in the northern Netherlands, between 2007 and 2012. Details on these corpora could be found in the webpages: ASDBank: https://talkbank.org/asd/access/Dutch/Asymmetries.html, CHILDES https://talkbank.org/childes/access/DutchAfrikaans/Asymmetries.html.

Supplementary Table 7. Demographics and clinical data of the Dutch Asymmetries corpus.

| Variables | TD | ASD | ADHD | Test | p values |
|---|---|---|---|---|---|
| Count | 69 | 46 | 37 | / | / |
| Age (by month) | 79.00 (47.00) | 110.50 (28.75) | 103.00 (33.00) | KW test | 0.000 |
| Sex (Female%) | 0.00% | 0.00% | 0.00% | $X^2$ test | 0.005 |
| Word Count | 46.50 (13.50) | 48.50 (12.44) | 50.50 (23.25) | KW test | 0.064 |

Note: Continuous variables were represented as median (interquartile range). KW test: Kruskal-Wallis test.

Missing values:

Age: TD: 0 ASD: 0 ADHD: 0

Sex: TD: 0 ASD: 0 ADHD: 0

WordCount: TD: 0 ASD: 0 ADHD: 0

### 1.8 The TOPSY cohort

The first psychosis cohort included spontaneous speech from 39 healthy controls (HC), 72 patients with first-episode psychosis (FEP), and 20 patients with chronic schizophrenia (CSZ), , as a part of the Tracking Outcomes in Psychosis (TOPSY) study (https://clinicaltrials.gov/study/NCT02882204).[5,6] In brief, FEP subjects had <2 weeks of lifetime antipsychotic exposure and in most cases were assessed in the first week of referral to the first-episode psychosis team. As such, the median dose of antipsychotic exposure, calculated by converting the various prescribed antipsychotic medication doses to a common equivalent on the basis of Defined Daily Dose (DDD) provided by the WHO Collaborating Centre for Drug Statistics and Methodology (https://www.whocc.no/atc_ddd_index_and_guidelines/guidelines/) and multiplying by the days of exposure to this dose, was <3 DDD-days in this sample. Only the data from FEP whose diagnosis remained stable (as schizophrenia, excluding those who had bipolar disorder or depressive psychosis) after 6 months of follow-up are included in this study. CSZ consisted of 20 participants that were clinically stable on long-acting injectable medications with >3 years since illness onset and no recorded hospitalization in the past year and receiving community-based care from physicians affiliated to a first-episode clinic (PEPP, London Ontario). Importantly, all participants were recruited regardless of the status of disorganization/thought disorder in their prior history, which was in order not to bias our sample towards language-related symptomatology. All diagnostic assessments were reviewed using a Best Estimate Procedure for clinical consensus (treating physician, a research psychiatrist and evaluators).[7] All patients provided written informed consent as stipulated by the Research Ethics Committee of University of Western Ontario, London, Canada (ID 108268).

All subjects were asked to describe three pictures from the Thematic Apperception Test and were given one minute for each image. During the speech, if any participant would finish their descriptions in less than one minute, the interviewer would prompt them to speak more, and if they were continuing beyond one minute, the interviewer would interrupt them. This procedure makes the quantity of speech relatively similar across groups. The recorded speech was transcribed by research assistants. Elicited speech was scored using the Thought Language Index (TLI), with scores for impoverishment of thought and for disorganization in thinking.[8] Patients with FEP and CSZ were assessed with the Positive and Negative Syndrome Scale-8 items version (PANSS), with delusions (PANSS8P1), conceptual disorganization (PANSS8P2), hallucinatory behavior (PANSS8P3), blunted affect (PANSS8N1), passive/apathetic social withdrawal (PANSS8N4), lack of spontaneity/flow of conversation (PANSS8N6), mannerisms/posturing (PANSS8G5), and unusual thought content (PANSS8G9).[9]

Supplementary Table 8. Demographics and clinical data of the TOPSY dataset.

| Variables | HC | FEP | CSZ | Test | p values |
|---|---|---|---|---|---|
| Count | 39 | 72 | 20 | / | / |
| Age | 22.00 (3.00) | 22.00 (5.00) | 28.00 (6.00) | KW test | 0.000 |
| Sex (Female%) | 33.33% | 18.31% | 20.00% | $X^2$ test | 0.190 |
| Education | 14.00 (3.00) | 12.00 (2.00) | 12.00 (2.50) | KW test | 0.001 |
| SES | 3.00 (2.00) | 4.00 (3.00) | 4.00 (2.00) | KW test | 0.425 |
| DUP | / | 20.00 (67.00) | 30.00 (96.50) | KW test | 0.601 |
| DUI | | 104.00 (189.50) | 104.00 (281.25) | KW test | 0.533 |
| P1: Delusion | 1.00 (0.00) | 5.00 (1.00) | 2.50 (2.75) | KW test | 0.000 |
| P2: Conceptual disorganization | 1.00 (0.00) | 3.00 (3.00) | 1.00 (1.00) | KW test | 0.000 |
| P3: Hallucinatory behavior | 1.00 (0.00) | 4.00 (2.00) | 3.00 (3.00) | KW test | 0.000 |
| N1: Blunted affect | 1.00 (0.00) | 2.00 (3.00) | 1.00 (1.00) | KW test | 0.000 |
| N4: Passive/apathetic social withdrawal | 1.00 (0.00) | 3.00 (4.00) | 1.00 (1.00) | KW test | 0.000 |
| N6: Lack of spontaneity and flow of conversation | 1.00 (0.00) | 1.00 (2.00) | 1.00 (0.00) | KW test | 0.000 |
| G5: Mannerisms & posturing | 1.00 (0.00) | 1.00 (2.00) | 1.00 (0.00) | KW test | 0.000 |
| G9: Unusual thought content | 1.00 (0.00) | 4.00 (2.00) | 1.50 (1.75) | KW test | 0.000 |
| PANSS Positive | 3.00 (0.00) | 12.00 (3.25) | 7.50 (5.75) | KW test | 0.000 |
| PANSS Negative | 2.00 (0.00) | 5.00 (3.00) | 3.00 (2.00) | KW test | 0.000 |
| PANSS General | 3.00 (0.00) | 7.00 (8.00) | 3.50 (1.00) | KW test | 0.000 |
| PANSS TOTAL | 8.00 (0.00) | 25.00 (9.00) | 14.50 (9.00) | KW test | 0.000 |
| TLI DIT | 0.00 (0.08) | 0.17 (0.50) | 0.04 (0.17) | KW test | 0.000 |
| TLI IOT | 0.00 (0.08) | 0.08 (0.25) | 0.08 (0.23) | KW test | 0.006 |
| SOFAS | 81.00 (5.00) | 40.00 (14.50) | 60.50 (12.00) | KW test | 0.000 |
| YRMS Total | 0.00 (0.00) | 12.50 (8.25) | 0.00 (4.50) | KW test | 0.000 |
| Word Count | 140.33 (46.00) | 123.67 (67.17) | 134.83 (49.08) | KW test | 0.086 |

Note: Continuous variables were represented as median (interquartile range). KW test: Kruskal-Wallis test. SES: Socioeconomic status. DUP: duration of untreated psychosis. DUI: duration of untreated illness. TLI: thought and language index. TLI includes two general types of disturbances: disorganization in thought (DIT), and impoverishment of thought (IOT). SOFAS: Social and Occupational Functioning Assessment Scale. YMRS: Young Mania Rating Scale.

Missing values:

Age: HC: 0 FEP: 1 CSZ: 2

Sex: HC: 0 FEP: 1 CSZ: 0

Education: HC: 1 FEP: 3 CSZ: 1

SES: HC: 0 FEP: 4 CSZ: 0

DUP: HC: 39 FEP: 16 CSZ: 6

DUI: HC: 39 FEP: 16 CSZ: 4

P1: Delusion: HC: 0 FEP: 7 CSZ: 2

P2: Conceptual disorganization: HC: 0 FEP: 6 CSZ: 1

P3: Hallucinatory behavior: HC: 0 FEP: 8 CSZ: 2

N1: Blunted affect: HC: 0 FEP: 7 CSZ: 2

N4: Passive/apathetic social withdrawal: HC: 0 FEP: 7 CSZ: 2

N6: Lack of spontaneity and flow of conversation: HC: 0 FEP: 7 CSZ: 2

G5: Mannerisms & posturing: HC: 0 FEP: 7 CSZ: 2

G9: Unusual thought content: HC: 0 FEP: 7 CSZ: 2

PANSS Positive: HC: 0 FEP: 8 CSZ: 2

PANSS Negative: HC: 0 FEP: 7 CSZ: 2

PANSS General: HC: 0 FEP: 7 CSZ: 2

PANSS TOTAL: HC: 0 FEP: 8 CSZ: 2

TLI Disorganization: HC: 5 FEP: 2 CSZ: 2

TLI Impoverishment: HC: 3 FEP: 2 CSZ: 2

SOFAS: HC: 0 FEP: 1 CSZ: 0

YRMS Total: HC: 0 FEP: 0 CSZ: 0

Word Count: HC: 0 FEP: 0 CSZ: 0

### 1.9 The Marburg cohort

The second psychosis cohort recruited German speakers from Marburg, including 43 healthy controls (HC), 42 patients with major depressive disorder (MDD), 22 patients with schizoaffective disorders, and 20 patients with non-affective schizophrenia.[10] This is part of the FOR2107 MACS cohort (data freeze of the October 20, 2022, www.for2107.de). Details on speech elicitation and recruitment procedure can be found in Schneider et al. (2023).[10]

Supplementary Table 9. Demographics and clinical data of the Marburg cohort.

| Variables | HC | MDD | SZA | SZH | Test | p values |
|---|---|---|---|---|---|---|
| Count | 43 | 42 | 22 | 20 | / | / |
| Age | 43.12 (12.11) | 41.64 (12.25) | 41.68 (13.19) | 39.70 (10.36) | KW test | 0.777 |
| Sex (Female%) | 32.56% | 35.71% | 45.45% | 25.00% | $X^2$ test | 0.560 |
| Education | 14.68 (3.16) | 12.76 (2.57) | 11.52 (1.74) | 12.44 (2.06) | KW test | 0.000 |
| Delusion | 0.07 (0.34) | 0.07 (0.34) | 3.50 (4.59) | 6.06 (6.45) | KW test | 0.000 |
| Bizarre | 0.00 (0.00) | 0.24 (0.79) | 0.59 (0.94) | 0.41 (0.84) | KW test | 0.002 |
| Hallucination | 0.00 (0.00) | 0.05 (0.22) | 1.82 (4.37) | 3.59 (4.90) | KW test | 0.000 |
| pFTD | 0.88 (2.41) | 2.27 (3.64) | 4.55 (5.49) | 8.33 (9.38) | KW test | 0.000 |
| SAPS Positive | 0.65 (1.57) | 2.63 (4.12) | 10.45 (11.46) | 17.00 (14.88) | KW test | 0.000 |
| Affect | 1.00 (2.29) | 2.46 (2.18) | 8.11 (6.02) | 5.72 (5.66) | KW test | 0.000 |
| Alogia | 0.19 (0.63) | 1.56 (2.25) | 3.27 (3.15) | 2.00 (2.36) | KW test | 0.000 |
| Apathy | 0.07 (0.34) | 1.78 (2.39) | 4.86 (3.65) | 3.83 (3.58) | KW test | 0.000 |
| Anhendonia | 0.12 (0.39) | 2.71 (4.62) | 6.52 (5.66) | 2.89 (3.07) | KW test | 0.000 |
| Attention | 0.44 | 1.57 | 1.05 | 1.00 | KW test | 0.017 |
| | (0.96) | (2.02) | (1.97) | (2.05) | | |
| SANS Negative | 1.87 | 7.08 | 23.29 | 15.44 | KW test | 0.000 |
| | (3.11) | (18.60) | (15.04) | (11.72) | | |
| YMRS | 0.52 | 0.55 | 4.86 | 7.26 | KW test | 0.000 |
| | (1.74) | (1.05) | (6.01) | (7.76) | | |
| Verbal episodic memory | 57.83 | 57.83 | 48.86 | 44.58 | KW test | 0.000 |
| (VLMT Sum Corrects) | (10.29) | (9.15) | (7.61) | (10.61) | | |
| Executive function | 24.10 | 29.00 | 34.63 | 43.00 | KW test | 0.023 |
| (TMT Time Difference) | (15.00) | (19.97) | (19.81) | (32.38) | | |
| Word Count | 282.48 | 261.71 | 239.28 | 234.36 | KW test | 0.052 |
| | (87.76) | (92.14) | (110.98) | (93.18) | | |

Note: Continuous variables were represented as median (interquartile range). KW test: Kruskal-Wallis test. pFTD: positive formal thought disorder. SAPS: scale for the assessment of positive symptoms. SANS: scale for the assessment of negative symptoms. YMRS: Young Mania Rating Scale. VLMT: verbal learning and memory test. TMT: Trail Making Test .

Missing values:
Age: HC: 0 MDD: 0 SZA: 0 SZH: 0
Sex: HC: 0 MDD: 0 SZA: 0 SZH: 0
Education: HC: 2 MDD: 0 SZA: 1 SZH: 2
Delusion: HC: 1 MDD: 1 SZA: 0 SZH: 3
Bizarre: HC: 2 MDD: 1 SZA: 0 SZH: 3
Hallucination: HC: 1 MDD: 1 SZA: 0 SZH: 3
pFTD: HC: 2 MDD: 1 SZA: 0 SZH: 2
SAPS Positive: HC: 3 MDD: 1 SZA: 0 SZH: 3
Affect: HC: 19 MDD: 14 SZA: 3 SZH: 2
Alogia: HC: 1 MDD: 1 SZA: 0 SZH: 2
Apathy: HC: 2 MDD: 1 SZA: 0 SZH: 2
Anhendonia: HC: 1 MDD: 1 SZA: 1 SZH: 2
Attention: HC: 2 MDD: 2 SZA: 0 SZH: 2
SANS Negative: HC: 13 MDD: 2 SZA: 1 SZH: 2
YMRS: HC: 1 MDD: 0 SZA: 0 SZH: 1
Verbal episodic memory (VLMT sum corrects): HC: 3 MDD: 1 SZA: 1 SZH: 1
Executive function (TMT time Difference): HC: 1 MDD: 1 SZA: 3 SZH: 2
Word Count: HC: 0 MDD: 0 SZA: 0 SZH: 0

## 2 Additional information on fMRI preprocessing

### 2.1 Anatomical data preprocessing

A total of 1 T1-weighted (T1w) images were found within the input BIDS dataset. The T1-weighted (T1w) image was corrected for intensity non-uniformity (INU) with N4BiasFieldCorrection,[11] distributed with ANTs 2.3.3,[12] and used as T1w-reference throughout the workflow. The T1w-reference was then skull-stripped with a Nipype implementation of the antsBrainExtraction.sh workflow (from ANTs), using OASIS30ANTs as target template.

Brain tissue segmentation of cerebrospinal fluid (CSF), white-matter (WM) and gray-matter (GM) was performed on the brain-extracted T1w using fast (FSL 6.0.5.1:57b01774, RRID:SCR_002823).[13] Brain surfaces were reconstructed using recon-all (FreeSurfer 6.0.1, RRID:SCR_001847),[14] and the brain mask estimated previously was refined with a custom variation of the method to reconcile ANTs-derived and FreeSurfer-derived segmentations of the cortical gray-matter of Mindboggle (RRID:SCR_002438).[15] Volume-based spatial normalization to one standard space (MNI152NLin2009cAsym) was performed through nonlinear registration with antsRegistration (ANTs 2.3.3), using brain-extracted versions of both T1w reference and the T1w template. The following template was selected for spatial normalization: ICBM 152 Nonlinear Asymmetrical template version 2009c [RRID:SCR_008796; TemplateFlow ID: MNI152NLin2009cAsym].[16]

### 2.2 Functional data preprocessing

For each of the 1 BOLD runs found per subject (across all tasks and sessions), the following preprocessing was performed. First, a reference volume and its skull-stripped version were generated using a custom methodology of fMRIPrep. Head-motion parameters with respect to the BOLD reference (transformation matrices, and six corresponding rotation and translation parameters) are estimated before any spatiotemporal filtering using mcflirt (FSL 6.0.5.1:57b01774).[17] The BOLD time-series (including slice-timing correction when applied) were resampled onto their original, native space by applying the transforms to correct for head-motion. These resampled BOLD time-series will be referred to as preprocessed BOLD in original space, or just preprocessed BOLD. The BOLD reference was then co-registered to the T1w reference using bbregister (FreeSurfer) which implements boundary-based registration.[18] Co-registration was configured with six degrees of freedom. Several confounding time-series were calculated based on the preprocessed BOLD: framewise displacement (FD), DVARS and three region-wise global signals. FD was computed using two formulations following Power (absolute sum of relative motions)[19] and Jenkinson (relative root mean square displacement between affines).[17] FD and DVARS are calculated for each functional run, both using their implementations in Nipype (following the definitions by Power et al.[19]). The three global signals are extracted within the CSF, the WM, and the whole-brain masks. Additionally, a set of physiological regressors were extracted to allow for component-based noise correction (CompCor)[20]. Principal components are estimated after high-pass filtering the preprocessed BOLD time-series (using a discrete cosine filter with 128s cut-off) for the two CompCor variants: temporal (tCompCor) and anatomical (aCompCor). tCompCor components are then calculated from the top 2% variable voxels within the brain mask. For aCompCor, three probabilistic masks (CSF, WM and combined CSF+WM) are generated in anatomical space. The implementation differs from that of Behzadi et al.[20] in that instead of eroding the masks by 2 pixels on BOLD space, the aCompCor masks are subtracted a mask of pixels that likely contain a volume fraction of GM. This mask is obtained by dilating a GM mask extracted from the FreeSurfer's aseg segmentation, and it ensures components are not extracted from voxels containing a minimal fraction of GM. Finally, these masks are resampled into BOLD space and binarized by thresholding at 0.99 (as in the original implementation). Components are also calculated separately within the WM and CSF masks. For each CompCor decomposition, the k components with the largest singular values are retained, such that the retained components' time series are sufficient to explain 50 percent of variance across the nuisance mask (CSF, WM, combined, or temporal). The remaining components are dropped from consideration. The head-motion estimates calculated in the correction step were also placed within the corresponding confounds file. The confound time series derived from head motion estimates and global signals were expanded with the inclusion of temporal derivatives and quadratic terms for each.[21] Frames that exceeded a threshold of 0.5 mm FD or 1.5 standardised DVARS were annotated as motion outliers. The BOLD time-series were resampled into standard space, generating a preprocessed BOLD run in MNI152NLin2009cAsym space. First, a reference volume and its skull-stripped version were generated using a custom methodology of fMRIPrep. All resamplings can be performed with a single interpolation step by composing all the pertinent transformations (i.e. head-motion transform matrices, susceptibility distortion correction when available, and co-registrations to anatomical and output spaces). Gridded (volumetric) resamplings were performed using antsApplyTransforms (ANTs), configured with Lanczos interpolation to minimize the smoothing effects of other kernels.[22] Non-gridded (surface) resamplings were performed using mri_vol2surf (FreeSurfer).

## 3 BERT models

Supplementary Table 10. Names and commit hash of BERT models per language.

| Languages | | Bert model names | Commit Hash |
|---|---|---|---|
| Names | Codes | | |
| Arabic | ar | CAMeL-Lab/bert-base-arabic-camelbert-mix | 97522efce17efa33036ac619802d5cec238dcad9 |
| Catalan | ca | PlanTL-GOB-ES/roberta-base-ca | 9eddec657c2fa3ff828845def0096efbba1b0e08 |
| Czech | cs | ufal/robeczech-base | 67cfdc0dd05fa2690e14d7f88220c10cb8cb4c64 |
| German | de | dbmdz/bert-base-german-uncased | b705f0ee07840c383f9d9cb1e69cb2b60abbe780 |
| Greek | el | nlpaueb/bert-base-greek-uncased-v1 | ec2b8f88dd215b5246f2f850413d5bff90d7540d |
| English | en | google-bert/bert-base-uncased | 86b5e0934494bd15c9632b12f734a8a67f723594 |
| Spanish | es | bertin-project/bertin-roberta-base-spanish | d1c9c4565c9d6731e57ed7f027b802697bad861e |
| Basque | eu | ixa-ehu/berteus-base-cased | bb695ac9ac5f26ac15124c733024732850265149 |
| Persian | fa | HooshvareLab/bert-fa-base-uncased | a04aa40c97bcdde570ae11986a534542c2995a62 |
| French | fr | almanach/camembertv2-base | a75967561c78f2aa81cc41045378d3b4ee25af9e |
| Hungarian | hu | SZTAKI-HLT/hubert-base-cc | 028baac7feb87a7b2f042bbdaa5deec6513c6060 |
| Indonesian | id | indolem/indobert-base-uncased | 7ccb3cd0f5b08ffbaa465aade22328e8600e23eb |
| Japanese | ja | cl-tohoku/bert-base-japanese-v2 | e4211d7c20b078ac29b022be35ae4b63f3fe1679 |
| Korean | ko | snunlp/KR-Medium | 1779cc0982ada0216dd6de0dd4e86fb78201926d |
| Dutch | nl | pdelobelle/robbert-v2-dutch-base | aeb2fedfdd322b3a22195f250a43cd59c40832d3 |
| Polish | pl | dkleczek/bert-base-polish-cased-v1 | 50e33e0567be0c0b313832314c586e3df0dc2297 |
| Tamil | ta | l3cube-pune/tamil-bert | 7f8d7b97378c5dcc37d3f070153401033a256c91 |
| Turkish | tr | dbmdz/bert-base-turkish-uncased | 6cb8cd880acc6f7d9723161b573fce0dfd23b39b |
| Vietnamese | vi | FPTAI/vibert-base-cased | c1e37c5c86f918761049cef6fa216b4779d0d01d |
| Chinese | zh | hfl/chinese-roberta-wwm-ext | a986e004d2a7f2a1c2f5a3edef4e20604a974ed1 |

## 4 Supplementary results on associations between uncertainty measures and linguistic features

### 4.1 Correlations between surprisal and linguistic measures

Correlations between frequency-based surprisal and linguistic measures:

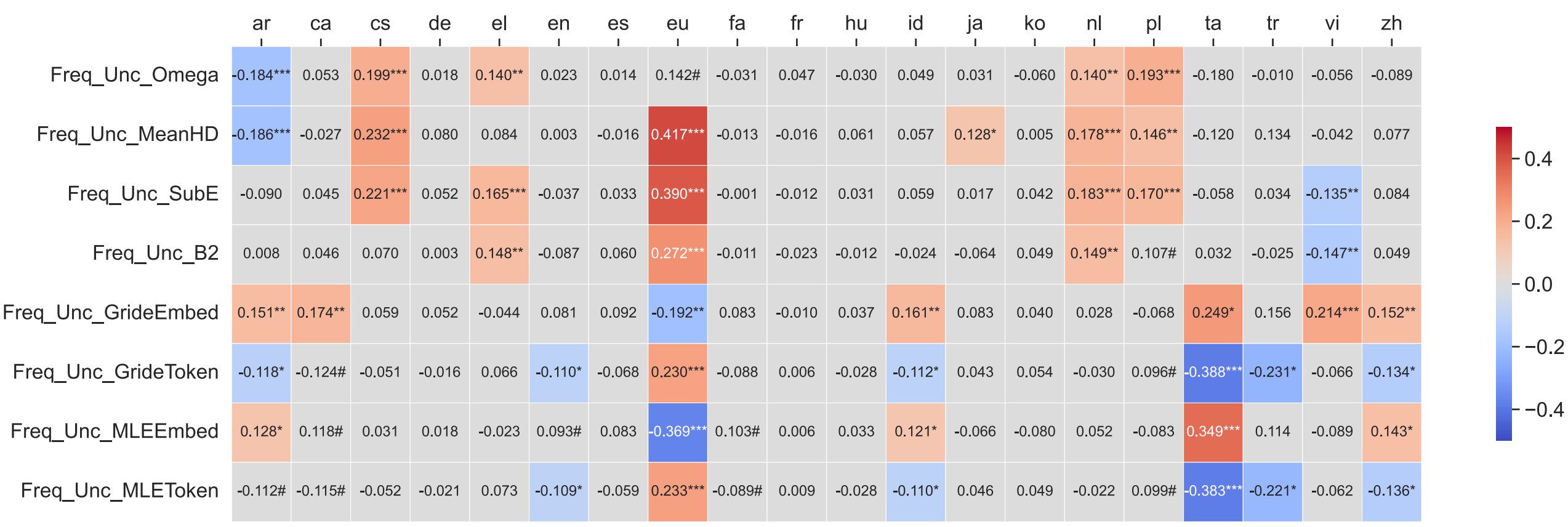

Supplementary Figure 1. Heatmap of Spearman correlations between frequency-based surprisal and each structural or semantic metric across 20 languages (columns). Colored cells indicate significant correlations between uncertainty reduction and linguistic variables (rows) in a certain language data (columns). Warm colors indicate positive correlations, cool colors negative correlations, and asterisks mark FDR-corrected significance levels. * $p < 0.05$, ** $p < 0.01$, *** $p < 0.001$. The linguistic measures are abbreviated as: dependency optimality (Omega), mean hierarchical distance (MeanHD), subtree size unevenness (SubE), B2 index (B2), intrinsic dimensions on the embedding feature manifold estimated by GRIDE (GrideEmbed), intrinsic dimensions on the token manifold estimated by GRIDE (GrideToken), intrinsic dimensions on the embedding feature manifold estimated by MLE (MLEEmbed), and intrinsic dimensions on the token manifold estimated by MLE (MLEToken).

Correlations between BERT-based surprisal and linguistic measures:

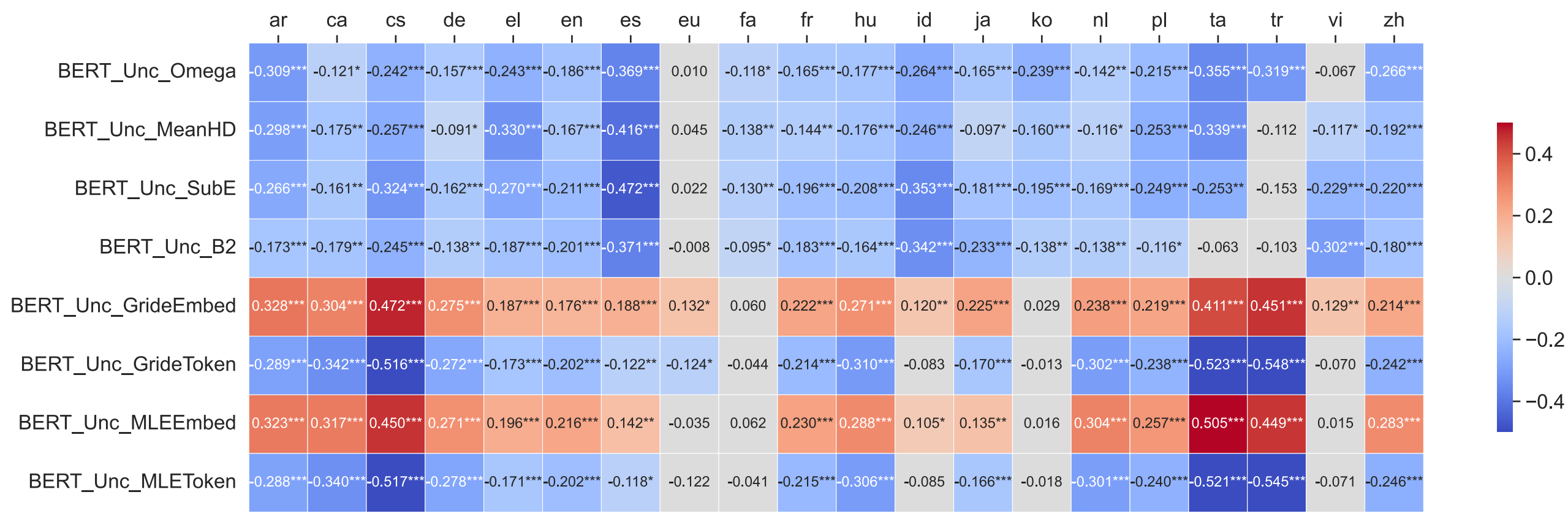

Supplementary Figure 2. Heatmap of Spearman correlations between BERT-based surprisal and each structural or semantic metric across 20 languages (columns). Colored cells indicate significant correlations

between uncertainty reduction and linguistic variables (rows) in a certain language data (columns). Warm colors indicate positive correlations, cool colors negative correlations, and asterisks mark FDR-corrected significance levels. * $p < 0.05$, ** $p < 0.01$, *** $p < 0.001$. The linguistic measures are abbreviated as: dependency optimality (Omega), mean hierarchical distance (MeanHD), subtree size unevenness (SubE), B2 index (B2), intrinsic dimensions on the embedding feature manifold estimated by GRIDE (GrideEmbed), intrinsic dimensions on the token manifold estimated by GRIDE (GrideToken), intrinsic dimensions on the embedding feature manifold estimated by MLE (MLEEmbed), and intrinsic dimensions on the token manifold estimated by MLE (MLEToken).

### 4.2 Correlations between surprisal reduction and MLE-based ID

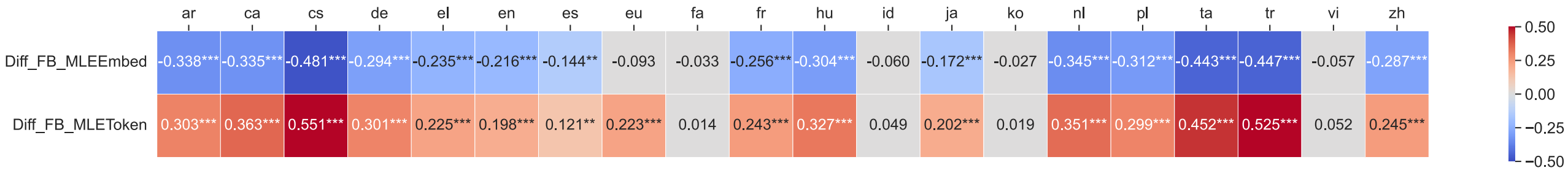

Supplementary Figure 3. Heatmap of Spearman correlations between surprisal reduction and ID values estimated by the MLE algorithm across 20 languages (columns). Colored cells indicate significant correlations between uncertainty reduction and linguistic variables (rows) in a certain language data (columns). Warm colors indicate positive correlations, cool colors negative correlations, and asterisks mark FDR-corrected significance levels. * $p < 0.05$, ** $p < 0.01$, *** $p < 0.001$.

### 4.3 Meta analysis

Supplementary Table 11. Meta analysis of the correlations between surprisal reduction and linguistic measures across 20 languages.

| Linguistic measures | Total Effect | CI_Lower | CI_Upper | Q | p_value_Q |
|---|---|---|---|---|---|
| Omega | 0.231 | 0.188 | 0.275 | 91.369 | 1.90E-11 |
| MeanHD | 0.229 | 0.184 | 0.274 | 106.886 | 2.98E-14 |
| SubE | 0.264 | 0.219 | 0.310 | 128.210 | 3.18E-18 |
| B2 | 0.213 | 0.176 | 0.251 | 69.945 | 9.39E-08 |
| GrideToken | 0.252 | 0.184 | 0.319 | 225.965 | 2.19E-37 |
| GrideEmbed | -0.236 | -0.293 | -0.180 | 152.850 | 6.17E-23 |
| MLEEmbed | -0.335 | -0.304 | -0.182 | 172.442 | 9.45E-27 |
| MLEToken | 0.380 | 0.194 | 0.333 | 240.429 | 2.67E-40 |
| Diff_Rev | 0.939 | 0.929 | 0.950 | 299.818 | 2.18E-52 |

Note: The linguistic measures are abbreviated as: dependency optimality (Omega), mean hierarchical distance (MeanHD), subtree size unevenness (SubE), B2 index (B2), intrinsic dimensions on the embedding feature manifold estimated by GRIDE (GrideEmbed), intrinsic dimensions on the token manifold estimated by GRIDE (GrideToken), intrinsic dimensions on the embedding feature manifold estimated by MLE (MLEEmbed), intrinsic dimensions on the token manifold estimated by MLE (MLEToken), and the surprisal reduction from normal to reversed word order (Diff_Rev).

The forest plots are below:

**Confidence interval**

```
ar  0.30( 0.21 to 0.39)
ca  0.15( 0.05 to 0.25)
cs  0.30( 0.21 to 0.39)
de  0.17( 0.08 to 0.25)
el  0.33( 0.24 to 0.40)
en  0.20( 0.11 to 0.28)
es  0.41( 0.34 to 0.48)
eu  0.02(-0.09 to 0.14)
fa  0.13( 0.04 to 0.21)
fr  0.19( 0.10 to 0.27)
hu  0.19( 0.10 to 0.27)
id  0.30( 0.22 to 0.38)
ja  0.19( 0.11 to 0.28)
ko  0.23( 0.14 to 0.31)
nl  0.22( 0.13 to 0.30)
pl  0.30( 0.21 to 0.37)
ta  0.33( 0.16 to 0.47)
tr  0.35( 0.21 to 0.48)
vi  0.04(-0.05 to 0.13)
zh  0.28( 0.20 to 0.36)
```

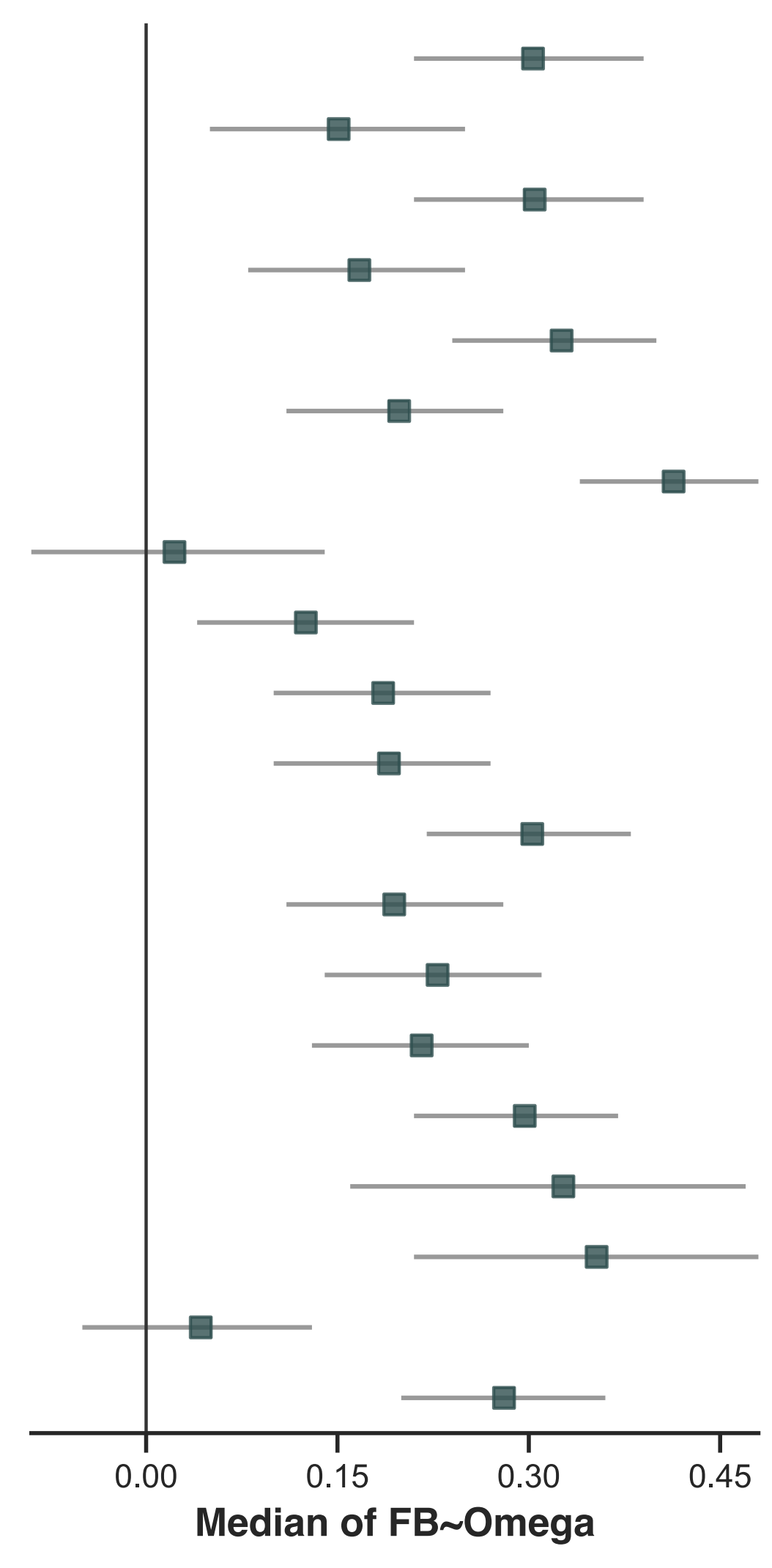

Supplementary Figure 4. Forest plot showing the correlation between surprisal reduction (FB) and dependency optimality (Omega). Squares denote the language-specific point estimates and horizontal lines indicate 95% confidence intervals. The vertical reference line marks $\rho = 0$.

**Confidence interval**

ar 0.29( 0.20 to 0.38)
ca 0.19( 0.08 to 0.29)
cs 0.33( 0.24 to 0.42)
de 0.12( 0.03 to 0.20)
el 0.40( 0.32 to 0.47)
en 0.17( 0.08 to 0.25)
es 0.45( 0.38 to 0.52)
eu 0.11(-0.01 to 0.22)
fa 0.16( 0.07 to 0.24)
fr 0.15( 0.06 to 0.24)
hu 0.20( 0.11 to 0.28)
id 0.30( 0.22 to 0.38)
ja 0.15( 0.06 to 0.23)
ko 0.18( 0.09 to 0.26)
nl 0.20( 0.11 to 0.28)
pl 0.32( 0.24 to 0.40)
ta 0.32( 0.16 to 0.47)
tr 0.16( 0.00 to 0.31)
vi 0.12( 0.04 to 0.21)
zh 0.23( 0.15 to 0.31)

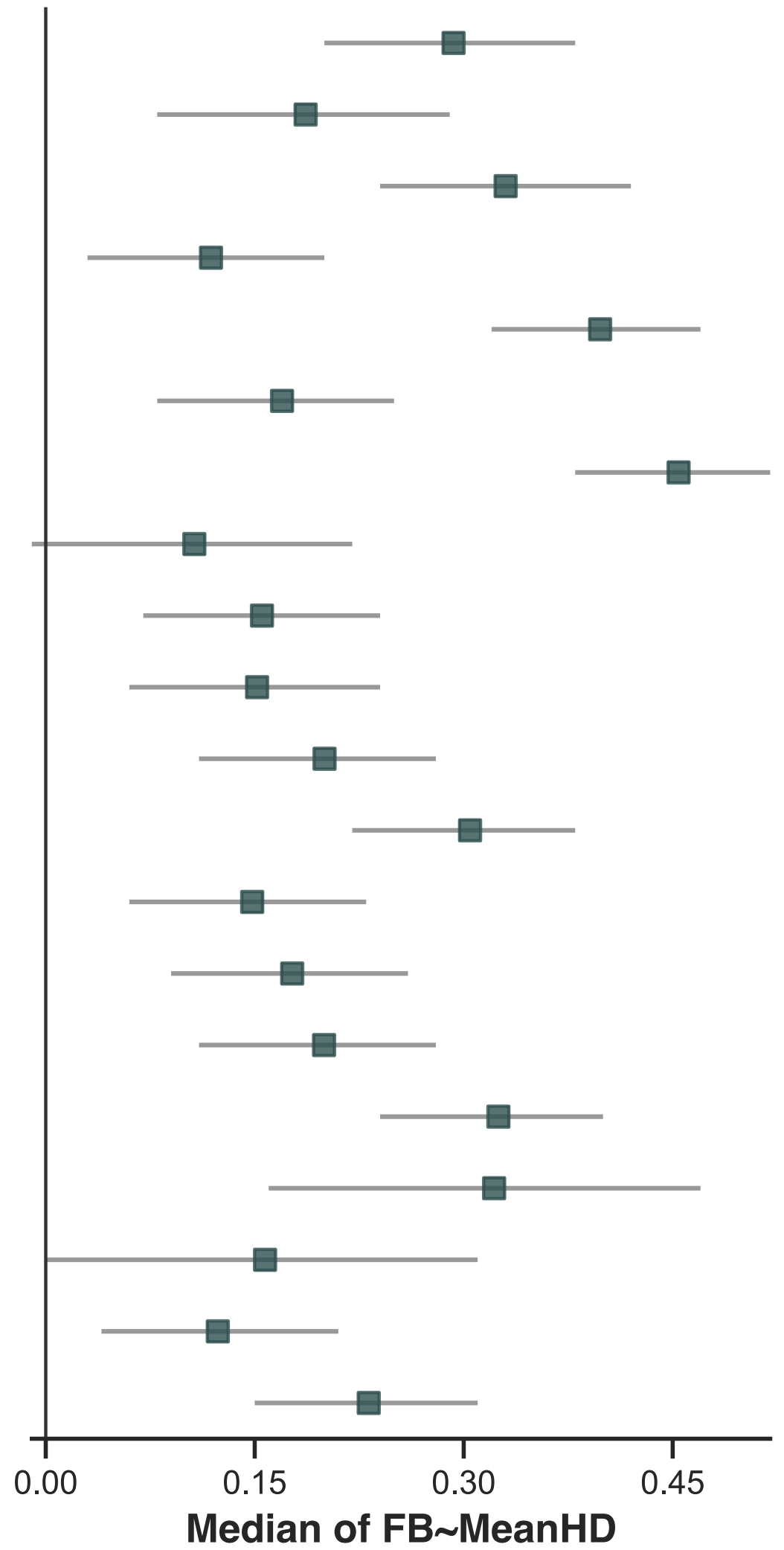

Supplementary Figure 5. Forest plot showing the correlation between surprisal reduction (FB) and mean hierarchical distance (MeanHD). Squares denote the language-specific point estimates and horizontal lines indicate 95% confidence intervals. The vertical reference line marks $\rho = 0$.

Confidence interval

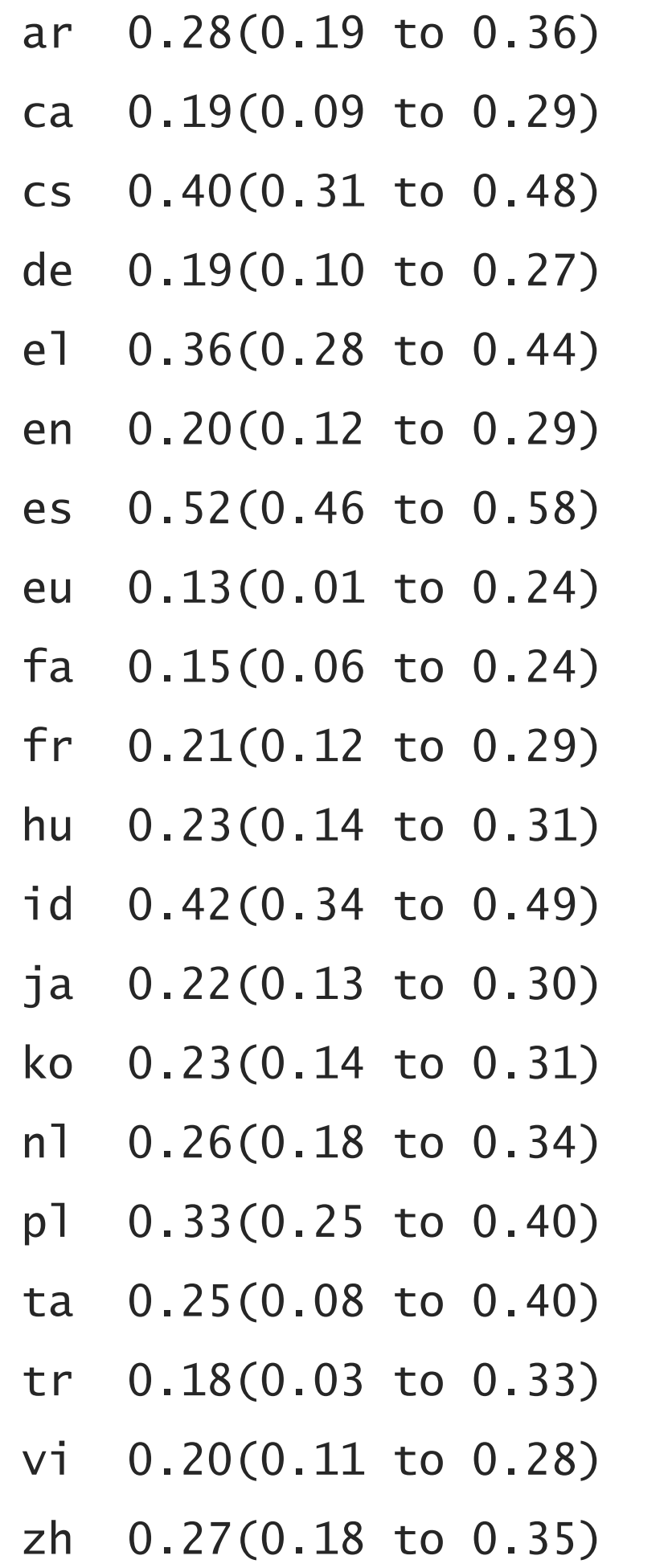

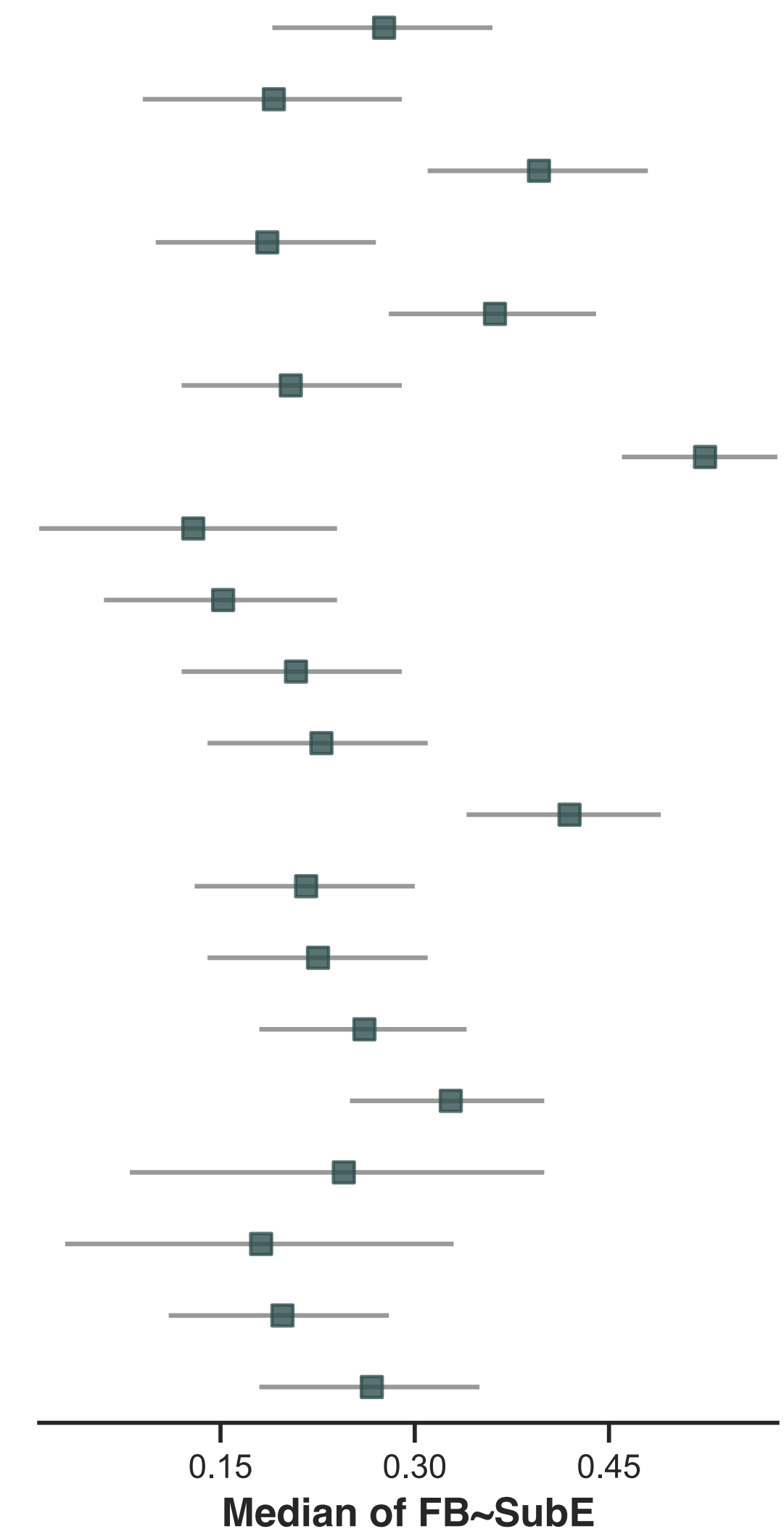

Supplementary Figure 6. Forest plot showing the correlation between surprisal reduction (FB) and subtree size unevenness (SubE). Squares denote the language-specific point estimates and horizontal lines indicate 95% confidence intervals. The vertical reference line marks $\rho = 0$.

**Confidence interval**

ar 0.19( 0.10 to 0.28)
ca 0.21( 0.11 to 0.31)
cs 0.28( 0.18 to 0.37)
de 0.15( 0.06 to 0.23)
el 0.27( 0.19 to 0.35)
en 0.18( 0.10 to 0.27)
es 0.42( 0.34 to 0.49)
eu 0.12( 0.00 to 0.23)
fa 0.12( 0.03 to 0.20)
fr 0.19( 0.11 to 0.27)
hu 0.17( 0.09 to 0.26)
id 0.38( 0.30 to 0.45)
ja 0.25( 0.16 to 0.33)
ko 0.17( 0.08 to 0.25)
nl 0.21( 0.12 to 0.29)
pl 0.16( 0.08 to 0.25)
ta 0.08(-0.09 to 0.25)
tr 0.11(-0.05 to 0.26)
vi 0.25( 0.17 to 0.33)
zh 0.22( 0.13 to 0.30)

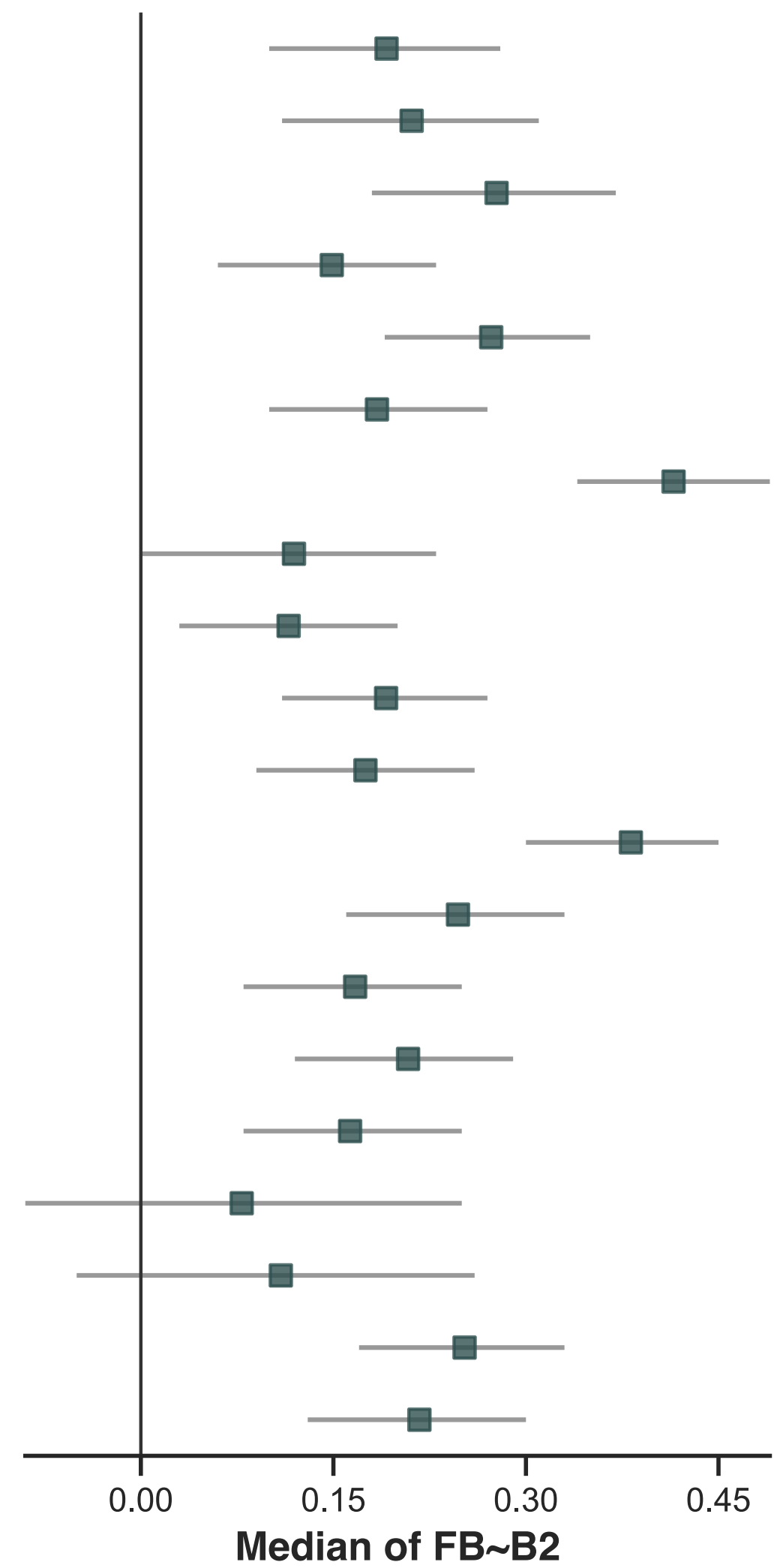

Supplementary Figure 7. Forest plot showing the correlation between surprisal reduction (FB) and B2 index (B2). Squares denote the language-specific point estimates and horizontal lines indicate 95% confidence intervals. The vertical reference line marks $\rho = 0$.

**Confidence interval**

| | |
|---|---|
| ar | -0.34(-0.42 to -0.25) |
| ca | -0.31(-0.40 to -0.21) |
| cs | -0.50(-0.57 to -0.42) |
| de | -0.29(-0.36 to -0.20) |
| el | -0.23(-0.31 to -0.15) |
| en | -0.18(-0.26 to -0.09) |
| es | -0.19(-0.28 to -0.11) |
| eu | -0.22(-0.33 to -0.11) |
| fa | -0.04(-0.13 to 0.05) |
| fr | -0.25(-0.33 to -0.17) |
| hu | -0.28(-0.36 to -0.20) |
| id | -0.06(-0.15 to 0.02) |
| ja | -0.23(-0.32 to -0.15) |
| ko | -0.01(-0.10 to 0.08) |
| nl | -0.27(-0.35 to -0.19) |
| pl | -0.27(-0.35 to -0.18) |
| ta | -0.37(-0.51 to -0.21) |
| tr | -0.46(-0.57 to -0.32) |
| vi | -0.05(-0.14 to 0.04) |
| zh | -0.20(-0.29 to -0.12) |

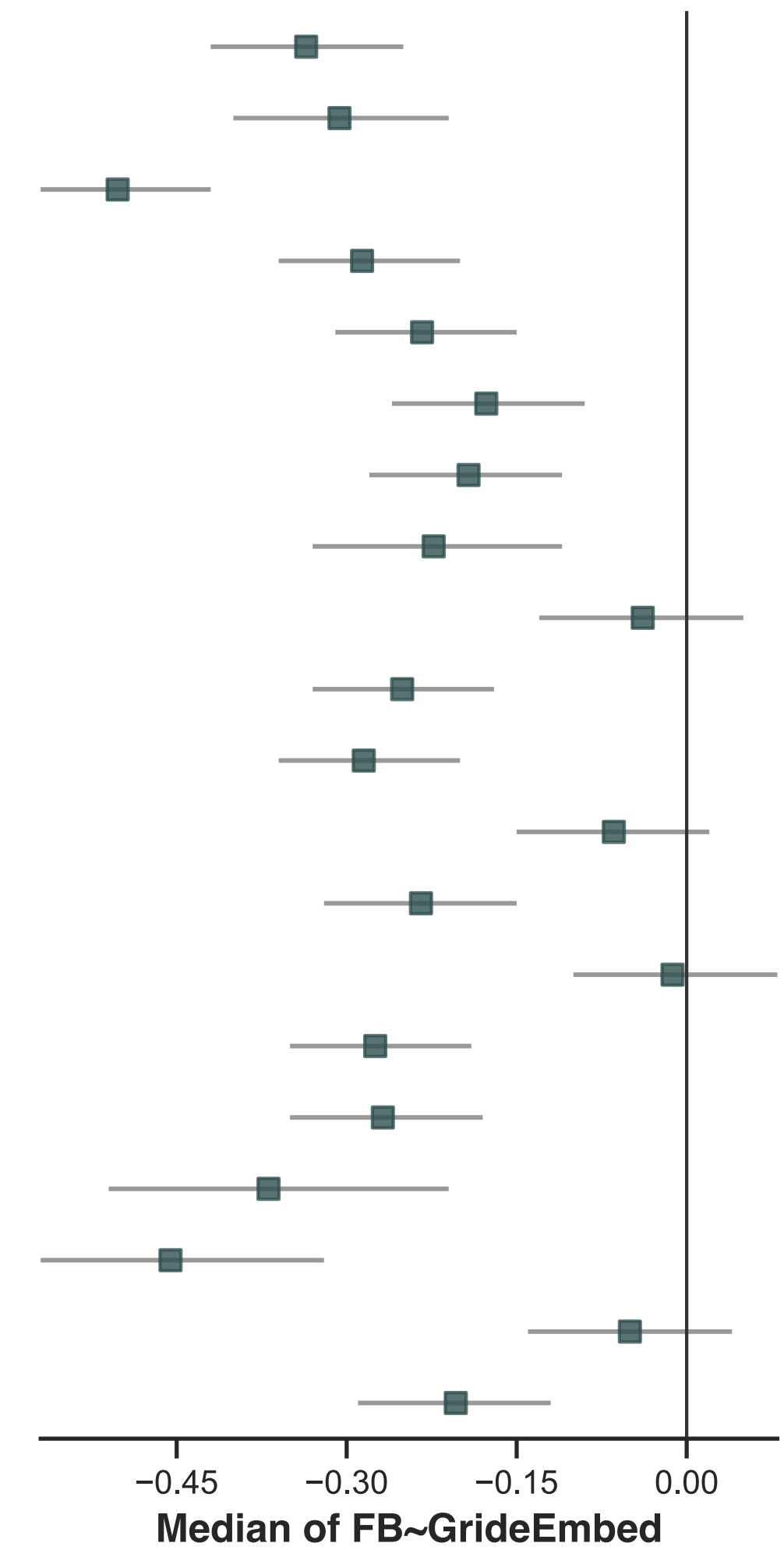

Supplementary Figure 8. Forest plot showing the correlation between surprisal reduction (FB) and intrinsic dimensions on the embedding feature manifold estimated by GRIDE (GrideEmbed). Squares denote the language-specific point estimates and horizontal lines indicate 95% confidence intervals. The vertical reference line marks $\rho = 0$.

**Confidence interval**

| | |
|---|---|
| ar | 0.30( 0.21 to 0.39) |
| ca | 0.36( 0.27 to 0.45) |
| cs | 0.55( 0.48 to 0.62) |
| de | 0.29( 0.21 to 0.37) |
| el | 0.22( 0.14 to 0.31) |
| en | 0.20( 0.11 to 0.28) |
| es | 0.12( 0.04 to 0.21) |
| eu | 0.23( 0.11 to 0.33) |
| fa | 0.02(-0.07 to 0.11) |
| fr | 0.24( 0.16 to 0.32) |
| hu | 0.33( 0.25 to 0.41) |
| id | 0.05(-0.04 to 0.13) |
| ja | 0.21( 0.12 to 0.29) |
| ko | 0.02(-0.07 to 0.11) |
| nl | 0.35( 0.27 to 0.42) |
| pl | 0.30( 0.21 to 0.37) |
| ta | 0.45( 0.31 to 0.58) |
| tr | 0.53( 0.40 to 0.63) |
| vi | 0.05(-0.04 to 0.14) |
| zh | 0.24( 0.15 to 0.32) |

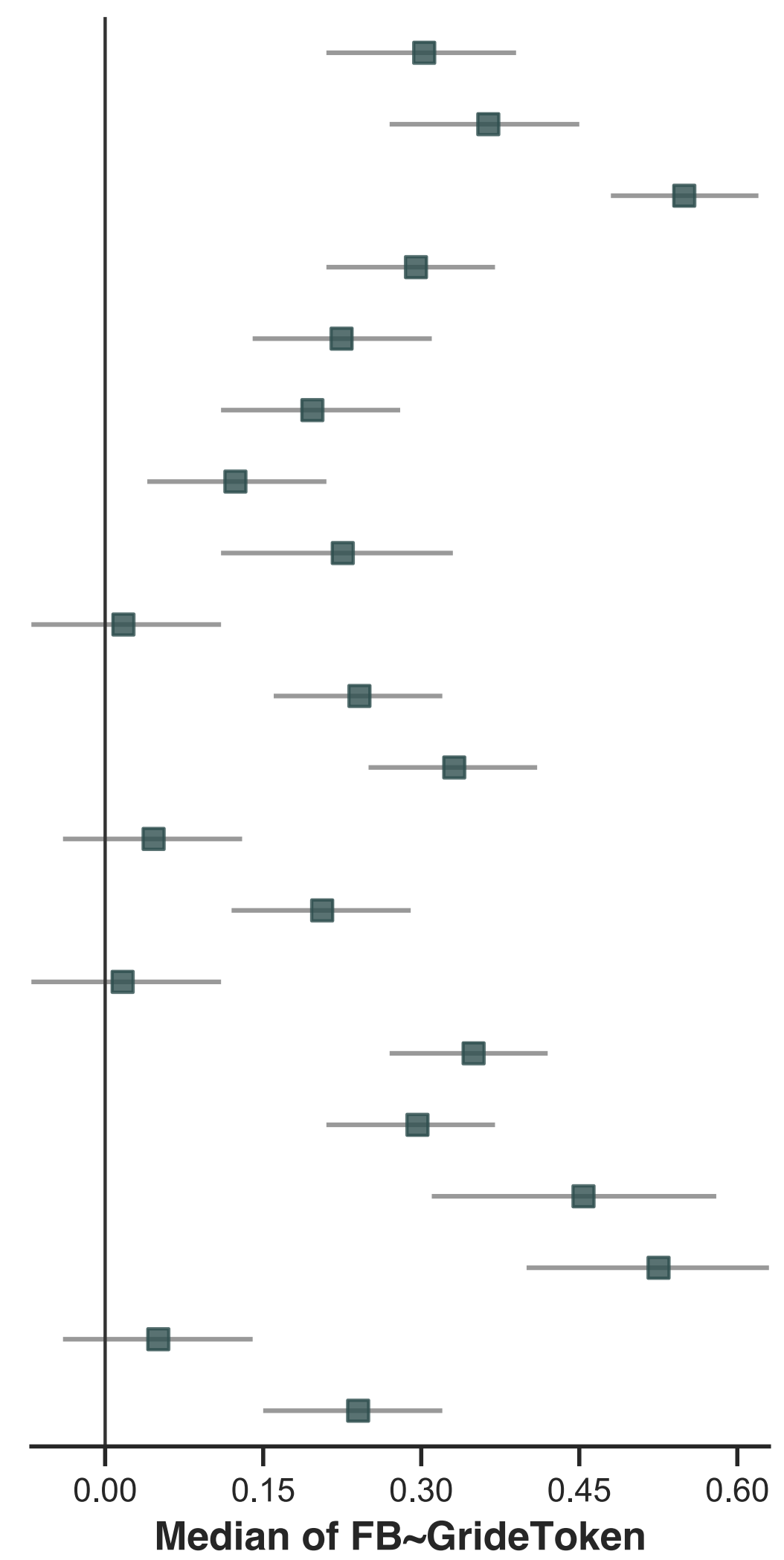

Supplementary Figure 9. Forest plot showing the correlation between surprisal reduction (FB) and intrinsic dimensions on the token manifold estimated by GRIDE (GrideToken). Squares denote the language-specific point estimates and horizontal lines indicate 95% confidence intervals. The vertical reference line marks $\rho = 0$.

**Confidence interval**

```
ar  -0.34(-0.42 to -0.25)
ca  -0.33(-0.43 to -0.24)
cs  -0.48(-0.55 to -0.40)
de  -0.29(-0.37 to -0.21)
el  -0.23(-0.32 to -0.15)
en  -0.22(-0.30 to -0.13)
es  -0.14(-0.23 to -0.06)
eu  -0.09(-0.21 to  0.02)
fa  -0.03(-0.12 to  0.06)
fr  -0.26(-0.34 to -0.17)
hu  -0.30(-0.38 to -0.22)
id  -0.06(-0.15 to  0.03)
ja  -0.17(-0.26 to -0.09)
ko  -0.03(-0.12 to  0.06)
nl  -0.35(-0.42 to -0.27)
pl  -0.31(-0.39 to -0.23)
ta  -0.44(-0.57 to -0.29)
tr  -0.45(-0.56 to -0.31)
vi  -0.06(-0.14 to  0.03)
zh  -0.29(-0.37 to -0.20)
```

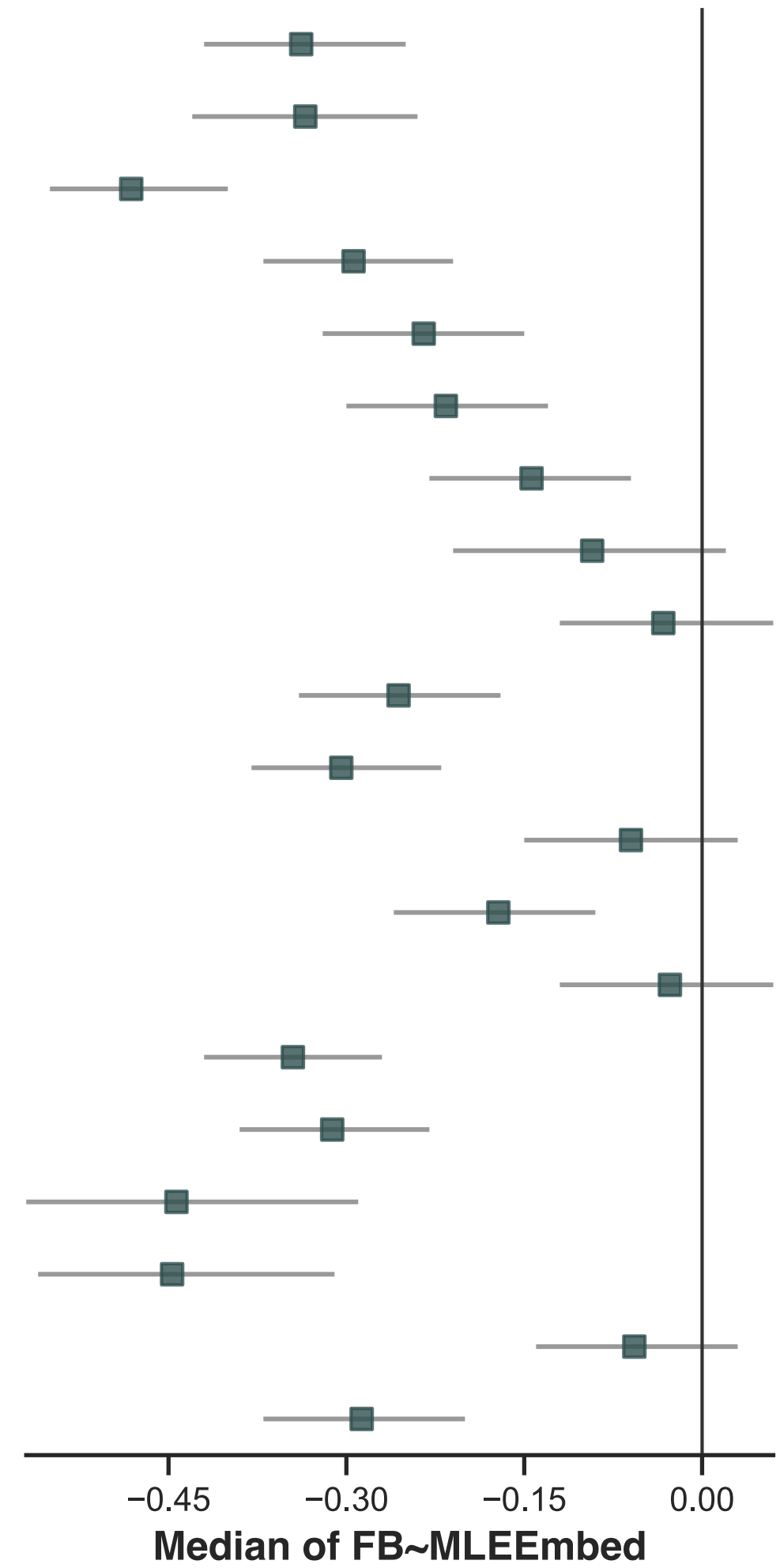

Supplementary Figure 10. Forest plot showing the correlation between surprisal reduction (FB) and intrinsic dimensions on the embedding feature manifold estimated by MLE (MLEEmbed). Squares denote the language-specific point estimates and horizontal lines indicate 95% confidence intervals. The vertical reference line marks $\rho = 0$.

**Confidence interval**

ar 0.30( 0.21 to 0.39)
ca 0.36( 0.27 to 0.45)
cs 0.55( 0.48 to 0.62)
de 0.30( 0.22 to 0.38)
el 0.22( 0.14 to 0.31)
en 0.20( 0.11 to 0.28)
es 0.12( 0.03 to 0.21)
eu 0.22( 0.11 to 0.33)
fa 0.01(-0.07 to 0.10)
fr 0.24( 0.16 to 0.32)
hu 0.33( 0.25 to 0.40)
id 0.05(-0.04 to 0.14)
ja 0.20( 0.12 to 0.28)
ko 0.02(-0.07 to 0.11)
nl 0.35( 0.27 to 0.43)
pl 0.30( 0.22 to 0.38)
ta 0.45( 0.30 to 0.58)
tr 0.52( 0.40 to 0.63)
vi 0.05(-0.04 to 0.14)
zh 0.24( 0.16 to 0.33)

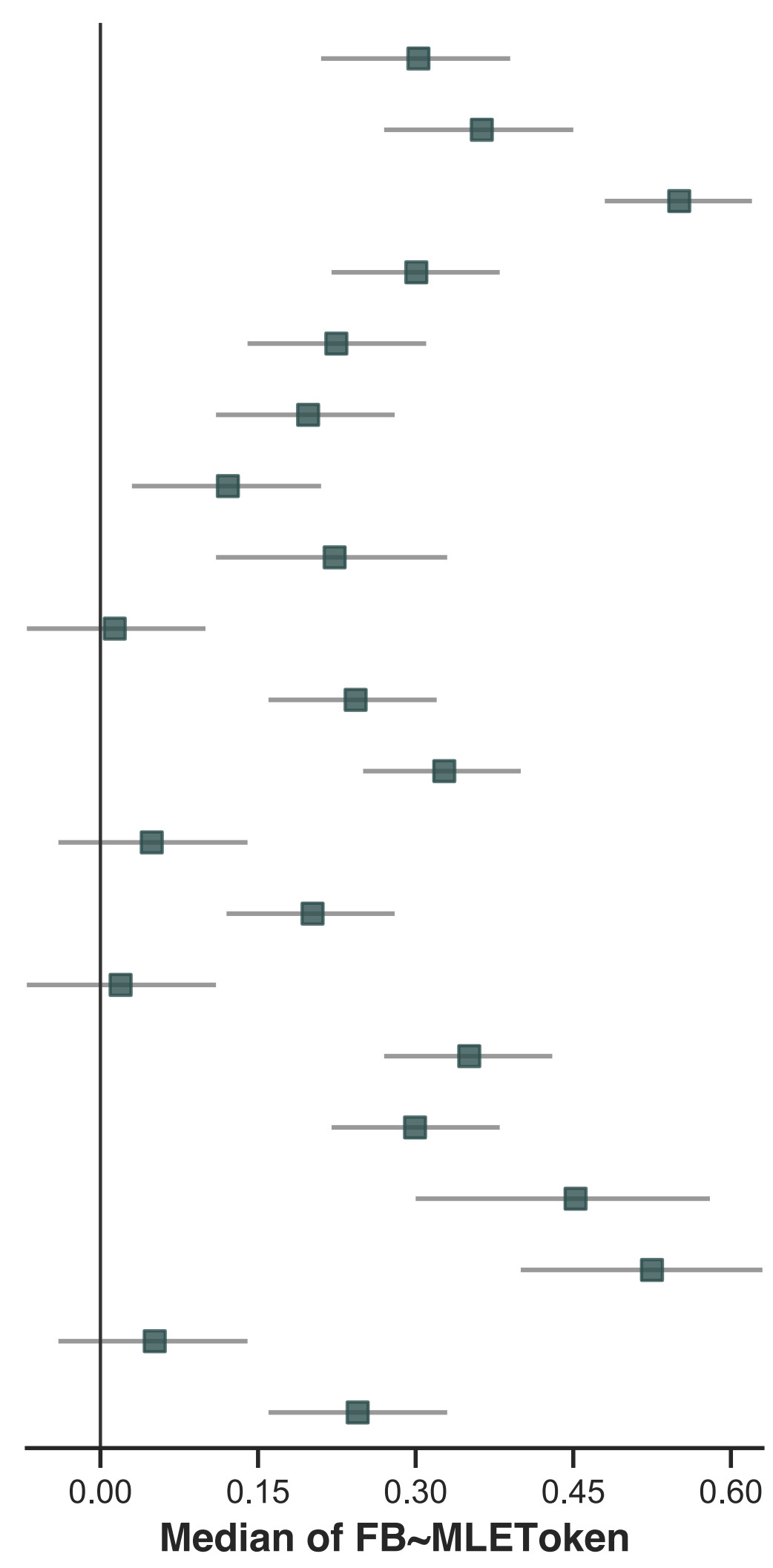

Supplementary Figure 11. Forest plot showing the correlation between surprisal reduction (FB) and intrinsic dimensions on the token manifold estimated by MLE (MLEToken). Squares denote the language-specific point estimates and horizontal lines indicate 95% confidence intervals. The vertical reference line marks $\rho = 0$.

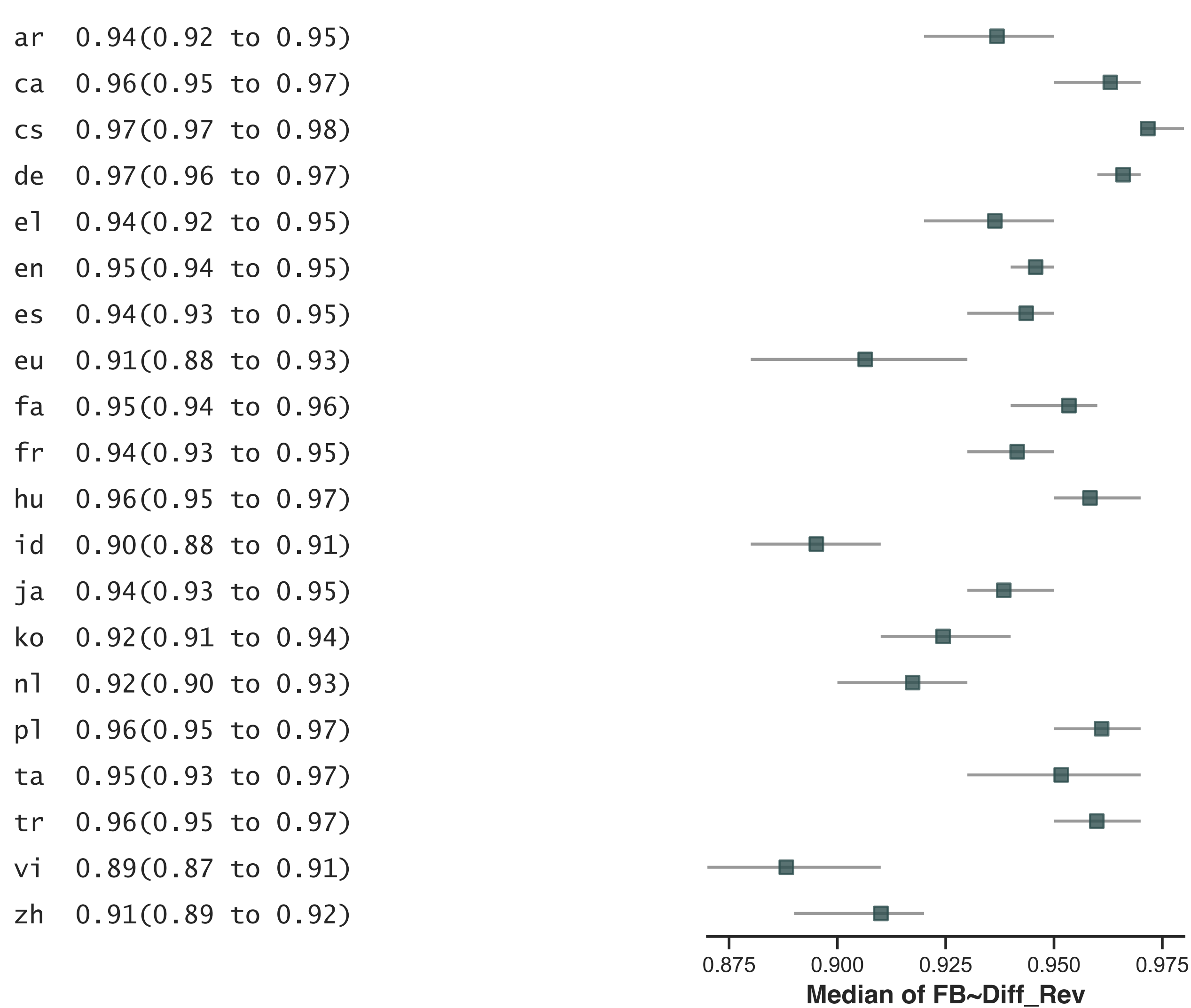

Supplementary Figure 12. Forest plot showing the correlation between surprisal reduction (FB) and the surprisal reduction from normal to reversed word order (Diff_Rev). Squares denote the language-specific point estimates and horizontal lines indicate 95% confidence intervals.

## 5 Validation with generative language models

We replicated BERT-based analysis using Qwen3 (Qwen/Qwen3-4B), a multilingual generative language model (GLM) with different model architecture and much larger parameter size. Similar to BERT, given a sentence, we encoded it into subword tokens, and the probability of each subword token, $p_{\mathrm{GLM}}(t_i \mid x_{<i})$, was computed by masking each subword token and all the tokens after it, running the model's forward pass, and extracting the probability assigned to the true subword at the masked position. These probabilities were converted to subword surprisal $-\log_2 p_{\mathrm{GLM}}(t_i \mid x_{<i})$. Zero probabilities were floored at $10^{-9}$. Word-level contextual surprisal was then computed as the sum of subword surprisals aligned to each word, while skipping punctuation tokens (i.e., tokens present in the punctuation-aware list but absent from the punctuation-removed list). Sentence-level contextual surprisal was computed as the sum of word-level surprisal across the sentence.

$$S_{\mathrm{GLM}}(x) = -\sum_{i=1}^{N(x)} \log_2 p_{\mathrm{GLM}}\left( t_i \mid x_{<i} \right)$$

where $x_{<i}$ denotes the sentence with only preceding tokens. All analyses were conducted at the text level to enable cross-lingual comparisons under a common unit of observation. For each text $T = (x_1, \dots, x_{n(T)})$, where each sentence represented as a token sequence $x = (t_1, \dots, t_{N(x)})$, we first computed surprisal values and syntactic measures at the sentence level. ID values, as noted before, were computed at the text level. Syntactic measures were averaged across sentences. Number of tokens, as well as the three surprisal values, $S_{\mathrm{freq}}(x)$, $S_{\mathrm{GLM}}(x)$, and $S_{\mathrm{GLM}}(x^{rev})$, were summed across sentences within the text. Then, we normalized the text-level surprisal (i.e. the sum of the sentence-level surprisal) by total number of words of the text $N(T)$, thus we have:

$$\overline{T_{\mathrm{freq}}}(x) = \frac{1}{N(T)} \sum_{i=1}^{n(T)} S_{\mathrm{freq}}(x)$$

$$\overline{T_{\mathrm{GLM}}}(x) = \frac{1}{N(T)} \sum_{i=1}^{n(T)} S_{\mathrm{GLM}}(x)$$

$$\overline{T_{\mathrm{GLM}}}(x^{\mathrm{rev}}) = \frac{1}{N(T)} \sum_{i=1}^{n(T)} S_{\mathrm{GLM}}(x^{\mathrm{rev}})$$

To quantify the reduction of uncertainty from grammar-insensitive to grammar-sensitive conditions, we derived a normalized surprisal-reduction index contrasting the lexical with contextual surprisal ($\Delta Diff_{\mathrm{FG}}(x)$), and an analogous index contrasting the reversed-order (grammar-distorted) with the intact sentences ($\Delta Diff_{\mathrm{REV}}(x)$).

$$\Delta Diff_{\mathrm{FG}}(x) = \frac{\overline{T_{\mathrm{freq}}}(x) - \overline{T_{\mathrm{GLM}}}(x)}{\overline{T_{\mathrm{freq}}}(x)}$$

$$\Delta Diff_{\mathrm{REV}}(x) = \frac{\overline{T_{\mathrm{GLM}}}(x^{\mathrm{rev}}) - \overline{T_{\mathrm{GLM}}}(x)}{\overline{T_{\mathrm{GLM}}}(x^{\mathrm{rev}})}$$

Similarly, we observed uncertainty reduction in all 20 languages except Basque, as shown in Supplementary Figure 13. This divergent pattern may reflect weaker performance of Qwen3 in Basque (much higher surprisal values compared to other languages), a relatively low-resource language. To examine this possibility, we additionally tested a comparable model that had been specifically fine-tuned for Basque (orai-nlp/Gemma-Kimu-2b-it), and replicated the BERT-based result, as shown in Supplementary Figure 14.

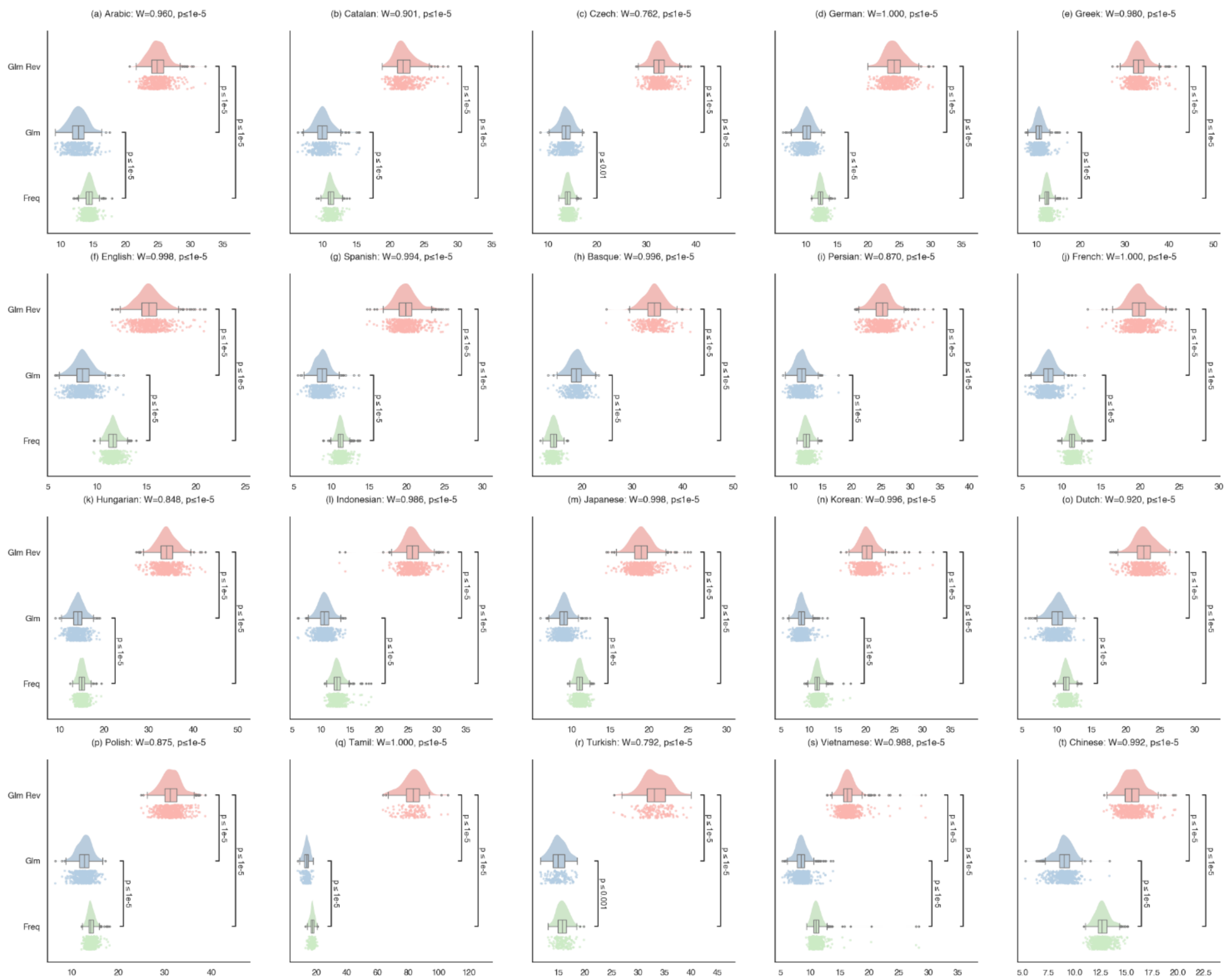

Supplementary Figure 13. Uncertainty estimation across 20 languages with a multilingual Qwen3 (a–t). For each language, distributions of sentence-level surprisal are shown for frequency-based estimates (Freq, green), contextual surprisal in the original order (Glm, blue), and contextual surprisal for reversed sentences (Glm Rev, red). Violin plots with overlaid points represent individual texts. Differences across conditions were tested using Friedman's test (reported in each panel title) with Siegel's post-hoc pairwise comparisons (reported in the brackets). Across all languages, contextual surprisal is lower than frequency-based surprisal, and reversing word order reliably increases contextual surprisal.

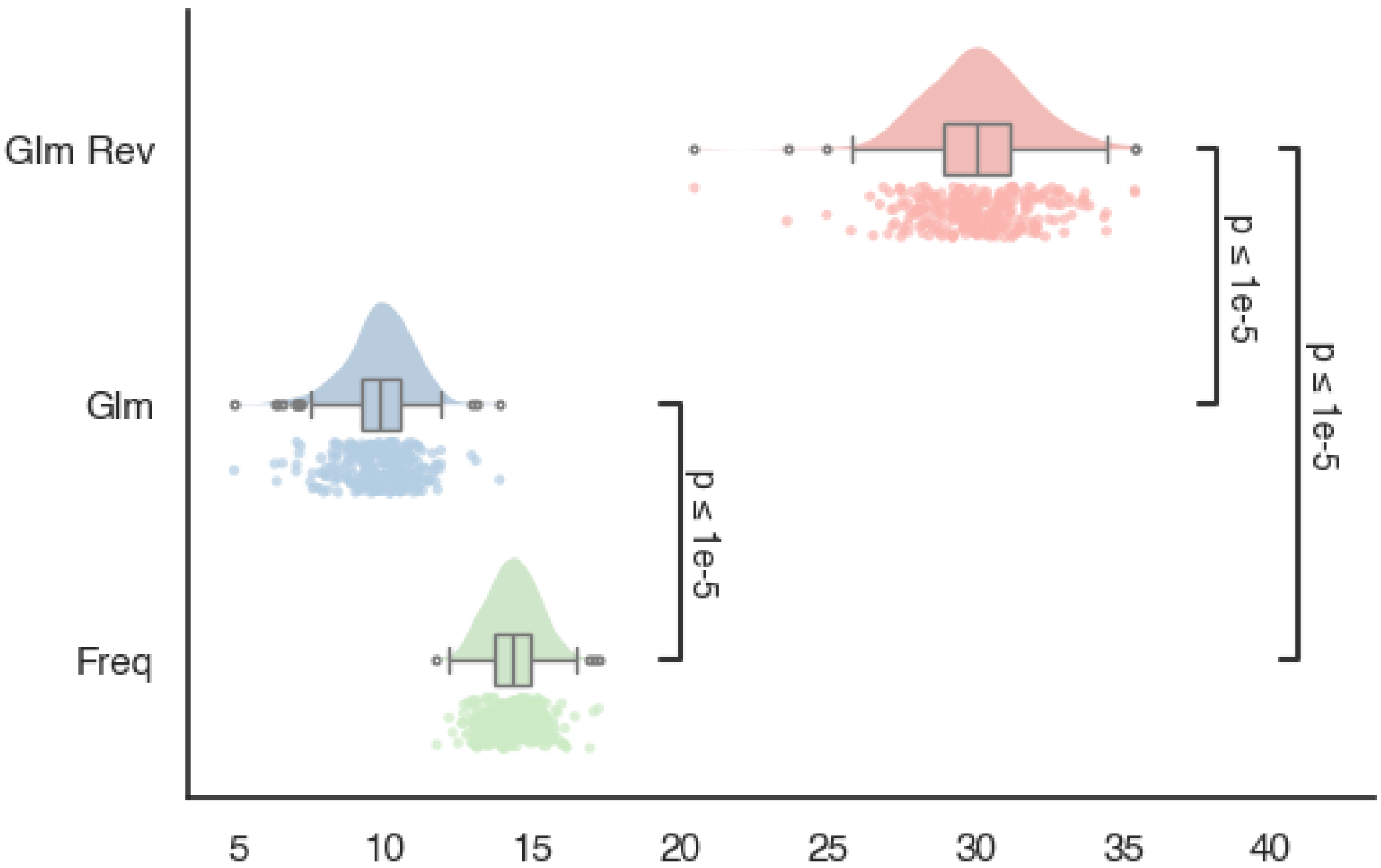

Supplementary Figure 14. Uncertainty estimation in Basque with a Gemma2 model pretrained on Basque (orai-nlp/Gemma-Kimu-2b-it). Distributions of sentence-level surprisal are shown for frequency-based estimates (Freq, green), contextual surprisal in the original order (Glm, blue), and contextual surprisal for reversed sentences (Glm Rev, red). Violin plots with overlaid points represent individual texts. Differences across conditions were tested using Friedman's test (reported in each panel title) with Siegel's post-hoc pairwise comparisons (reported in the brackets). Across all languages, contextual surprisal is lower than frequency-based surprisal, and reversing word order reliably increases contextual surprisal.

Correlations with syntactic and semantic features were also generally replicated, with generally stronger correlation strength than the BERT results, as in Supplementary Figure 15.

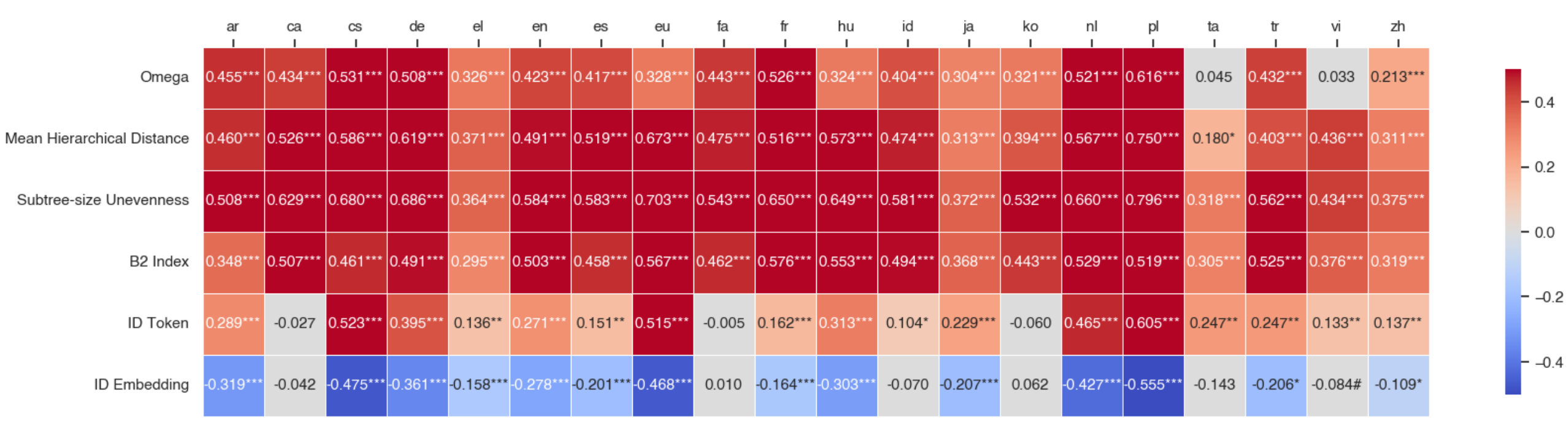

Supplementary Figure 15. Heatmap of Spearman correlations between normalized surprisal reduction and each structural or semantic metric across 20 languages (columns). Colored cells indicate significant correlations between uncertainty reduction and linguistic variables (rows) in a certain language data (columns). Warm colors indicate positive correlations, cool colors negative correlations, and asterisks mark FDR-corrected significance levels. * *p*<0.05, ** *p*<0.01, *** *p*<0.001.

Meta analyses also yielded similar results. Across all 20 languages, surprisal reduction closely tracked the surprisal cost from sentence reversal (meta-analytic Spearman's $\rho$=0.769, 95% CI: 0.731 to 0.807). Surprisal reduction was positively correlated with B2 index (meta-analytic Spearman's ρ=0.459, 95% CI: 0.422 to 0.497), subtree-size unevenness (meta-analytic Spearman's ρ=0.566, 95% CI: 0.510 to 0.621), dependency distance optimality (meta-analytic Spearman's ρ=0.386, 95% CI: 0.322 to 0.449), mean hierarchical distance (meta-analytic Spearman's ρ=0.489, 95% CI: 0.433 to 0.544), and intrinsic dimensionality on the token manifold (meta-analytic Spearman's ρ=0.243, 95% CI: 0.160 to 0.326), and negatively correlated with intrinsic dimensionality on the embedding manifold (meta-analytic Spearman's ρ=-0.227, 95% CI: -0.303 to -0.151). All of these results replicated the correlation direction as in the main BERT-based analyses.

## 6 Surprisal comparisons in clinical cohorts

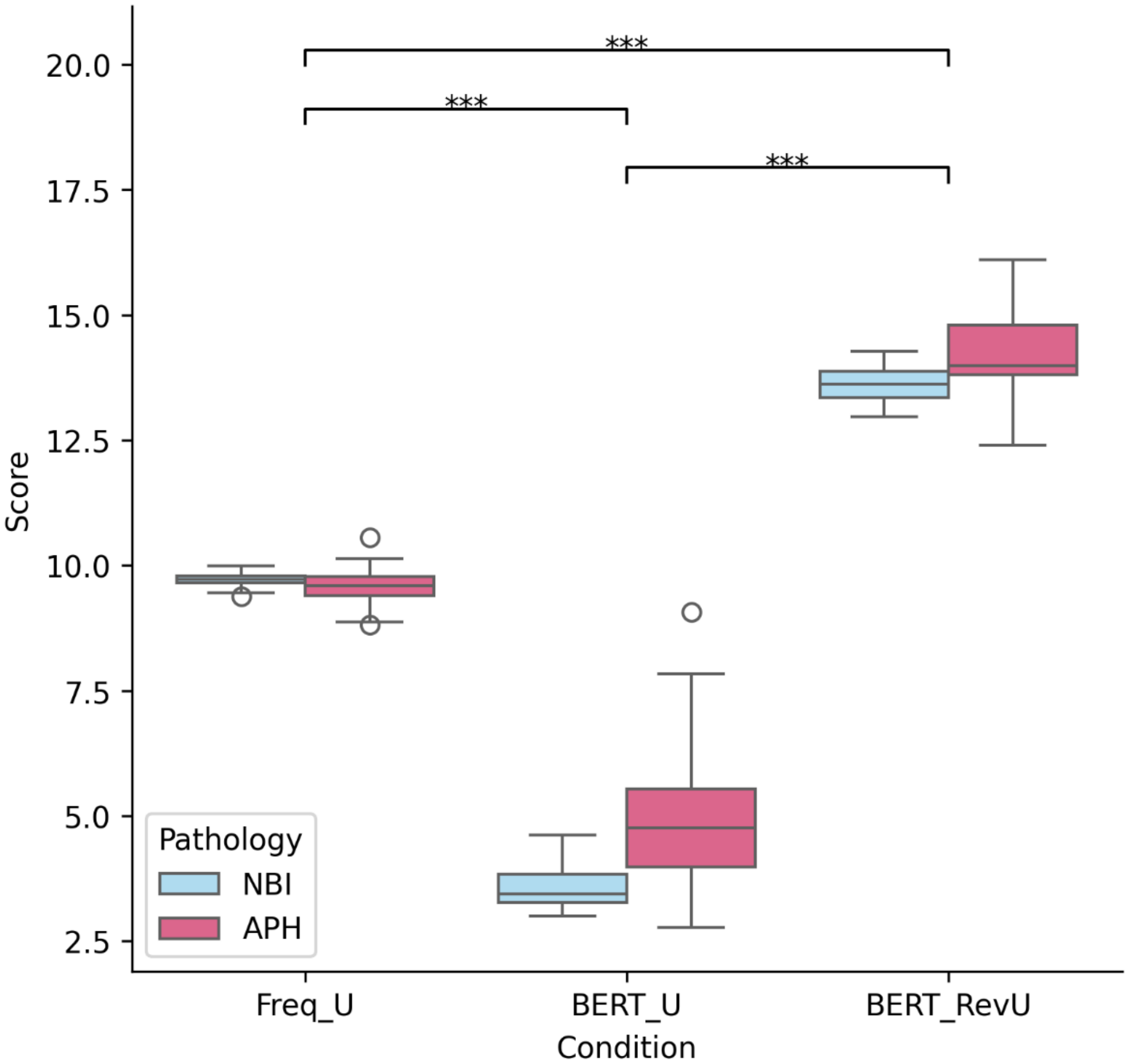

Supplementary Figure 16. Surprisal estimation in the Olness aphasia cohort. Distributions of sentence-level surprisal are shown for frequency-based estimates (Freq_U), contextual surprisal in the original order (BERT_U), and contextual surprisal for reversed sentences (BERT_RevU). Differences across conditions were tested using Friedman's test with Siegel's post-hoc pairwise comparisons (reported in the brackets). Different colors indicate different clinical groups: non brain injury (NBI), and aphasia (APH).

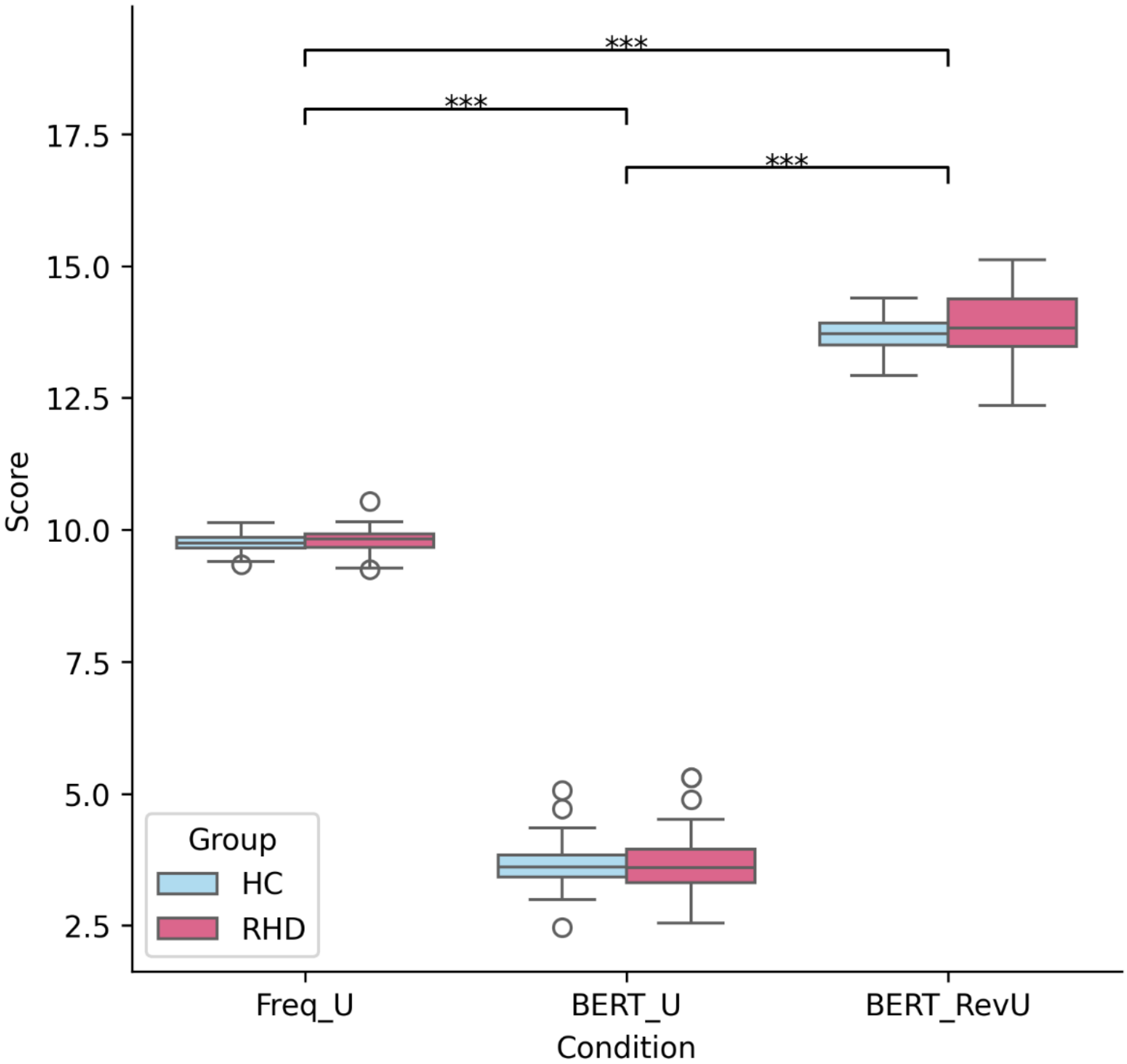

Supplementary Figure 17. Surprisal estimation in the Minga RHD cohort. Distributions of sentence-level surprisal are shown for frequency-based estimates (Freq_U), contextual surprisal in the original order (BERT_U), and contextual surprisal for reversed sentences (BERT_RevU). Differences across conditions were tested using Friedman's test with Siegel's post-hoc pairwise comparisons (reported in the brackets). Different colors indicate different clinical groups: health control (HC), and right hemisphere damage (RHD).

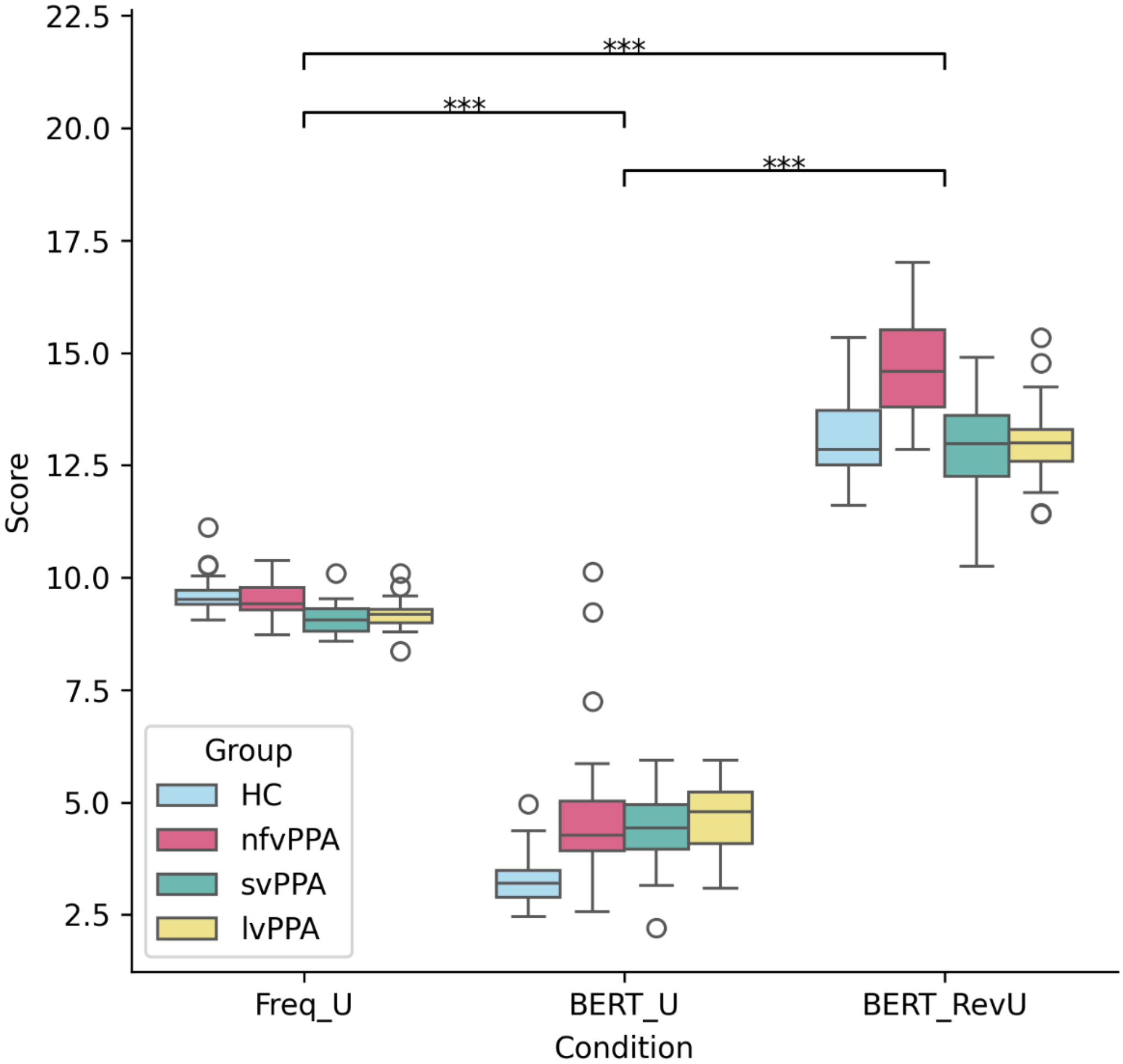

Supplementary Figure 18. Surprisal estimation in the Rezaii PPA cohort. Distributions of sentence-level surprisal are shown for frequency-based estimates (Freq_U), contextual surprisal in the original order (BERT_U), and contextual surprisal for reversed sentences (BERT_RevU). Differences across conditions were tested using Friedman's test with Siegel's post-hoc pairwise comparisons (reported in the brackets). Different colors indicate different clinical groups: health control (HC), non-fluent variant of primary progressive aphasia (nfvPPA), semantic variant of primary progressive aphasia (svPPA) and logopenic variant of primary progressive aphasia (lvPPA).

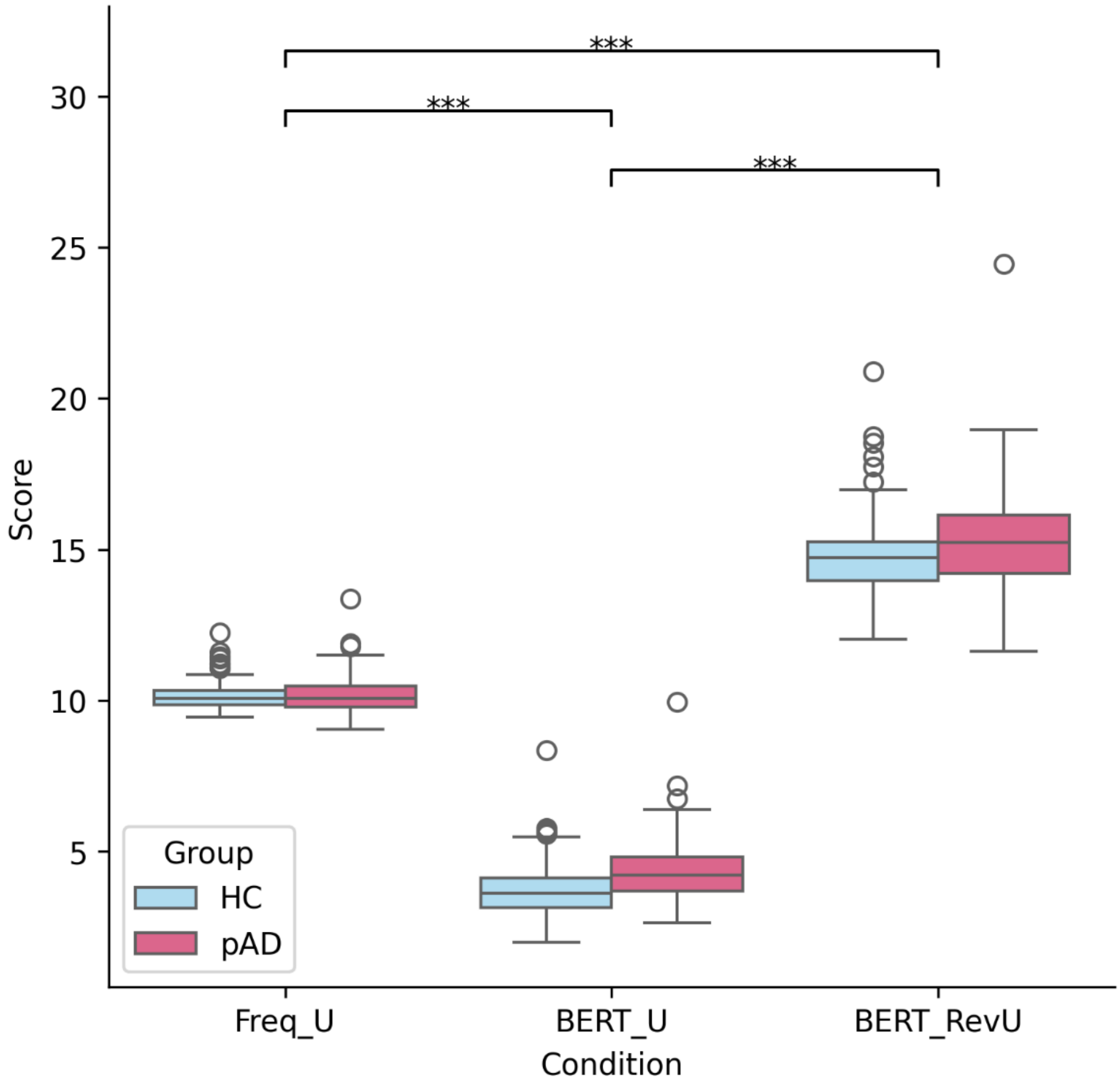

Supplementary Figure 19. Surprisal estimation in the ADReSS dataset. Distributions of sentence-level surprisal are shown for frequency-based estimates (Freq_U), contextual surprisal in the original order (BERT_U), and contextual surprisal for reversed sentences (BERT_RevU). Differences across conditions were tested using Friedman's test with Siegel's post-hoc pairwise comparisons (reported in the brackets). Different colors indicate different clinical groups: health control (HC), and probable Alzheimer's disease (pAD).

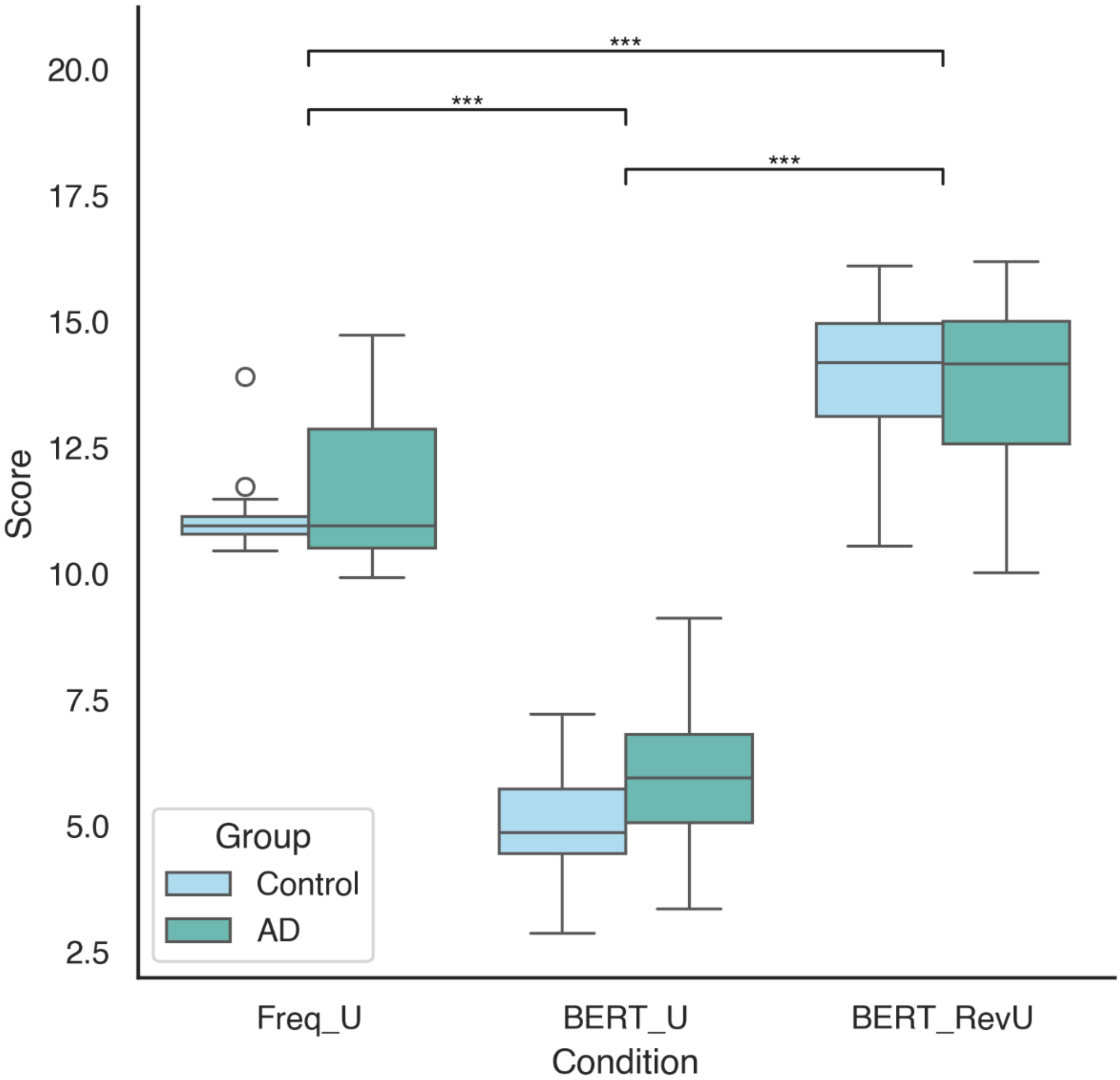

Supplementary Figure 20. Surprisal estimation in the Zhuhai AD dataset. Distributions of sentence-level surprisal are shown for frequency-based estimates (Freq_U), contextual surprisal in the original order (BERT_U), and contextual surprisal for reversed sentences (BERT_RevU). Differences across conditions were tested using Friedman's test with Siegel's post-hoc pairwise comparisons (reported in the brackets). Different colors indicate different clinical groups: health control (HC), and Alzheimer's disease (AD).

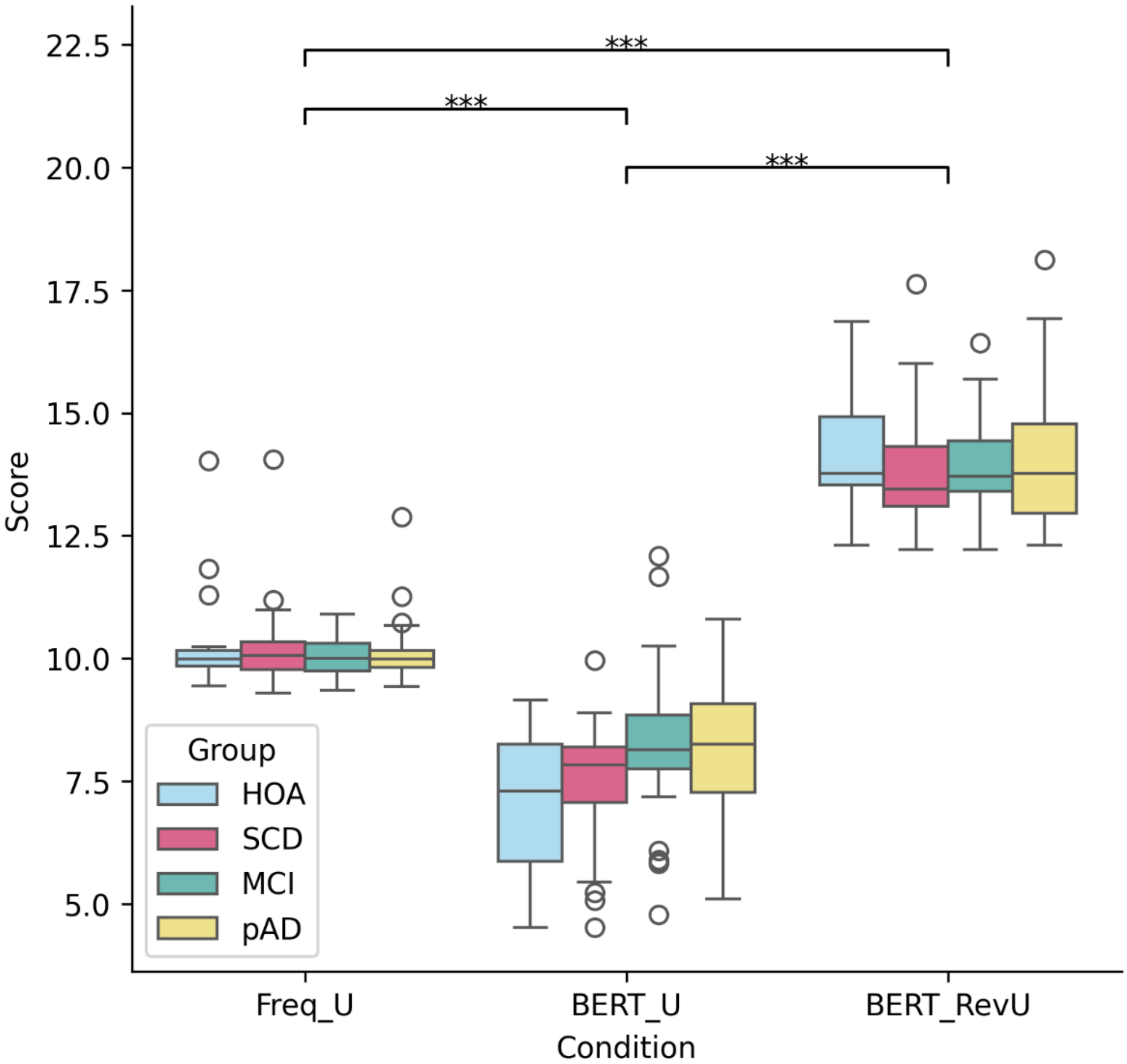

Supplementary Figure 21. Surprisal estimation in the ACE corpus. Distributions of sentence-level surprisal are shown for frequency-based estimates (Freq_U), contextual surprisal in the original order (BERT_U), and contextual surprisal for reversed sentences (BERT_RevU). Differences across conditions were tested using Friedman's test with Siegel's post-hoc pairwise comparisons (reported in the brackets). Different colors indicate different clinical groups: health older adults (HOA), subjective cognitive decline (SCD), mild cognitive impairment (MCI), and probable Alzheimer's disease (pAD).

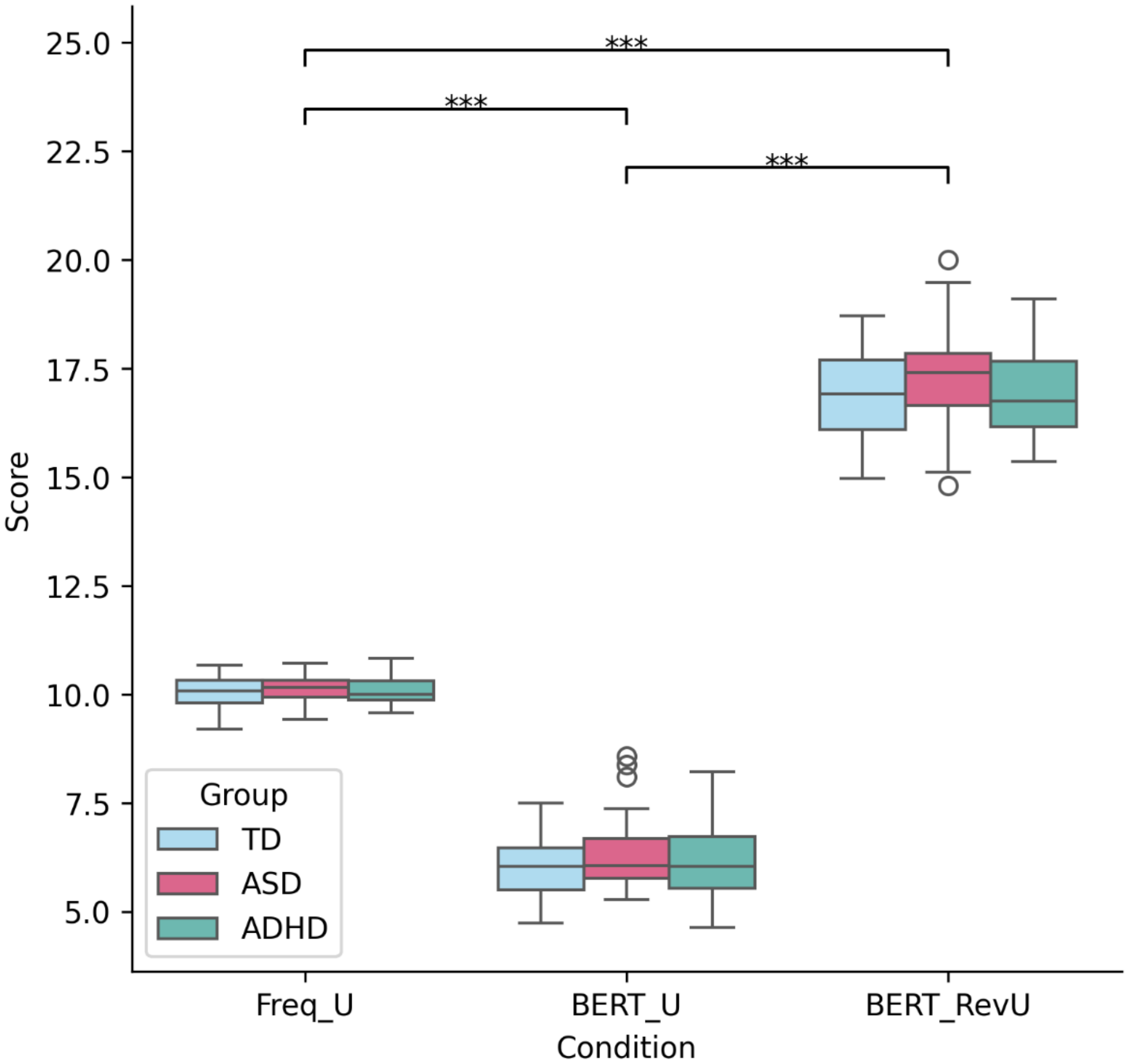

Supplementary Figure 22. Surprisal estimation in the Asymmetries corpus. Distributions of sentence-level surprisal are shown for frequency-based estimates (Freq_U), contextual surprisal in the original order (BERT_U), and contextual surprisal for reversed sentences (BERT_RevU). Differences across conditions were tested using Friedman's test with Siegel's post-hoc pairwise comparisons (reported in the brackets). Different colors indicate different clinical groups: typical development (TD), autism spectrum disorder (AS) and attention deficit hyperactivity disorder (ADHD).

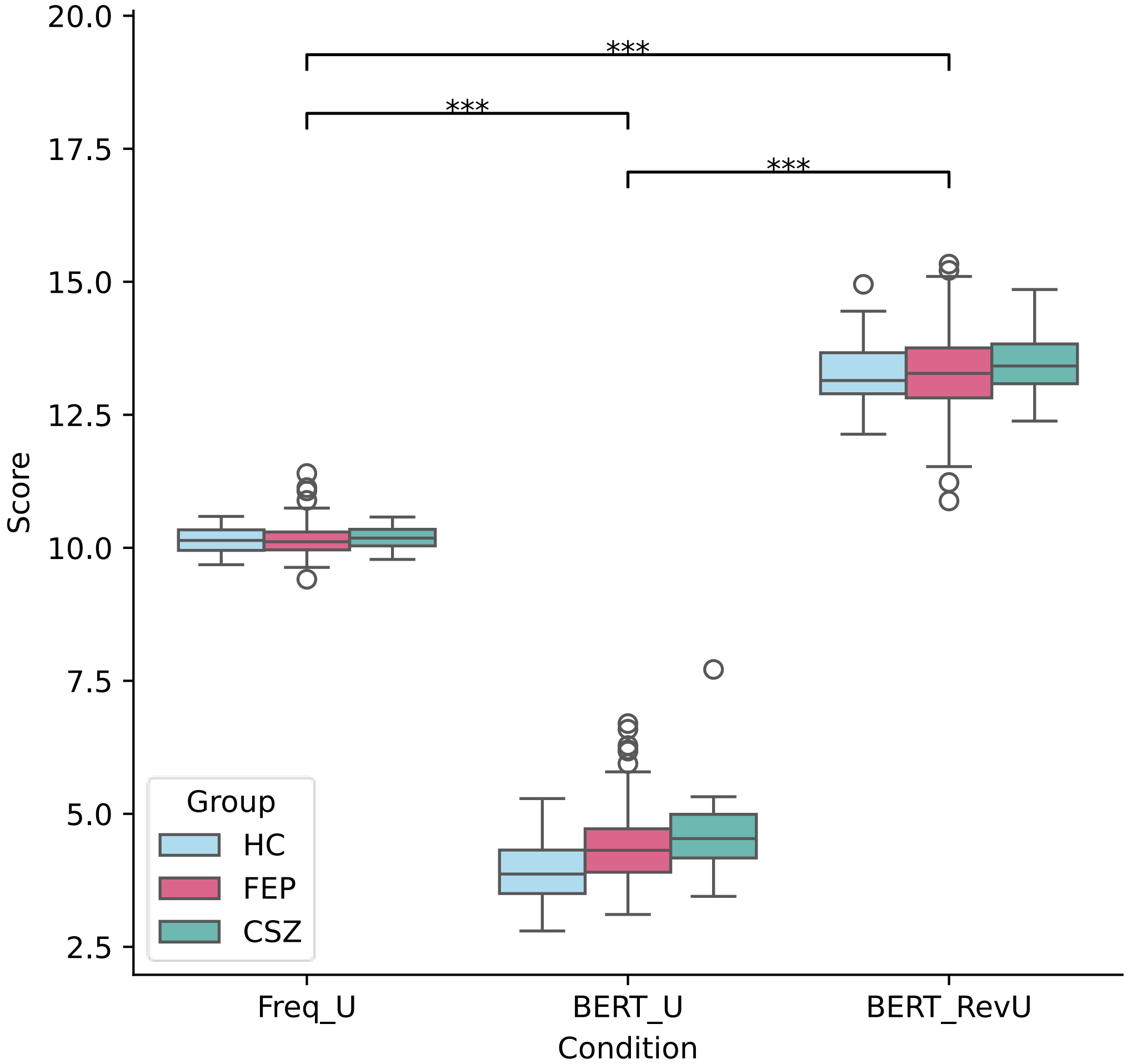

Supplementary Figure 23. Surprisal estimation in the TOPSY dataset. Distributions of sentence-level surprisal are shown for frequency-based estimates (Freq_U), contextual surprisal in the original order (BERT_U), and contextual surprisal for reversed sentences (BERT_RevU). Differences across conditions were tested using Friedman's test with Siegel's post-hoc pairwise comparisons (reported in the brackets). Different colors indicate different clinical groups: health control (HC), first-episode psychosis (FEP), and chronic schizophrenia (CSZ).

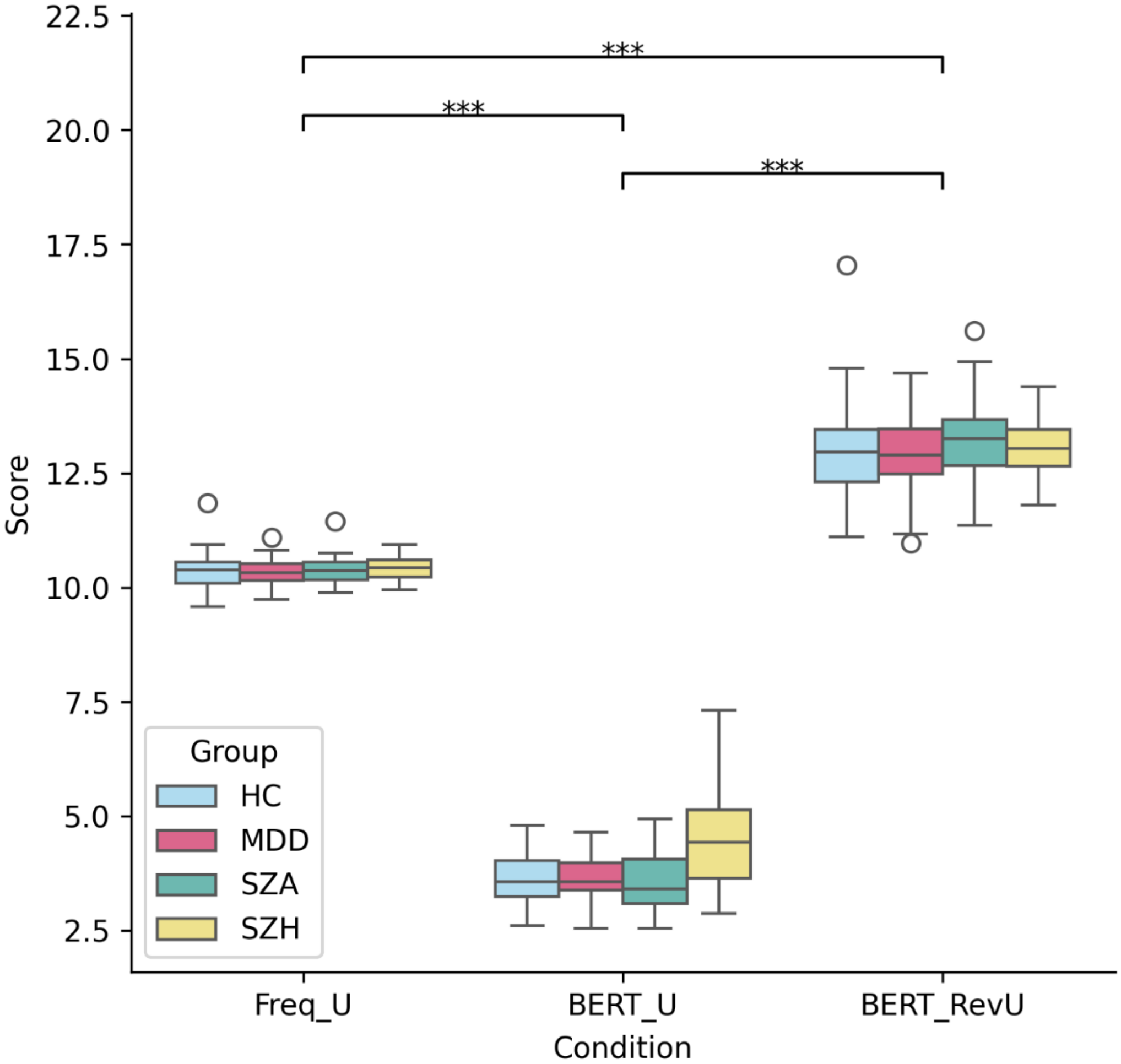

Supplementary Figure 24. Surprisal estimation in the Marburg cohort. Distributions of sentence-level surprisal are shown for frequency-based estimates (Freq_U), contextual surprisal in the original order (BERT_U), and contextual surprisal for reversed sentences (BERT_RevU). Differences across conditions were tested using Friedman's test with Siegel's post-hoc pairwise comparisons (reported in the brackets). Different colors indicate different clinical groups: health control (HC), major depressive disorder (MDD), schizoaffective (SZA), and schizophrenia (SZH).